\documentclass[runningheads]{llncs}
\usepackage[T1]{fontenc}
\usepackage{graphicx}
\usepackage{version}
\usepackage{multirow}
\usepackage{rotating}
\usepackage{amssymb,amsmath}
\usepackage{mathtools}
\DeclarePairedDelimiter\floor{\lfloor}{\rfloor}
\usepackage{hyperref}
\usepackage{enumerate}
\usepackage{xcolor}
\graphicspath{{./test/img/}}

\newcommand{\SSS}{\mathcal{S}}
\newcommand{\GG}{\mathcal{G}}

\newcommand{\llbracket}{[\![}
\newcommand{\rrbracket}{]\!]}

\usepackage{color}
\usepackage{version}
\usepackage{multirow}

\usepackage{amssymb}
\usepackage{textcomp}
\usepackage{xcolor}
\usepackage{url}
\usepackage{mathtools}
\definecolor{review}{rgb}{0, 0, 0}
\definecolor{review2}{rgb}{0, 0, 0}
\definecolor{review3}{rgb}{0, 0, 0}
\definecolor{rodrigo_review}{rgb}{0, 1, 0}

\graphicspath{{./img/}}

\newcommand{\eqsize}{\normalsize}
\newcommand{\PP}{\mathbf{P}}
\newcommand{\sa}{S_i^{a\rightarrow \{\mathbb{N}^{2}\}}}

\begin{document}
\title{Generalized Shortest Path-based Superpixels for 3D Spherical Image Segmentation}
\author{R{\'e}mi Giraud\inst{1} %\orcidID{0000-1111-2222-3333} 
\and
Rodrigo Borba Pinheiro\inst{1}
\and
Yannick Berthoumieu\inst{1}%\orcidID{1111-2222-3333-4444}
}
\authorrunning{R. Giraud et al.}
\institute{
Univ. Bordeaux, Bordeaux INP, IMS, CNRS UMR 5218, France.\\ 
\email{remi.giraud@ims-bordeaux.fr}
}
\maketitle              % typeset the header of the contribution

\begin{abstract}
The growing use of wide angle image capture devices
and the need for fast and accurate image analysis in computer visions
have enforced the need for dedicated under-representation approaches.
Most recent decomposition methods segment an image into a small number of irregular homogeneous regions, called \textit{superpixels}.
Nevertheless, these approaches are generally designed to segment standard 2D planar images, \emph{i.e.},
captured with a 90$^\text{o}$ angle view without distortion.
In this work, we introduce a new general superpixel method called SphSPS (for Spherical Shortest Path-based Superpixels)\footnote{Available code at: \url{https://github.com/rgiraud/sphsps}}, dedicated
to wide 360$^\text{o}$ spherical or omnidirectional images.
Our method respects the geometry of the 3D spherical acquisition space and
generalizes the notion of shortest path between a pixel and a superpixel center,
to fastly extract relevant clustering features.
We demonstrate that considering the geometry of the acquisition space to compute
the shortest path enables to jointly improve the
segmentation accuracy and the shape regularity of superpixels.
To evaluate this regularity aspect,
we also generalize a global regularity metric to the spherical space,
addressing the limitations of the only existing spherical compactness measure.
Finally, the proposed SphSPS method is validated
on the reference 360$^\text{o}$ spherical panorama segmentation dataset and on synthetic road omnidirectional images.
{\color{review2}
Our method significantly outperforms both planar and spherical state-of-the-art approaches in terms of segmentation accuracy, robustness to noise and regularity, providing a very interesting tool for superpixel-based applications on 360\textsuperscript{o} images.
}
\end{abstract}

\section{Introduction}

Many computer vision pipelines now
use under-representation or low-level segmentation approaches
to overcome the growth in resolution and quantity of image data,
which may lead to an important computational load.
Among existing approaches,
irregular image decomposition techniques
were mainly popularized with \cite{achanta2012},
to decompose an image into \textit{superpixels}, \emph{i.e.},
small, connected regions having homogeneous colors.
The image domain is thus generally reduced to hundreds of regions instead of millions of pixels.
By processing such regions in an under-representation scale that fits to the image content,
the result may be obtained in a very fast manner while being very close to the optimal result at pixel scale.
{\color{review}
The scale of the under-representation, \emph{i.e.}, the number of superpixels,
is generally application-driven, regarding the data quality and the processing time requirements.
While larger high quality images, containing many objects may
require
% necessitate
more superpixels to accurately fit to the image content, a higher number of regions may lead to a higher processing time.
Hence, the superpixel approach is particularly interesting, since it offers a direct control on the number of elements to process.
}

Over the years, superpixel methods have been successfully applied for many
computer vision tasks such as:
semantic segmentation \cite{tighe2010,wang2013med,priya2015superpixels,farag2016bottom},
object tracking \cite{reso2013,oron2015locally,lee2018tracking},
optical flow estimation \cite{menze2015object,wang2019semflow}
or style transfer \cite{liu2017photo}.
However, with such under-representation,
the structural irregularity between adjacent regions may become an issue when using
standard neighborhood-based tools.
Several works have then proposed different methods to address
this setback, \emph{e.g.},
graph-based approach \cite{gould2014},
neighborhood structure \cite{giraud2017_spm},
or
hierarchical superpixel decomposition \cite{nakamura2017hierarchical}.
{\color{review}
From this perspective,
the shape regularity of the superpixels is a particularly interesting  property
for the extraction of
consistent neighborhoods between regions.
Generally, the optimal superpixel decomposition is a trade-off between the shape and size regularity of regions,
and their ability to fit to the image content, \emph{i.e.}, object borders or color changes.
These aspects must be jointly evaluated to asses the relevance of a superpixel decomposition method.
}

Additionally to the increase of image resolution,
new acquisition devices capturing wide angles,
such as fish eyes, or covering a 360$^\text{o}$ field of view
have become increasingly popular.
With these devices, the entire environment can be captured
to offer a global understanding of a scene.
Hence, in applications such as autonomous driving,
spherical or omnidirectional images are particularly interesting and may be
used for semantic segmentation \cite{yang2020omnisupervised,yang2021context}.
Moreover, with the joint use of a depth-aware system,
the captured intensity can be projected on a 3D point cloud, offering additional modalities.
Generally, the captured 360$^\text{o}$ environment is
projected on a discrete 2D plane, generating an equirectangular image.
Naturally, such projection on a rectangular space introduces distortions \cite{zorin1995correction}.

{\color{review2}
In this context, standard planar superpixel algorithms have been used
on such equirectangular images \cite{cabral2014piecewise,sakurada2015change}
for instance for 3D scene reconstruction applications.
Nevertheless, as stated in \cite{da20223d},
using planar approaches is not entirely suited since they
do not consider the geometry distortions in the projected image.
This may lead to poor segmentation accuracy and difficulties in interpreting the planar over-segmentation
in the spherical acquisition space around the poles, \emph{i.e.}, top and bottom of the equirectangular image.
}

{\color{review}
A first method has been proposed in \cite{zhao2018}, adapting the standard SLIC algorithm \cite{achanta2012} to spherical images. The generated decomposition considers the geometry of the 360\textsuperscript{o} acquisition and computes superpixels that seem irregular in the planar space, but regular in the spherical one.
However,
since the method is based on a straightforward adaptation of \cite{achanta2012}, that considers very simple features and clustering distance, it suffers from the same limitations, \emph{i.e.},
poor segmentation accuracy, robustness to noise, and ability to capture thin object contours.
Moreover, \cite{zhao2018} introduced a spherical regularity evaluation metric based on the planar circularity measure \cite{schick2012}, that appears to present the same limitations.
These have been demonstrated and addressed in \cite{giraud2017_jei} for the planar case.
}

\smallskip

\begin{figure*}[t]
 \centering
 \includegraphics[width=0.95\textwidth]{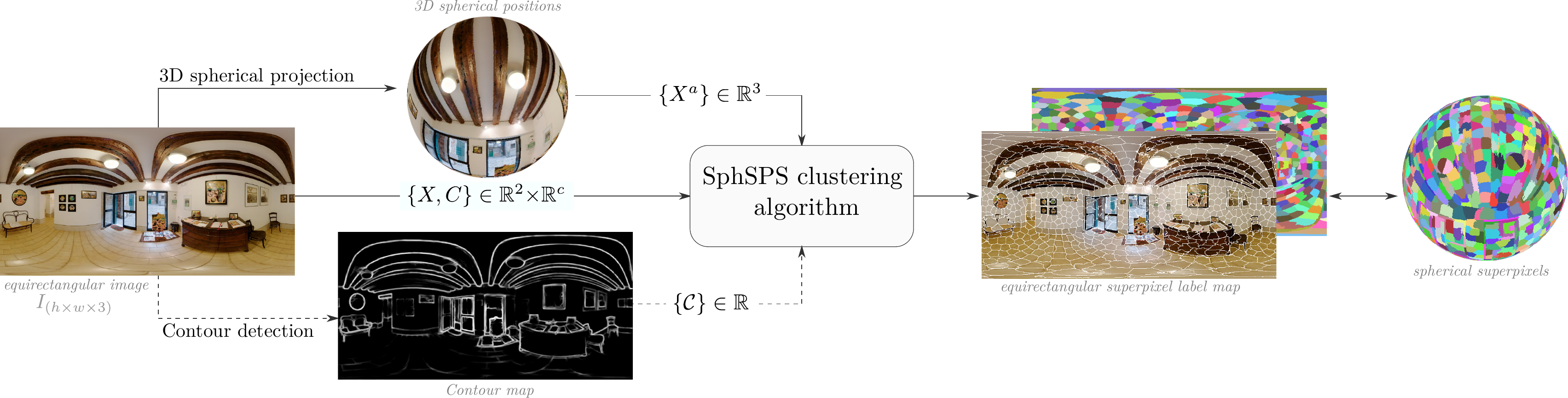}
 \caption{Pipeline of the SphSPS method.
 A projection in the spherical space is computed from the equirectangular image $I$ to obtain 3D coordinates $X^a$.
Along with 2D positions $X$ and colors $C$ ($c$ channels), a contour map $\mathcal{C}$ can be provided (dotted arrows) to
enforce the respect of image contours. SphSPS provides regular superpixels in the spherical space}
\label{fig:scheme}
\end{figure*}

{\color{review2}
 \subsubsection*{Contributions}
%\textit{Contributions:}

In this work, we propose an accurate method called SphSPS (Spherical Shortest Path-based Superpixels), to decompose a 2D equirectangular image into 3D spherically regular superpixels.
The core contributions of this work are listed as follows:
\begin{itemize}
 \item (i) New superpixel segmentation framework for spherical images;
\item (ii) Adaptation of a spherical shortest path algorithm to extract features on equirectangular images;
\item (iii) Optimized recursive computation of such spherical shortest paths;
\item (iv) New superpixel shape regularity metric for validation;
\item (v) Evaluation of algorithms on a standard natural and a new synthetic dataset.
\end{itemize}

(i)
The pipeline of the SphSPS method, considering both 2D and 3D positions is summarized in Figure \ref{fig:scheme}.
The core of
SphSPS is based on the adaptation of the $K$-means iterative clustering framework \cite{achanta2012} to the spherical space by \cite{zhao2018}, but addresses its limitations in terms of accuracy, robustness and computational time.

(ii)
To this end, we generalize of the notion of shortest path between a pixel and a superpixel \cite{giraud2018_scalp} to the spherical case
to extract more relevant features during the clustering process.
With such path, when comparing a pixel to a superpixel, the color of the pixels along the path to the superpixel barycenter can be considered to ensure the color homogeneity.
Moreover, contour information provided by any prior contour map can also be integrated to prevent the segmentation from crossing object contours in our method.

(iii) To extract such features in a fast manner, we
use an algorithm inspired of \cite{shoemake1985animating} and
also provide an optimized recursive implementation of the spherical shortest path. Hence, SphSPS is able to converge in very limited processing time and generate accurate and regular spherical superpixels (see Figure \ref{fig:scheme}).

(iv) To relevantly evaluate the regularity aspect in the spherical space, we also propose a generalization of the global regularity (GR) metric \cite{giraud2017_jei}, that addresses the limitations of the previously existing spherical metric  \cite{zhao2018}.

(v)
The performance of SphSPS is compared to the ones of planar and spherical state-of-the-art methods on the
reference 360$^\text{o}$ spherical panorama segmentation dataset (PSD) \cite{wan2018} and on a novel set of synthetic omnidirectional road images from the Omniscape project \cite{sekkat2020omniscape}.
SphSPS obtains the highest segmentation and contour detection accuracy while
providing spherically regular superpixels,
and is robust to noise contrary to most state-of-the-art methods. \\

}

This paper is an extension of \cite{giraud2020_icpr},
with significant improvements and extensive additional experiments.
Compared to \cite{giraud2020_icpr}, we provide:
(1) more details on our method, with a substantial
analysis of its parameters (\emph{e.g.}, sampling frequency of the shortest path, initialization strategies),
(2) a modification of %the discrete downsampling step in the computation of
the Generalized Global Regularity (G-GR) metric to more relevantly evaluate the smoothness of the superpixel boundaries, along with a quantitative comparison to the spherical compactness measure proposed in \cite{zhao2018};
(3) an evaluation of the robustness of methods to noise,
including other state-of-the-art approaches (ETPS \cite{yao2015}, GMMSP \cite{Ban18} and SphLSC \cite{chen2017,zhao2018})
and comparison to an implementation of SphSPS using a linear path;
(4) an additional validation on a synthetic 360$^\text{o}$ road images \cite{sekkat2020omniscape}.

{\color{review}
\section{Related Works}

The majority of existing superpixel methods
concern standard 2D planar images.
To achieve such irregular decompositions,
grouping spatially close pixels with homogeneous colors,
many approaches have been proposed using
region growing \cite{levinshtein2009},
graph-based energy \cite{liu2011},
eikonal-based \cite{buyssens2014eikonal},
watershed \cite{machairas2015},
coarse-to-fine \cite{yao2015},
or even bayesian algorithms \cite{Uziel:ICCV:2019:BASS}.
Some of these methods may offer interesting properties,
in terms of exact control of the number of generated superpixels
or the ability to set the spatial regularity constraint to produce more or less grid-like decompositions.
Among the vast
literature of superpixel works, a
significant breakthrough was obtained with the SLIC algorithm \cite{achanta2012}.
This iterative method applies a pixel-wise $K$-means algorithm
to a small area around each superpixel barycenter, which are initialized as a regular grid.
Pixels are thus naturally grouped according to a trade-off between
distances in the spatial and color spaces (CIELab).
The method requires the approximate number of superpixel to generate and a spatial constraint term to produce a decomposition in a reduced processing time.
Nevertheless, SLIC may fail to jointly capture object borders
and provide regular shapes, while being highly sensitive to texture and noise.
Many methods have been proposed based on the SLIC algorithm, introducing
boundary constraint \cite{zhang2016},
advanced feature space \cite{chen2017},
non-iterative clustering \cite{achanta2017superpixels},
or a shortest path approach \cite{giraud2018_scalp}.

More recently, deep learning frameworks have also been proposed
to decompose an image into superpixels, \emph{e.g.},
\cite{jampani2018superpixel,yang2020superpixel}.
In \cite{jampani2018superpixel} for instance, the SLIC algorithm is made differentiable to learn adequate image features.
{\color{review2}
First, contrary to most other methods, they may not allow to set the shape regularity
which may be an important property to set according to the application \cite{giraud2017_jei}.
Then, along with the usual deep learning requirements
in terms of computational resources and substantial training time and image dataset,
these methods may have limited applicability to other images.
They may also even fail to handle large ones due to memory issue.
$K$-means-based approaches such as SLIC or the proposed SphSPS  do
not necessitate any learning and their clustering distance is computed on the image features.
Therefore, they are consistent with any other dataset,
and does not have restriction on the image dimension.
Their complexity is simply linear with the image dimension.
Hence, unsupervised approaches remain of interest,
while they may still consider features extracted from deep learning-based pipelines as input.
}

For spherical images,
the segmentation approach of \cite{felzenszwalb2004} for 2D planar images has been adapted  in \cite{yang2016efficient}. The approach consists in using a graph segmentation to generate segments in the image. First, a graph $G=\{V,E\}$ is constructed, where $V$ is the set of pixels in the image and $E$ is the set of edges that connects neighboring pixels. Each edge $E$ has a corresponding weight that is dependant on the dissimilarity between the two pixels connected by it. The edges are sorted by non-decreasing weight and the vertices are merged accordingly.
%%Nevertheless, it uses the same graph-based
 For the 3D approach the main difference is the addition of the edges connecting the pixels from the left and right boundaries. Although being able to segment the image, this clustering method generates
%clustering that generates
very irregular regions in terms of shape and size that may not be considered as superpixels.

The $K$-means-based iterative clustering method of SLIC \cite{achanta2012} has been extended in \cite{zhao2018} to generate spherically regular superpixels, \emph{i.e.}, having more consistent size and shape in the spherical domain.
The initial sampling of the superpixel barycenters is adapted to cover the sphere area.
The pixel positions are then projected on the unit sphere to compute the spatial constraints to gather homogeneous pixels not spatially too far in the spherical space.
With the interest of having visually regular superpixels when
working in the spherical domain, the respect of the acquisition space geometry
enables a more accurate segmentation of the image objects \cite{zhao2018}.

Recently, based on the same framework, a non-iterative and a hierarchical extension have been proposed in \cite{da2021fast}.
The non-iterative extension is based on the SNIC algorithm \cite{achanta2017superpixels}, a similar method to SLIC that does not need use $K$-means iterations and has to calculate less pixel distances, making it less memory consuming. The algorithm starts by sampling the grid with the initial N barycenter locations,
with a minimum distance associated to each.
A priority queue then feeds the superpixels with the image pixels. The pixel will be associated with the closest superpixel, the barycenter is updated, and its neighbors that are not yet associated to any region are pushed to the queue. These steps are repeated until the queue is empty. The spherical version SSNIC \cite{da2021fast} extends this algorithm for the spherical image domain
following the same adaptation for the SphSLIC method \cite{zhao2018}.
The hierarchical extension proposed in \cite{da2021fast} is a spherical version of the SH algorithm \cite{wei2018}. This algorithm generates regions in the image at different levels, hierarchically grouping pixels by color to form the superpixels. Contrary to SLIC and SNIC, SH does not produce regularly shaped superpixels. Similarly to the approach in \cite{yang2016efficient}, the algorithm considers the image as an undirected graph $G=\{V,E\}$ where $V$ is the set of pixels and $E$ the set of edges that connects neighbor pixels. The algorithm treats each vertex as a tree and the trees are iteratively aggregated to its neighbor according to the minimization of a cost function. For the spherical domain extension SSH, the set of edges $E$ in the graph is modified to reflect the connectivity of pixels in the spherical domain and a new cost function is proposed to aggregate the spherical features.

These methods use the same features as SLIC,
\emph{i.e.},
simple color and spatial distance between the pixel and the average superpixel features.
Therefore, as SLIC, they are very sensitive to highly textured areas and presence of noise in the images, which can lead to poor segmentation results.
These approaches also do not integrate any contour information,
reducing the object segmentation accuracy and
leading their algorithm to ignore thin contours.
These limitations due to the simple SLIC clustering framework are tackled in \cite{giraud2018_scalp} for standard 2D planar image segmentation,
where the notion of shortest path between the pixel and the superpixel is introduced to extract more relevant features providing higher segmentation accuracy.

}

\section{Spherical Shortest Path-based Superpixels}
\label{sec:sps}

{\color{review2}
In this section, we present the proposed SphSPS method.
First, we describe in details the frameworks used by SphSPS, \emph{i.e.},
the initial {$K$-means} clustering algorithm of SLIC \cite{achanta2012} (Section \ref{subsec:slic})
and its adaptation to spherical images introduced in \cite{zhao2018}
(Section \ref{subsec:spherical_geometry}).
Then, we present the core contributions,
which are generalizing the linear shortest path approach of \cite{giraud2018_scalp}  (Section \ref{subsubsec:planar_path}) to equirectangular images
in order to extract relevant features on a spherical shortest path (Sections \ref{subsubsec:gen_path} and \ref{subsec:sps_path}),
and proposing a fast sampling method to extract features on such path in a reduced computational time (Sections \ref{subsubsec:fast_path}).
}

\subsection{Planar $K$-means Iterative Clustering\label{subsec:slic}}

The proposed SphSPS method is based on the Simple Linear Iterative Clustering (SLIC) \cite{achanta2012}.
The algorithm is particularly interesting since it only requires as input parameters the approximate number of superpixels to produce and a shape regularity constant,
and it may achieve accurate performance in a few iterations.
The iterative $K$-means clustering is spatially constrained
through the whole process to respect a spatial regularity in the segmentation.
First, superpixels, denoted $S_i$ in the following, are initialized over the image domain
as square blocks of size $s{\times}s$.
Each superpixel is
described by the average CIELab colors $C_{S_i}$ %(CIELab colors for \cite{achanta2012})
and spatial barycenter position $X_{S_i}=[x_i,y_i]$ of all pixels within $S_i$.
At each iteration,
each superpixel $S_i$ is considered and compared to all pixels $p=[C_p,X_p]$,
of color $C_p$ having their position $X_p$ into
a $(2s$$+$$1)$${\times}$$(2s$$+$$1)$ square window $A_i$ around its barycenter $X_{S_i}$.
At the end of each iteration,
a pixel $p$ is associated to the superpixel minimizing the distance $D$ composed of a color $d_c$ and a spatial distance $d_s$ such as: %\vspace{-0.225cm}
{\eqsize
\begin{align}
d_c(p,{S_i})&={\|C_p-C_{S_i}\|}_2^2 ,  \label{dc}\\
 d_s(p,{S_i})&={\|X_p-X_{S_i}\|}_2^2 , \label{ds} \\
 D(p,S_i)&=d_c(p,{S_i}) + d_s(p,{S_i})\frac{m^2}{s^2}  , \label{slic}
\end{align}
}%
where $m$ is the trade-off parameter that enables to set the shape regularity.
Finally, a post-processing step is applied to
ensure the region connectivity.

\subsection{Adaptation to Spherical Images\label{subsec:spherical_geometry}}

\subsubsection{Spherical Geometry\label{subsubsec:spherical_geometry}}
\begin{figure}[t]
\centering
\begin{tabular}{cc}
\includegraphics[width=0.675\textwidth]{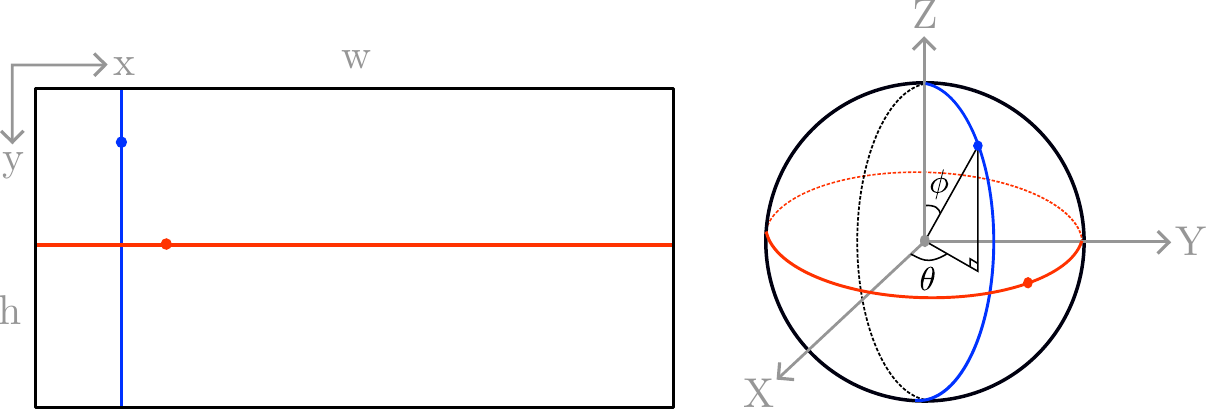}
\end{tabular}
\caption{2D Equirectangular to 3D spherical space projection system.
Vertical and horizontal planar coordinates correspond to the meridians and circles of latitude of the sphere.
The azimuthal angle $\theta$ and polar angle $\phi$ are considered in the projection system \eqref{xy_xyz}
}%
\label{fig:sphere_projection}
\end{figure}

The correspondence between the planar equirectangular 2D space and
the 3D spherical space,
can be seen as the respective projection of
vertical and horizontal planar coordinates on the meridians and circles of latitude of the sphere,
resulting in a spherical image having a width $w$ twice superior to its height $h$.
The projection system is illustrated in Figure \ref{fig:sphere_projection}.
Each image pixel $X=[x,y]$ in the 2D planar space,
corresponds to a 3D position $X^a=[x^a,y^a,z^a]$ in the spherical acquisition space
such as: %\vspace{-0.15cm}
{\eqsize
\begin{equation}
%\begin{array}{l}
  X=
%\left\{
%\begin{array}{ll}
\begin{bmatrix}
x=\hspace{-0.00cm} \floor{\frac{\theta w}{2\pi}} \\[-0.5ex]
\\[-0.25ex]
y=\hspace{-0.00cm} \floor{\frac{\phi h}{\pi}} \\
\end{bmatrix}
% \end{array}
% \right.%
% \end{array}
\leftrightarrow %\hspace{0.1cm}
X^a=
%\left\{
%\begin{array}{ll}
\begin{bmatrix}
x^a = \text{sin}(\frac{y\pi}{h})\text{cos}(\frac{2x\pi}{w}) \\[0.25ex]
y^a = \text{sin}(\frac{y\pi}{h})\text{sin}(\frac{2x\pi}{w}) \\[0.25ex]
z^a = \text{cos}(\frac{y\pi}{h})
\end{bmatrix}
% \end{array}
% \right.
.   \label{xy_xyz}
\end{equation}
}%

With $\theta=\text{arctan2}({y^a,x^a})$ the azimuthal angle, and $\phi=\text{arccos}(z^a)$ the polar angle.
Note that when going from 3D to 2D domain,
$x\in [-\frac{w}{2},\frac{w}{2}]$.
Hence, to map $x$ to the image domain, we compute
$x \leftarrow x+w$, if $x\leq0$.

\subsubsection{Spherical $K$-means Iterative Clustering\label{subsubsec:sphsps}}

SphSPS adapts the planar $K$-means method to the spherical geometry in three steps, in the same manner as \cite{zhao2018}.
The first one is the initialization of the $K$ superpixels,
to uniformly sample the 3D sphere.
In Figure \ref{fig:init}, we compare several planar and spherical seed initialization strategies for a number $K=400$ superpixels.
With planar-based sampling (Figures \ref{fig:init}(a) and (b)), the seeds are regularly spaced in the two dimensions in the equirectangular image but only along the circles of latitude in the spherical space, so many seeds are set close to the poles.
A random seed initialization (Figure \ref{fig:init}(c)) may cover the sphere volume but may lead to irregularities, especially with a low number of superpixels.
A geodesic approach, relying on subdivisions of icosahedron,
can be used to sample the 3D sphere very regularly (Figure \ref{fig:init}(d)) but
the number of seeds is limited to $K\in \{5{\times}2^{2n+1}+2, \text{ with } n\in\mathbb{N}\}$.
Finally, two other approaches are compared to uniformly sample the 3D sphere
using the Fibonacci sequence \cite{swinbank2006fibonacci} (Figure \ref{fig:init}(e)) or the Hammersley sampling \cite{wong1997sampling} (Figure \ref{fig:init}(f)).
Both generate in real-time a sufficiently regular spherical sampling with the desired number of superpixels.
In Section \ref{subsec:param_res}, we compare the use of each initialization strategy on the segmentation performance.
According to these results, we use in SphSPS the Hammersley sampling \cite{wong1997sampling}, as in \cite{zhao2018}.

{\color{review}
The whole $K$-means clustering process being deterministic with the same parameters, the generated result can only be influenced by the nature of the initialization.
All sampling methods of initial superpixels barycenters are deterministic for a given number $K$, except the random one (Figure \ref{fig:init}(c)), which may lead to slightly different segmentation results.
%
% Hence,
By using a Hammersley sampling, SphSPS
always generates the same segmentation result for the same parameter settings.}

\newcommand{\whsev}{0.1\textwidth}
\newcommand{\wsev}{0.2\textwidth}
\newcommand{\hsev}{0.1\textwidth}
\begin{figure*}[t]
{\footnotesize
\begin{tabular}{@{\hspace{0mm}}c@{\hspace{1mm}}c@{\hspace{3mm}}c@{\hspace{1mm}}c@{\hspace{3mm}}c@{\hspace{1mm}}c@{\hspace{0mm}}}
\includegraphics[width=\wsev,height=\hsev]{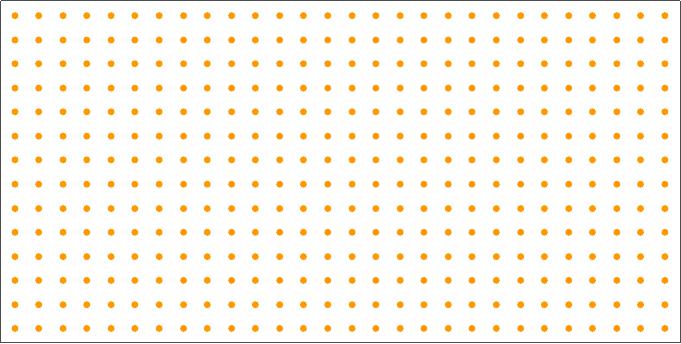}&
\includegraphics[width=\whsev,height=\whsev]{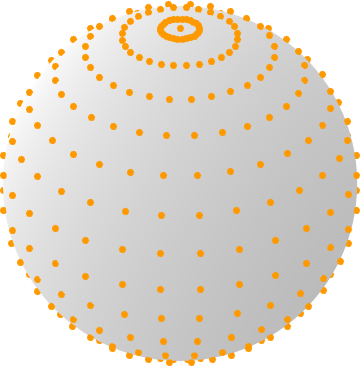}&
\includegraphics[width=\wsev,height=\hsev]{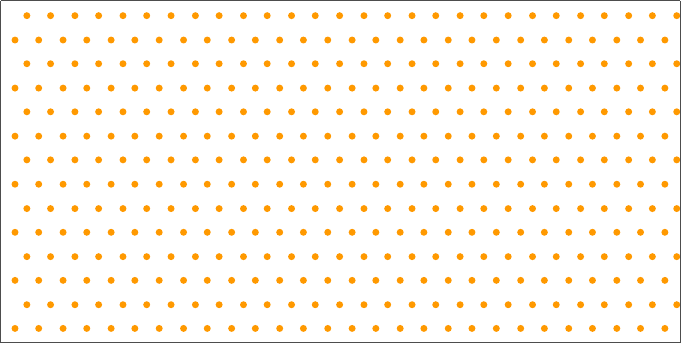}&
\includegraphics[width=\whsev,height=\whsev]{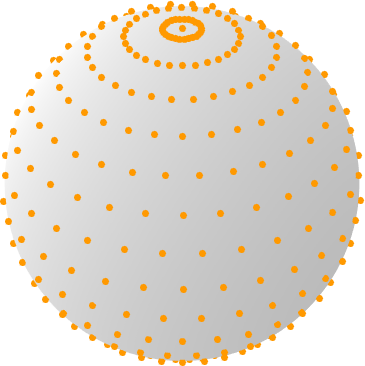}&
\includegraphics[width=\wsev,height=\hsev]{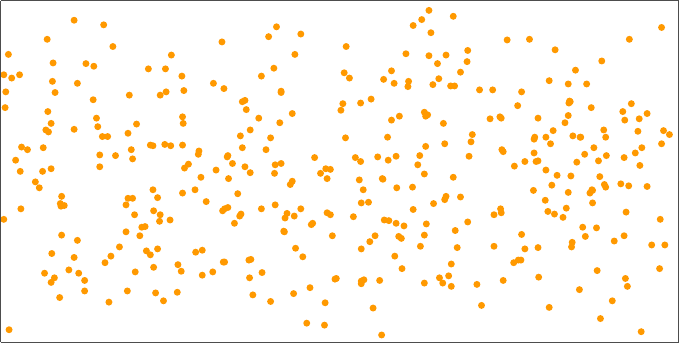}&
\includegraphics[width=\whsev,height=\whsev]{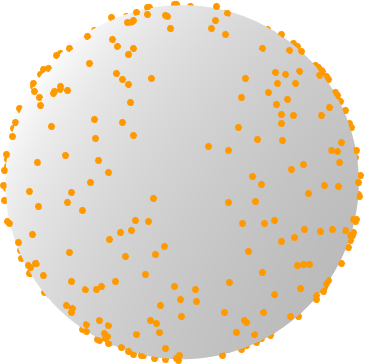}\\
 \multicolumn{2}{c}{(a) Square grid} & \multicolumn{2}{c}{(b) Hexagonal grid} & \multicolumn{2}{c}{(c) Random} \\[2ex]
\includegraphics[width=\wsev,height=\hsev]{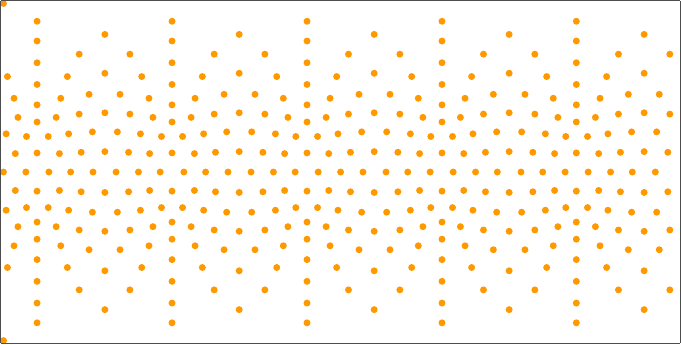}&
\includegraphics[width=\whsev,height=\whsev]{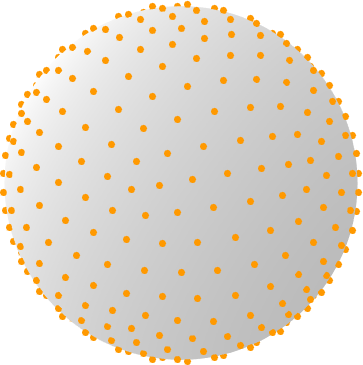}&
\includegraphics[width=\wsev,height=\hsev]{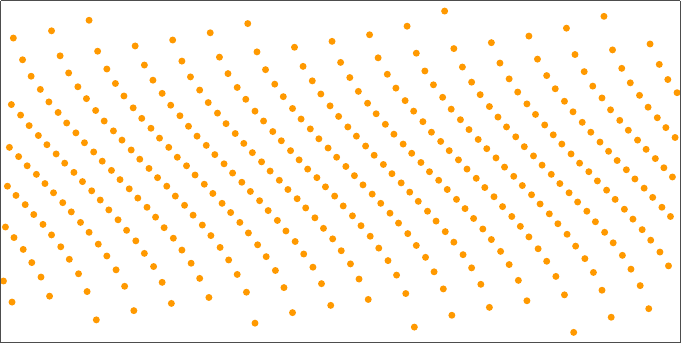}&
\includegraphics[width=\whsev,height=\whsev]{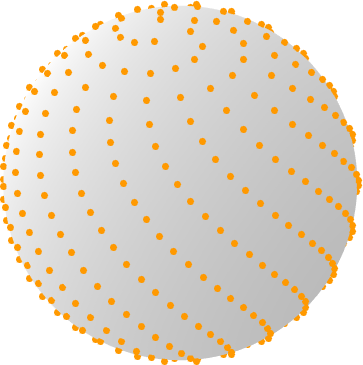}&
\includegraphics[width=\wsev,height=\hsev]{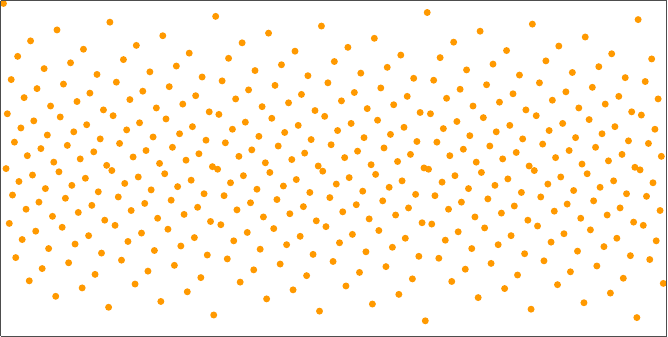}&
\includegraphics[width=\whsev,height=\whsev]{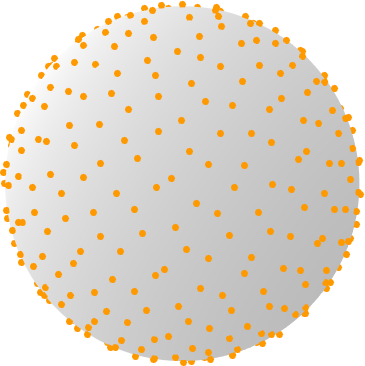}\\
 \multicolumn{2}{c}{(d) Geodesic} & \multicolumn{2}{c}{(e) Fibonacci} & \multicolumn{2}{c}{(f) Hammersley}
\end{tabular}
}
\caption{Comparison of superpixel seeds initialization strategies in the planar and spherical space}
\label{fig:init}
\end{figure*}

\begin{figure}[t]
\centering
\begin{tabular}{cc}
\includegraphics[width=0.45\textwidth]{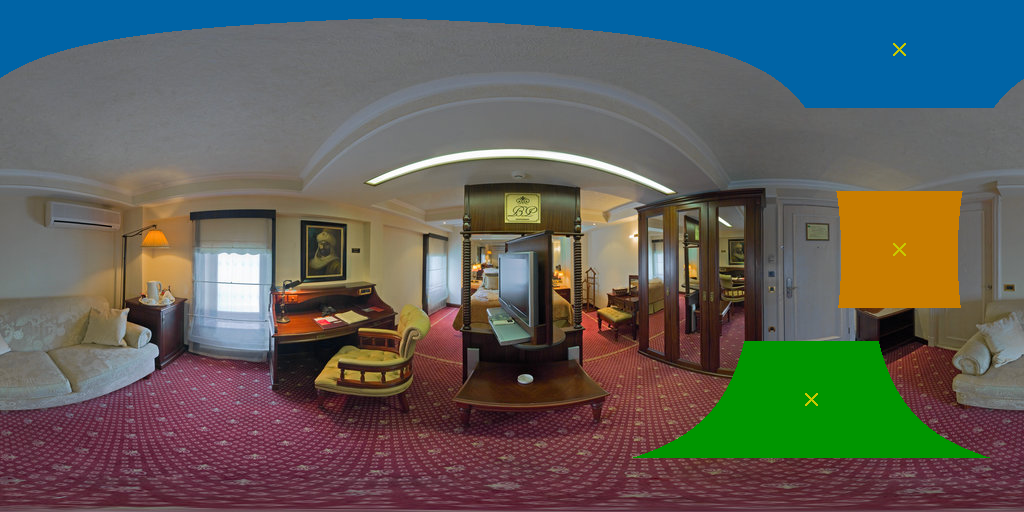}&
\includegraphics[width=0.225\textwidth]{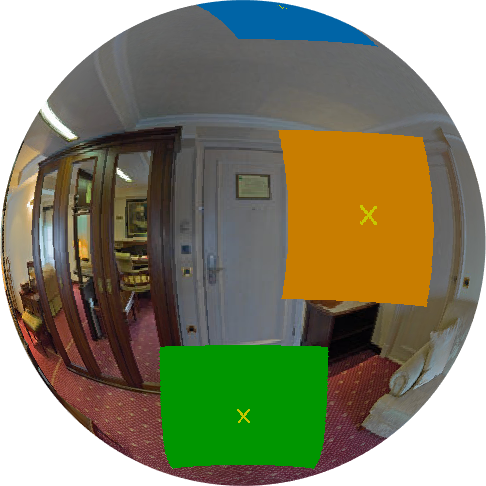}\\
\end{tabular}
\caption{Examples of spherical search areas for different superpixel barycenter positions with $K=400$ superpixels
}%
\label{fig:search_area}
\end{figure}

The second step is the adaptation of the
definition of the search area, which must consider the position of pixels in the spherical space.
For instance, superpixels near the sphere poles must have larger search areas in the equirectangular image.
The overall number of superpixels must also be considered to respect the regularity of the segmentation.
Hence, the search area $A_i$ for each superpixel $S_i$ of barycenter $X_{S_i}=[x_i,y_i]$, is defined as: %\vspace{-0.2cm}
{\eqsize
\begin{align}
\hspace{-0.05cm} A_i = \hspace{-0.05cm}\{[x,y] | x_i\hspace{-0.05cm}-\hspace{-0.05cm}\frac{S}{\text{sin}(\phi)}\leq \hspace{-0.05cm}x\hspace{-0.05cm}\leq x_i+\frac{S}{\text{sin}(\phi)} , y_i\hspace{-0.05cm}-\hspace{-0.05cm}S \leq y \leq y_i\hspace{-0.05cm}+\hspace{-0.05cm}S\},
\end{align}
}%
\noindent where
$S=w/\sqrt{K\pi}$ is the average superpixel size and
$\phi=y\pi/h$ is the polar angle corresponding to the $y$-th row for an image
of height $h$ and width $w$.
Since we deal with equirectangular images, the 360$^\text{o}$ aspect must also be handled to connect the pixels at the horizontal boundaries.
This is simply done using a horizontal warping of positions,
when the search region falls outside the image boundaries \cite{zhao2018}.
The spherical search areas are represented for several superpixel barycenter positions in Figure \ref{fig:search_area}.

The third modification to implement in the planar algorithm
is the computation of the spatial distance $d_s$ \eqref{slic}
in the spherical space.
To compute a spatial distance in this 3D space, \emph{i.e.} using
3D positions $X^a$ \eqref{xy_xyz},
we can use the Euclidean one, defined such as:
$d_s(X_p^a,X_{S_i}^a)=\|X^a_p - X^a_{S_i}\|_2^2$.
Nevertheless, as proposed in \cite{zhao2018},
we choose to use in SphSPS a spherical and computationally costless cosine dissimilarity distance defined as:
{\eqsize
\begin{equation}
d_s(X_p^a,X_{S_i}^a)=1-\left< X_p^a, X_{S_i}^a\right>,  \label{ds_sph}
\end{equation}
}%
with $\left< X_p^a, X_{S_i}^a\right>$, the scalar product between
the two position vectors.
Note that, although it does not exactly follow the sphere geometry,
the Euclidean and Cosine dissimilarity distances can achieve almost similar
performance for \cite{zhao2018}
with adjusted parameter $m$ \eqref{slic} (see Section \ref{sec:results}).

\subsection{Generalized Shortest Path Method}

\subsubsection{Feature extraction on a shortest path\label{subsubsec:planar_path}}

In \cite{giraud2018_scalp}, when processing a pixel $p$,
the color and contour information
of the pixels $q$ on
the planar shortest path $\mathbf{P}_{p,S_i}$
% $p$
towards the barycenter of the superpixel $S_i$
are considered in the clustering distance.
By using these additional features,
the clustering accuracy and the respect of object contours are improved.
Also, contrary to color-based geodesic paths used for instance in \cite{wang2013,rubio2016}, the planar shortest path increases the shape regularity
which is a desirable property \cite{giraud2017_jei},
by enforcing homogeneous star-convex shapes with respect to their barycenters \cite{gulshan2010geodesic}.
SphSPS integrates these features using the same clustering distance $D$ as \cite{giraud2018_scalp}, but with a different shortest path  definition since it is defined according to the spherical space.

{\color{review}
First, the color information along the path is compared to the average one of the superpixel to ensure the color homogeneity.
This way, the regularity is increased in a relevant manner,
preventing non convex shapes to appear \cite{giraud2018_scalp} since the color information must be consistent from any pixel to the superpixel barycenter.
}
The color distance of the pixels $q$ on the path $\mathbf{P}_{p,S_i}$  %to the superpixel
is added to the color distance $d_c$ \eqref{dc} such that: %\vspace{-0.2cm}
{\eqsize
\begin{equation}
d_c(p,S_i,\PP_{p,S_i})\hspace{-0.05cm}=\hspace{-0.05cm}\lambda d_c(p,S_i)\hspace{-0.1mm}+\hspace{-0.1mm}
\frac{1-\lambda}{|\PP_{p,S_i}|}\hspace{-0.1cm}\sum_{q\in \PP_{p,S_i}}\hspace{-0.1cm}d_c(q,S_i),   %\vspace{-0.05cm}
\label{path_color}
\end{equation}
}%
\noindent
{\color{review}where $\lambda$ is a trade-off parameter between the color of the considered pixel and the ones on the shortest path, which is usually set to $0.5$ to equally consider both terms.}

Along the shortest path,
a contour information can also be extracted to ensure
the respect of objects borders.
These borders may indeed not be in line with the ones obtained following the color homogeneity criteria.
Especially for thin contours, which are not yet captured in the clustering framework.
Any contour map, denoted $\mathcal{C}$, with normalized values between 0 and 1,
respectively indicating the absence or presence of contours,
can be computed and used to define a contour term $d_\mathcal{C}$ such that:
%\vspace{-0.15cm}
%
{\eqsize
\begin{equation}
d_{\mathcal{C}}(\PP_{p,S_i}) = 1 + \gamma \hspace{0.1cm} \underset{q\in \PP_{p,S_i}}{\text{max}}\hspace{0.05cm}\mathcal{C}(q) , \label{path_contour} %\vspace{-0.05cm}
\end{equation}
}%
\noindent
{\color{review}
where $\gamma \geq 0$ and empirically set to $10$, is the parameter penalizing the crossing of a contour.
With such term, the pixel $p$ will not be
associated to the superpixel $S_i$ when a high contour
intensity is found on the path, thus enforcing the respect of thin contours.
}

Finally, the clustering distance $D$ of the SphSPS method
ensuring more regular regions and respecting the image object contours,
is defined as:   %\vspace{-0.2cm}
{\eqsize
\begin{equation}
   D(p,S_i)=\left(d_c(p,S_i,\PP_{p,S_i}) + d_s(p,S_i)\frac{m^2}{s^2}\right)d_\mathcal{C}(\PP_{p,S_i})  , \label{newdist} %\vspace{-0.075cm}
\end{equation}}%
\noindent where $d_s$ is the spherical spatial distance using the cosine dissimilarity \eqref{ds_sph},
 and $\PP_{p,S_i}$  is the proposed spherical shortest path computed as follows.

\subsubsection{Generalized shortest path\label{subsubsec:gen_path}}

\begin{figure}[t]
\centering
\includegraphics[width=0.75\textwidth]{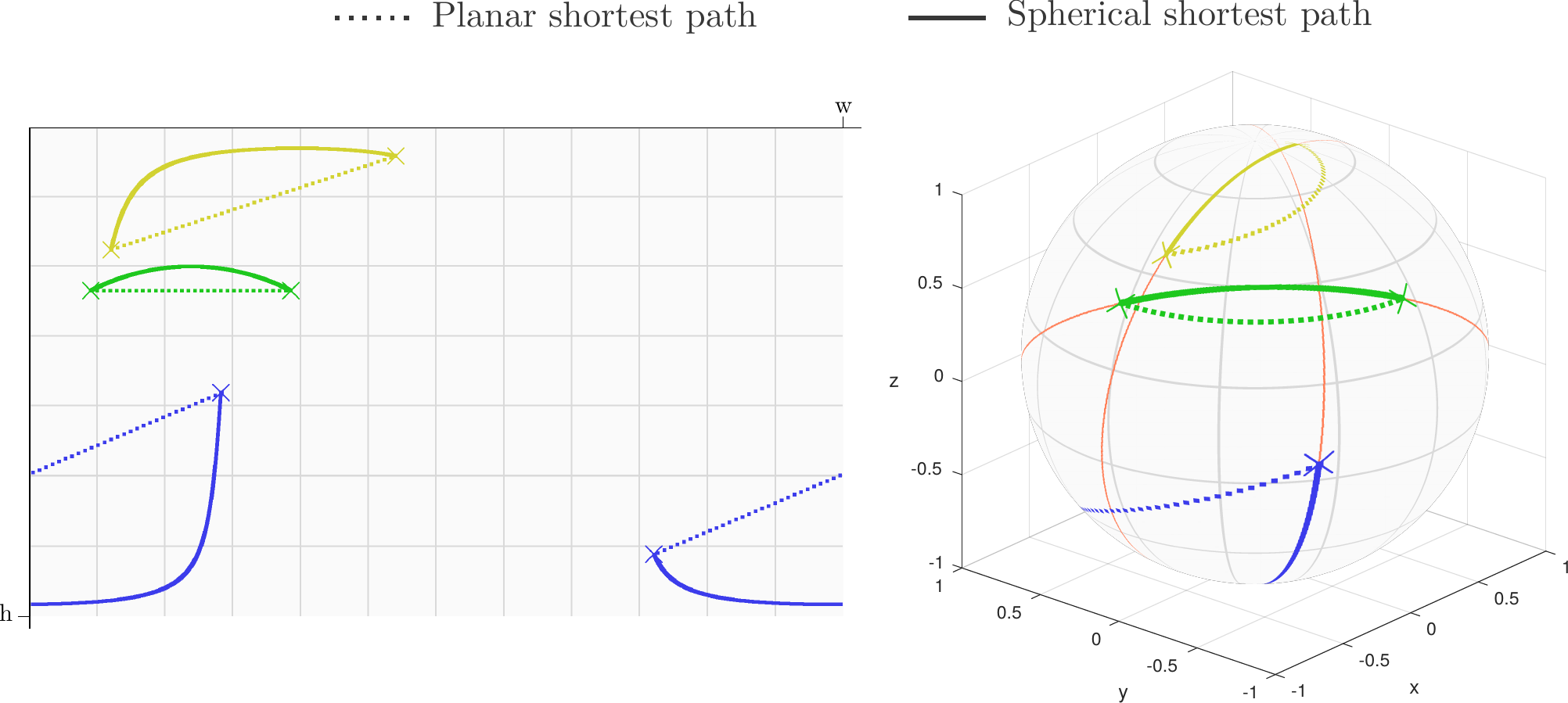}
\caption{Examples of planar (dotted lines) and spherical shortest paths (full lines) between points in the 2D planar image space (left) and 3D acquisition space (right).
The spherical path follows the shortest geodesic path on the sphere
}%
\label{fig:spherical_path}
\end{figure}

\begin{figure*}[t!]
\centering
{\footnotesize
\begin{tabular}{@{\hspace{0mm}}c@{\hspace{2mm}}c@{\hspace{2mm}}c@{\hspace{0mm}}}
\includegraphics[height=0.22\textwidth]{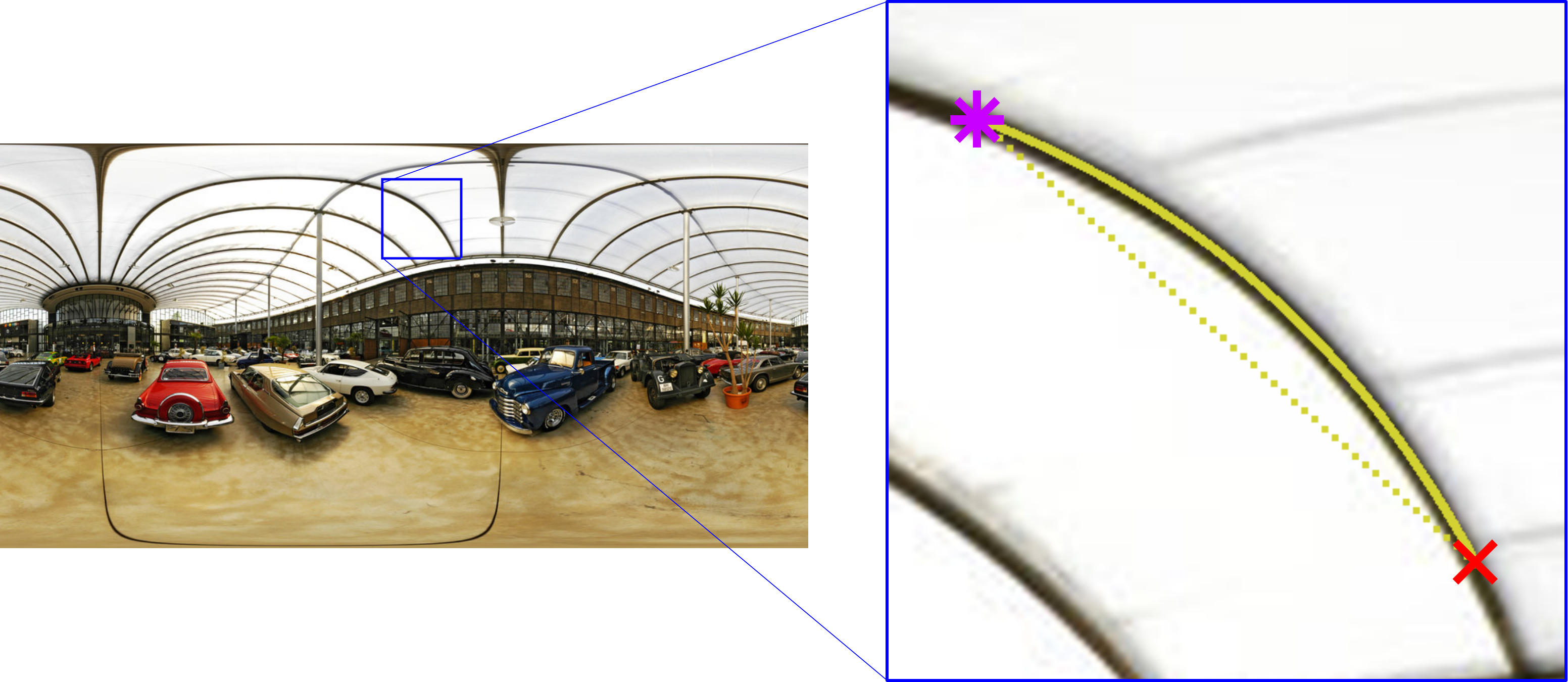}&
\includegraphics[height=0.22\textwidth]{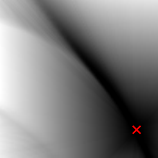}&
\includegraphics[height=0.22\textwidth]{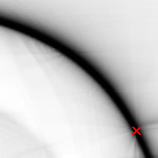}\\
(a) Equirectangular image & (b) Linear shortest  & (c) Spherical shortest \\
& path distance & path distance\\
\end{tabular}}
\caption{Comparison of linear shortest path (dotted line) and spherical shortest path (full line)
between an initial (red cross) and final (purple star) pixel position in (a).
In (b) and (c), the average color distance on all points of the path from the
the initial point (star) is represented (lighter color indicates a higher distance).
When trying to associate a point to a superpixel, such distance from the point to the superpixel barycenter is considered \eqref{path_color}.
Hence, using the spherical shortest path enables to follow the
structure of objects more accurately in the equirectangular image
}%
\label{fig:linear_vs_spherical}
\end{figure*}

{\color{review2}
In this section, we propose a general formulation of the shortest path
between a pixel and a superpixel.
}
For any image to segment, the shortest path should be computed in the
acquisition space, to respect the acquisition geometry.
In Figure \ref{fig:spherical_path},
we illustrate several examples of shortest paths in the planar space
as in \cite{giraud2018_scalp},
and in the spherical one as in SphSPS.
The difference of using these paths
to compute the color distance on the path to the superpixel barycenter \eqref{path_color} is also illustrated in Figure \ref{fig:linear_vs_spherical}.
For a reference point (red cross),
the average color distance to all the other points on
linear and spherical shortest paths are respectively represented in Figure \ref{fig:linear_vs_spherical}(b) and \ref{fig:linear_vs_spherical}(c).
The spherical path enables the algorithm to respect, within the equirectangular image, the linear structure of objects in the spherical space.

For the planar case, the acquisition and image spaces
 are equivalent.
Hence, the shortest path can be reduced to a linear path
that can be easily computed with a discrete algorithm such as \cite{bresenham1965}.
{\color{review}
On the other hand,
in the general case
with fully spherical or with different aperture angles, \emph{e.g.}, fisheyes,
distortions can be introduced between the
acquisition and the image space ($\mathbb{N}^2$).}
%
%computes it %
Therefore, the shortest path should be first explicitly computed in the acquisition space,
denoted as $\mathbf{P}_{p,S_i}^a$,
and projected back to the planar space
to provide the discrete sample positions in the 2D image.
Hence, the general formulation of the discrete shortest path is defined as:
{\eqsize
\begin{equation}
 \mathbf{P}_{p,S_i}  = \mathbf{P}_{p,S_i}^a \hspace{0.25cm} \overrightarrow{\text{\small proj}} \hspace{0.25cm} \{\mathbb{N}^2\} . \label{proj_path}
\end{equation}
}%

{\color{review}
With our general method, the projection equations \eqref{xy_xyz} and the way to extract the shortest path \eqref{sph_path} should be slightly modified according to the geometry.
Nevertheless, the core of the algorithm, \emph{i.e.}, the clustering distance \eqref{newdist}, with the projection of the shortest path on the 2D image \eqref{proj_path} remain identical and do not require different or additional parameter settings.
}

\subsubsection{Shortest path in the spherical space\label{subsec:sps_path}}

The shortest path in the spherical space consists in following the geodesic along the sphere \cite{gromov1983filling}.
This geodesic lies within the
\textit{great circle}
(in orange color in Figures \ref{fig:spherical_path} and \ref{fig:great_circle}),
containing the two considered points and
the center of the sphere to form a disc.
In the following, we rely on the Slerp (Spherical linear interpolation) formulation of the spherical geodesic path problem
proposed in \cite{shoemake1985animating} to sample the points of our shortest path.
Note that tangential approaches to extract way-points on the great circle
have been formalized, for instance in \cite{karney2013algorithms},
but such theoretical methods
use many trigonometric computations that may impact the processing time.
 To the best of our knowledge, no direct algorithm, such as \cite{bresenham1965} for the planar case,
 has been made explicitly available to extract in an equirectangular image
 the discrete shortest path between two pixels in the spherical space.

\subsubsection*{Spherical geodesic path implementation}

\begin{figure}[t!]
\centering
{\footnotesize
\includegraphics[width=0.75\textwidth]{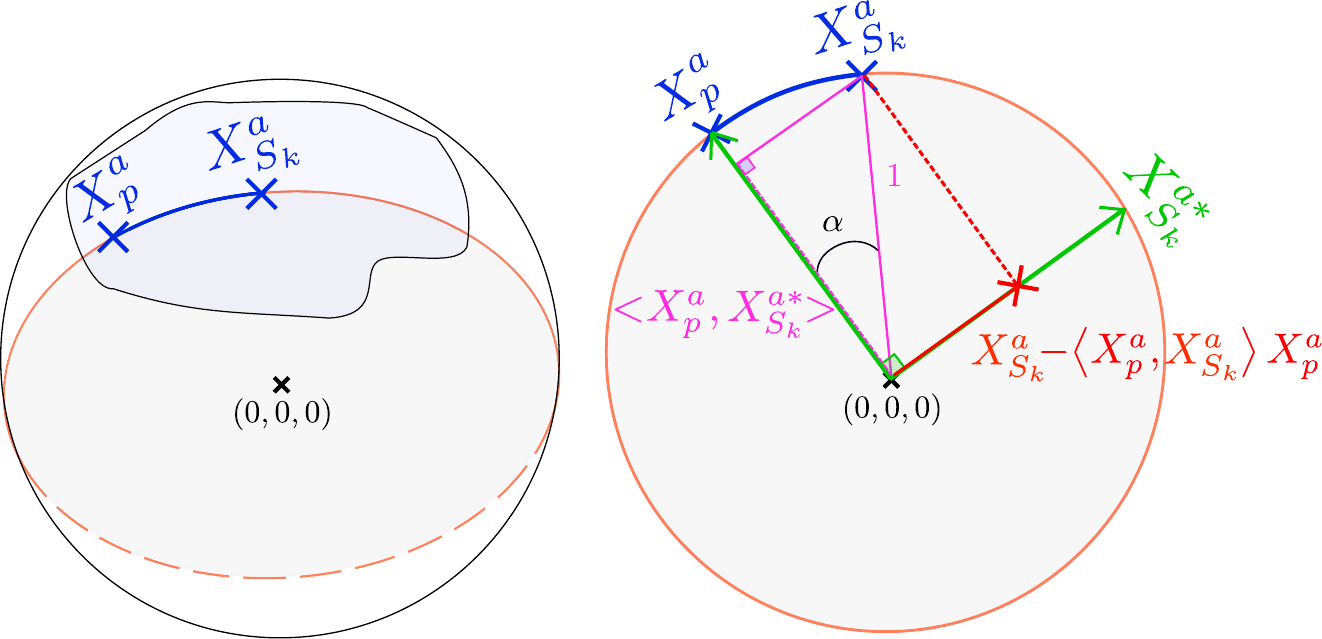}}%
\caption{Computation of the spherical shortest path.
The orthogonal coordinate system $[\vec{X_p^a},\vec{X_{S_i}^{a*}}]$ is computed from projection of  $X_{S_i}^a$ on $X_p^a$ \eqref{co_system}.
The  angle  $\alpha$ between the positions is then
used to sample 3D points on the path \eqref{sph_path}%
}%
\label{fig:great_circle}
\end{figure}

{\color{review2}
In this section, we propose a process to compute the spherical shortest path
between a pixel and a superpixel.
This process is illustrated in  Figure \ref{fig:great_circle}.
}
For each comparison of a pixel at position $X_p^a$
to a superpixel of spatial barycenter $X_{S_i}^a$,
we first compute
an orthogonal coordinate system $[\vec{X_p^a},\vec{X_{S_i}^{a*}}]$
that lies within their great circle.
To build such system, we use the Gram-Schmidt orthogonalization process \cite{bjorck1967solving}
to get the position ${X_{S_i}^{a*}}$,
an orthogonal vector to $X_p^a$
within the great circle such as:
%\vspace{-0.115cm}
%
{\eqsize
\begin{equation}
{X_{S_i}^a}^* = \frac{X_{S_i}^a - \left<X_p^a,X_{S_i}^a\right>X_p^a}{\left\|X_{S_i}^a - \left<X_p^a,X_{S_i}^a\right>X_p^a\right\|_2} .
\label{co_system}
\end{equation}
}%
\noindent
The cost of computing \eqref{co_system} is greatly reduced since the scalar product $\left<X_p^a,X_{S_i}^a\right>$ has already been computed for the spatial distance $d_s$ \eqref{ds_sph}.
Then, the angle $\alpha$ between the two points is simply obtained with
$\alpha = \text{arccos}\left(\left<X_p^a,X_{S_i}^a\right>\right)$.
Finally, $\alpha$ is used to linearly sample
the geodesic path $\mathbf{P}_{p,S_i}^a$,
starting from the pixel position and progressively
reaching the superpixel barycenter such as:
{\eqsize
    \begin{equation}
     \mathbf{P}_{p,S_i}^a = \text{cos}(\mathbf{\alpha_N})X^a_p + \text{sin}(\mathbf{\alpha_N})X^{a*}_{S_i} ,  \label{sph_path}
    \end{equation}
}%
with $\mathbf{\alpha_N}=\frac{\llbracket0, N-1\rrbracket}{N-1} \alpha\in\mathbb{R}^N$, intermediate angles to linearly sample $N$ points
within the coordinate system
$[\vec{X_p^a},\vec{X_{S_i}^{a*}}]$ between the pixel and barycenter positions.
The geodesic path in the acquisition space,
is finally projected to the planar space \eqref{xy_xyz} to get
the image pixel positions contained into the path $\mathbf{P}_{p,S_i}$ \eqref{proj_path}.
By this way, we compute the discrete shortest spherical path with simple calculations, dividing the processing time by a factor 2 compared to tangential approaches.

\subsubsection{Fast spherical shortest path sampling using recursive implementation\label{subsubsec:fast_path}}

Compared to the complexity of the SLIC method \cite{achanta2012},
a much  larger number of  pixels are considered in a shortest path,
leading to a significant increase of calculations.
Nevertheless, we can reduce the cost of this additional complexity
by using a recursive implementation.

First, as in \cite{giraud2018_scalp},
for each superpixel,
we store the color distance $d_c$ to each tested pixel,
making it directly available for the next pixels to process.
This first optimization reduces the processing time by $50\%$.
Here, we propose a general optimization framework using
the path redundancy with recursive implementation.
The principle is that if the path of a pixel to a superpixel crosses
a previously computed path to the same superpixel,
the rest of the path should be the same.
In the planar case using the algorithm of \cite{bresenham1965} this assumption is not exactly accurate,
but it becomes true in the spherical case
since all points lie on the same great circle.
Using recursive implementation, we can store
the average color distance and maximum contour intensity
on the path for each crossed pixel.
By this way, within a superpixel search area,
for most pixels, the large quantity of information contained in the shortest path
has already been computed once and is directly available.
This recursive scheme
also reduces the processing time by a factor 2.

\section{Generalized Global Regularity Measure}
\label{sec:grs}

Most superpixel methods try to decompose the image into regions
having approximately the same size, and
being homogeneous in terms of color, so they should naturally be contained into the image objects.
This is usually done by optimizing a trade-off between color and spatial terms.
Giving more weight to color would generally provide more accurate superpixels
in terms of respect of object contours, while increasing regularity would
prevent superpixel borders from following them.
Works such as \cite{giraud2017_jei} have demonstrated that having regular superpixels
may increase the performances of superpixel-based pipelines.
Hence, algorithms should try to produce both accurate and regular superpixels
and dedicated metrics should jointly evaluate
object segmentation and shape regularity performances.

Nevertheless,
this last aspect is rarely evaluated in the literature.
The standard compactness metric \cite{schick2012}
is the only one extended to the spherical space \cite{zhao2018},
although it was proven very limited \cite{giraud2017_jei}.
In this section, we propose a
more relevant metric to evaluate the shape regularity in any acquisition space.

\subsection{Limitation of the Compactness Measure}

The first reference regularity measure in the superpixel literature
is the compactness COM \cite{schick2012} that evaluates the
average circularity of each superpixel shape.
In \cite{zhao2018}, the extension of this metric to the spherical case is proposed.
The COM measure independently evaluates each superpixel $S_i$
 of a segmentation $\SSS=\{S_i\}$
 such as:
{\eqsize
\begin{equation}
\text{COM} = \frac{1}{\sum\limits_{S_i\in \mathcal{S}} |S_i|}\sum\limits_{S_i\in \mathcal{S}} Q_\text{sph}(S_i)|S_i| ,  \vspace{-0.075cm}
\label{circu}
\end{equation}
}%
with $Q_\text{sph}$ the spherical isoperimetric quotient \cite{osserman1978isoperimetric} defined as:
{\eqsize
\begin{equation}
Q_\text{sph}(S_i)= \frac{4\pi |S_i| - |S_i|^2}{|P(S_i)|^2},
\end{equation}
}%
 with $P(S_i)$ the superpixel perimeter.

Although this metric has been regularly used when evaluating the regularity,
it has been proven very limited \cite{giraud2017_jei}.
It relies on a reduced circularity criteria
such that ellipses can obtain higher regularity measures than squares,
and it is highly sensitive to boundary noise and inconsistent with the superpixel size.
The same limitations appear for the extension to the spherical case, to such extent that
the COM measure may even fail to differentiate spherical and planar-based methods \cite{zhao2018}.

\subsection{Generalized Global Regularity Metric}

\subsubsection{Global regularity metric}

To address the issues of the COM measure, a global regularity metric (GR) has been introduced in \cite{giraud2017_jei},
containing two terms.
First, instead of comparing the superpixels to a circle,
it robustly evaluates the convexity, the contour smoothness, and the 2D balanced repartition of each superpixel with the
Shape Regularity Criteria (SRC) term.
The convexity and smoothness properties are computed with respect to the discrete convex hull corresponding to the superpixel shape:
%
% as follows:
%
 {\eqsize
 \begin{equation}
 \text{SRC}(S_i) = \frac{\text{CC}(H_{S_i})}{\text{CC}(S_i)}\text{V}_{\text{xy}}(S_i) ,  \label{src}
 \end{equation}}%
 \noindent  where
 $\text{V}_\text{xy}(S_i) = {\min(\sigma_x,\sigma_y)/\max(\sigma_x,\sigma_y)}$,
 evaluates the balanced repartition of the shape
 with $\sigma_x$ and $\sigma_y$
 the square root of standard deviations of pixel positions $x$ and $y$ in $S_i$,
  $H_{S_i}$ is the convex hull containing $S_i$, and
 CC measures the ratio between the perimeter and the area of the shape.

As for the compactness COM \eqref{circu},
SRC is independently computed for each superpixel.
Hence, \cite{giraud2017_jei} introduces a Smooth Matching Factor (SMF) term
to also evaluate the consistency of all superpixel shapes.
First, each superpixel $S_i$ is registered on its barycenter to obtain $S_i^*$.
Then, these registered superpixel shapes   are averaged to compute $S^*$,
the average superpixel shape, created from
 the superposition of all superpixels.
 Finally, each registered superpixel $S_i^*$ is compared to the average shape $S^*$ such that:
% %\vspace{-0.35cm}
% %
%
 {\eqsize
\begin{equation}
 \text{SMF}(S_i) =  1 - \left\|\frac{S^*}{|S^*|} - \frac{S_i^*}{|S_i^*|}\right\|_1 \bigg/{2} . \label{smf}
 \end{equation}}%

Finally, the global regularity is defined with GR
combining these two metrics such that: %\vspace{-0.1cm}
{\eqsize
\begin{equation}
\text{GR}(\SSS) = \frac{1}{\sum\limits_{S_i\in \mathcal{S}} |S_i|}
\sum\limits_{S_i\in \mathcal{S}} |S_i|\text{SRC}(S_i)\text{SMF}(S_i) . \label{gr}
\end{equation}
}%
% \vspace{-0.175cm}

\subsubsection{Generalization in the acquisition space}

Ideally, the regularity of each discrete superpixel shape $S_i$,
should be evaluated in the acquisition space.
In our context,
we first deal with a set of 3D points $S_i^a$,
from the projections of the pixel positions of $S_i$
\eqref{xy_xyz}.
Since the GR metric relies on the computation of a convex hull for SRC \eqref{src},
and uses barycenter registration for SMF \eqref{smf},
it cannot be directly applied to such point clouds in $\mathbb{R}^3$.

\begin{figure*}[t]
\centering
\includegraphics[width=0.95\textwidth]{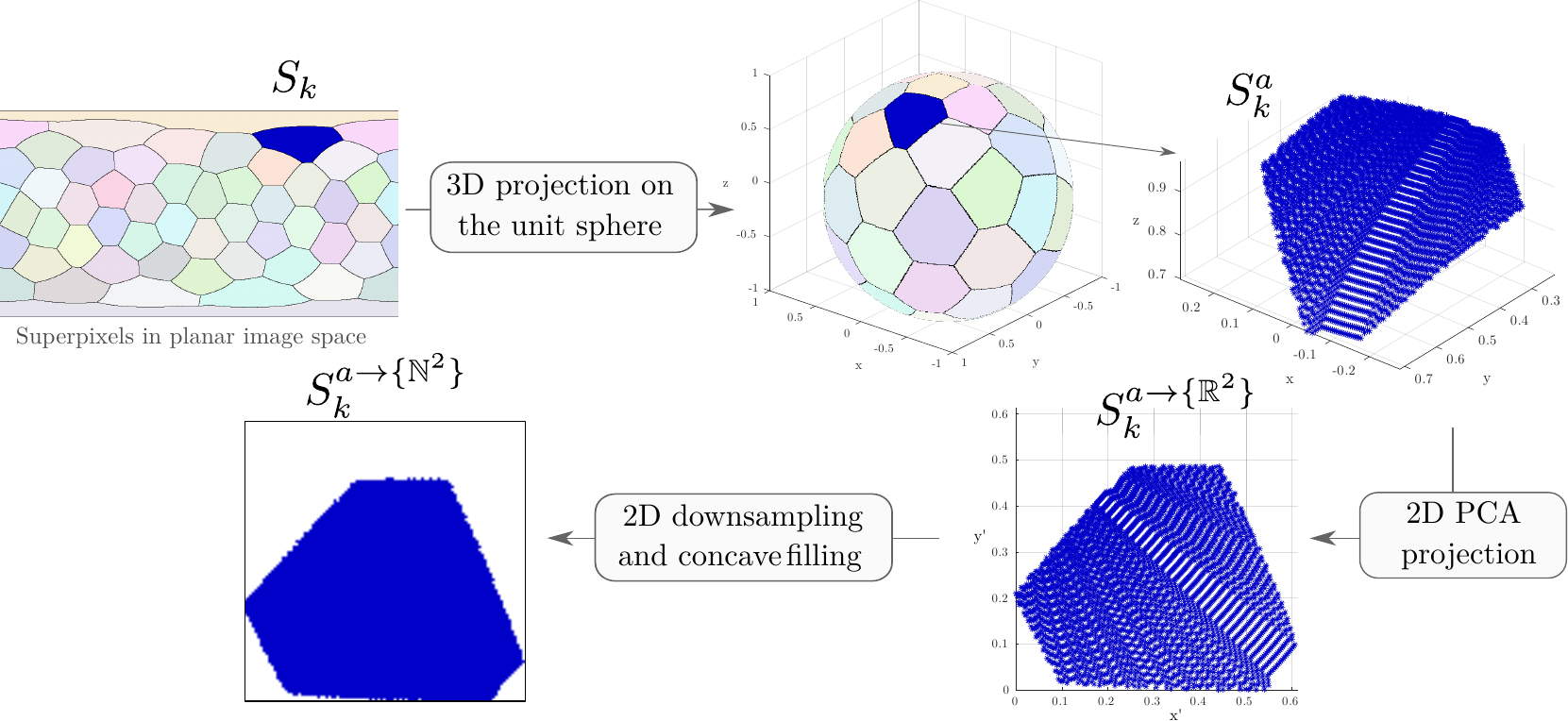}
\caption{Projection process for the proposed Generalized Global Regularity (G-GR) metric \eqref{grs}.
Each superpixel shape $S_i$ from the planar segmentation
is projected in its acquisition space to obtain $S_i^a$.
Then, a principal component analysis (PCA) is performed to project the 3D point cloud on a two dimensional
space using ($S_i^{a\rightarrow \{\mathbb{R}^2\}}$).
Finally, it is downsampled to generate a 2D matrix,
and the remaining potential holes are filled by considering the concave hull of the shape
($S_i^{a\rightarrow \{\mathbb{N}^2\}}$).
The G-GR metric especially relying on a convex hull can then be applied on each obtained dense 2D shape}%
\label{fig:grs}
\end{figure*}

To be able to use the GR metric,
we propose to reduce to 2D shapes relevantly approximating
each set of 3D points of $S_i^a$.
The whole projection process is illustrated in Figure \ref{fig:grs}.
Firstly, a superpixel $S_i$ in the discrete image space is projected
to its acquisition one, giving a spherical point cloud $S_i^a$.
Secondly,
to reduce to a 2D point cloud $S_i^{a\rightarrow \{\mathbb{R}^2\}}$,
we apply a principal component analysis (PCA) on $S_i^a$,
and project the points on its two most significant eigenvectors.
Finally,
to avoid holes in the projection and get a dense discrete shape,
a downsampling is performed to obtain a discrete 2D shape  $S_i^{a\rightarrow \{\mathbb{N}^2\}}$,
that is finally filled according to its concave hull.
Note that in \cite{giraud2020_icpr},
only a downsampling was used, with a higher scale,
resulting in a smaller and smoothed 2D shape,
thus reducing the relevance of the projection and the ability to measure the
the contour irregularities.

This way, each superpixel shape is relevantly projected from the 3D acquisition space into a discrete 2D shape.
The initial SRC and SMF measures can then be used in the
proposed Generalized Global Regularity (G-GR) metric as: %\vspace{-0.285cm}
{\eqsize
\begin{equation}
 \hspace{-0.05cm} \text{G-GR}(\SSS)  =  \frac{1}{\sum\limits_{S_i\in \mathcal{S}} \left|\sa\right|}\sum\limits_{S_i\in \mathcal{S}} \left|\sa\right|\text{SRC}(\sa)\text{SMF}(\sa).
 \label{grs}
\end{equation}
}%

Regularity results for SphSPS and state-of-the-art methods
are compared in Section \ref{subsec:soa}.
With the proposed G-GR metric, contrary to the COM measure,
a performance gap is now visible such that
no planar methods have higher regularity than spherical ones
for a given number of superpixels.

\section{Results}
\label{sec:results}

\subsection{Validation Framework}

\subsubsection{Dataset}

\begin{figure*}[t]
{\scriptsize
\begin{tabular}{@{\hspace{0mm}}c@{\hspace{1mm}}c@{\hspace{3mm}}c@{\hspace{1mm}}c@{\hspace{3mm}}c@{\hspace{1mm}}c@{\hspace{0mm}}}
% \multirow{2}{*}{\rotatebox{90}{\hspace{1cm} PSD \cite{wan2018} \hspace{-0.5cm}}}&
\includegraphics[width=\wsev,height=\hsev]{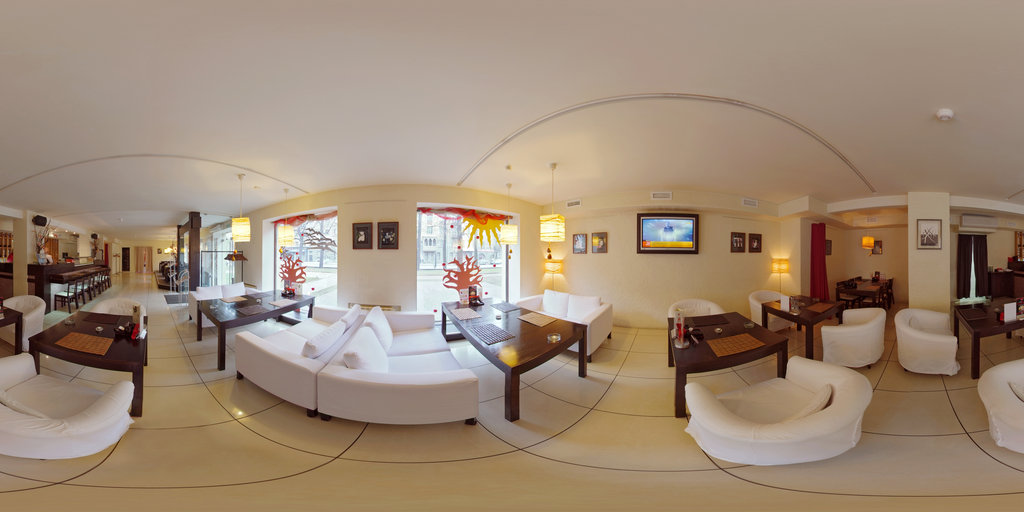}&
\includegraphics[width=\whsev,height=\whsev]{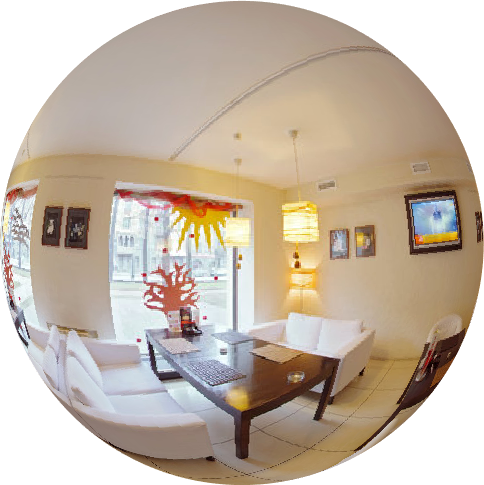}&
\includegraphics[width=\wsev,height=\hsev]{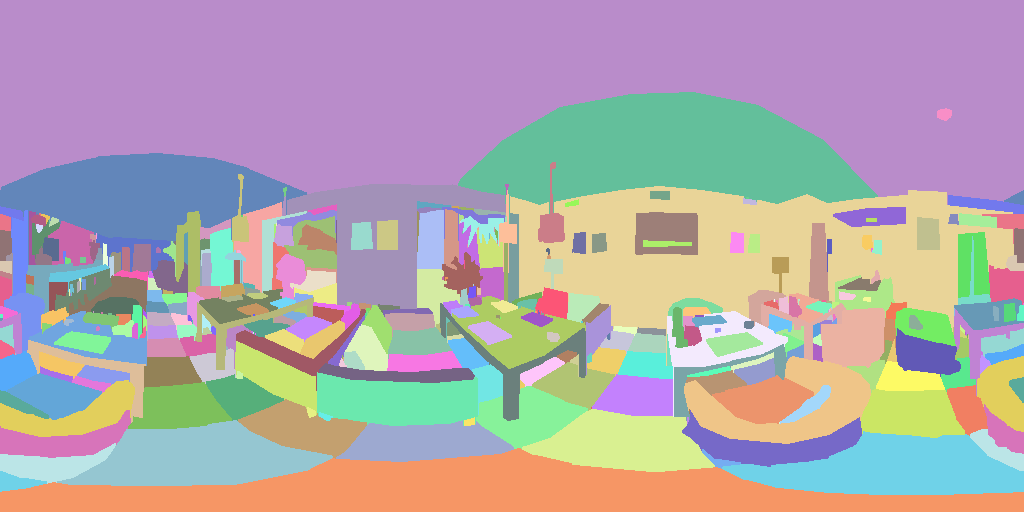}&
\includegraphics[width=\whsev,height=\whsev]{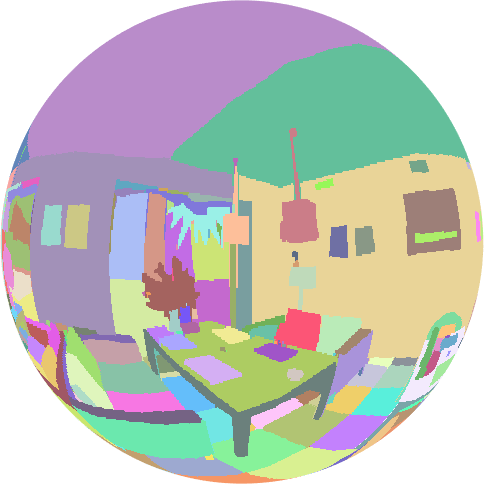}&
\includegraphics[width=\wsev,height=\hsev]{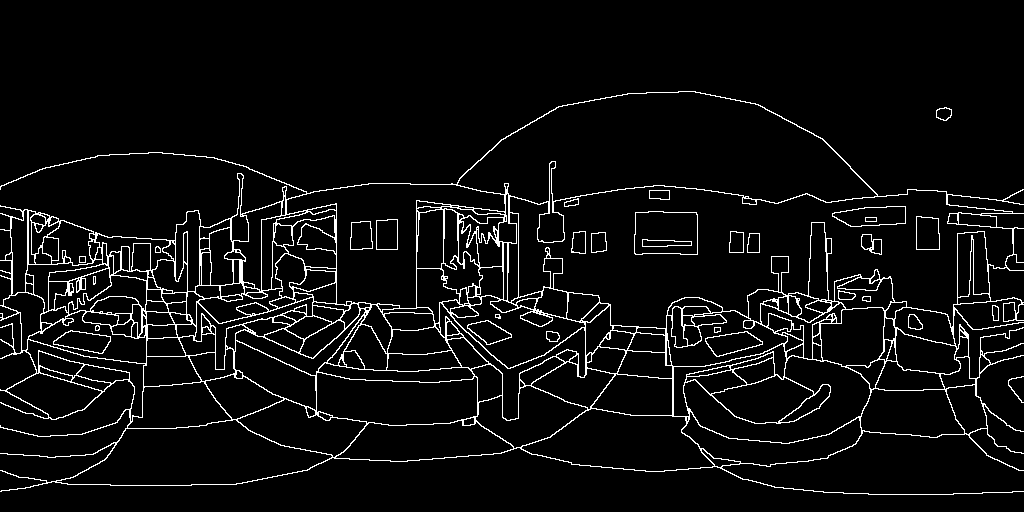}&
\includegraphics[width=\whsev,height=\whsev]{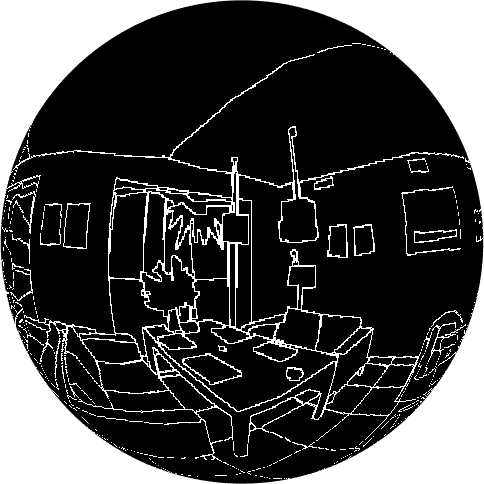}\\
\includegraphics[width=\wsev,height=\hsev]{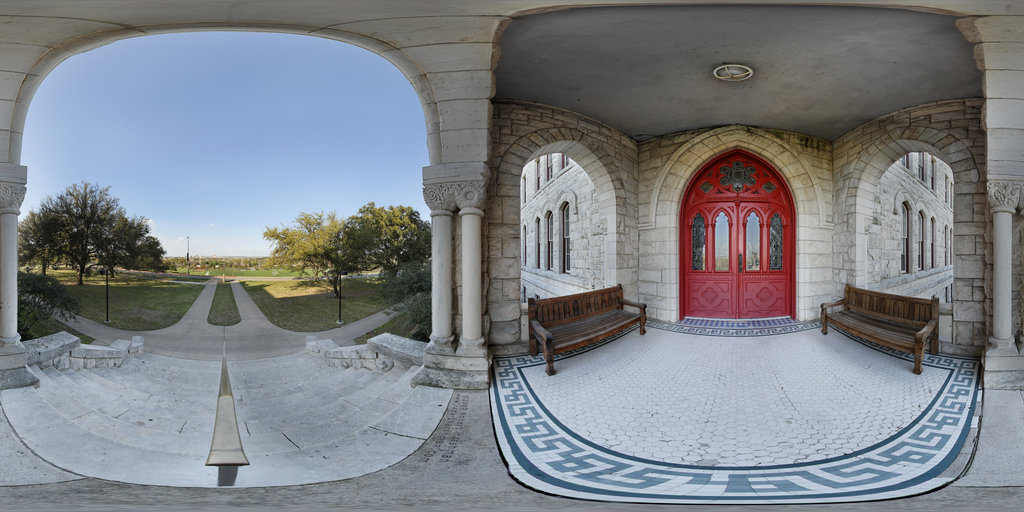}&
\includegraphics[width=\whsev,height=\whsev]{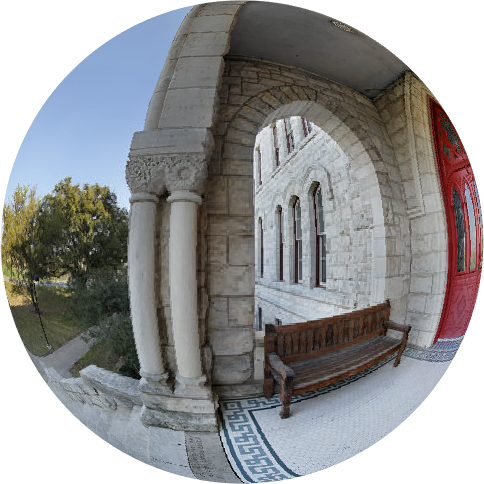}&
\includegraphics[width=\wsev,height=\hsev]{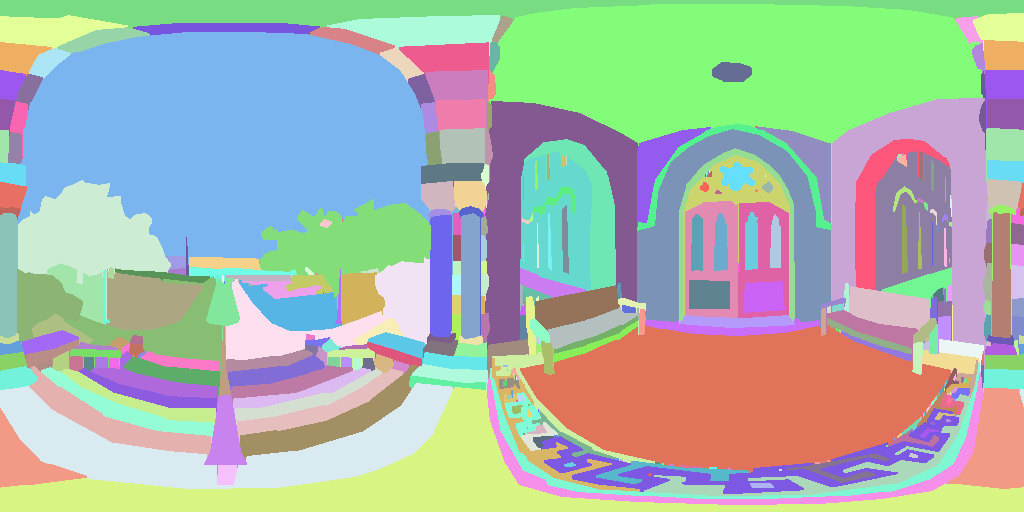}&
\includegraphics[width=\whsev,height=\whsev]{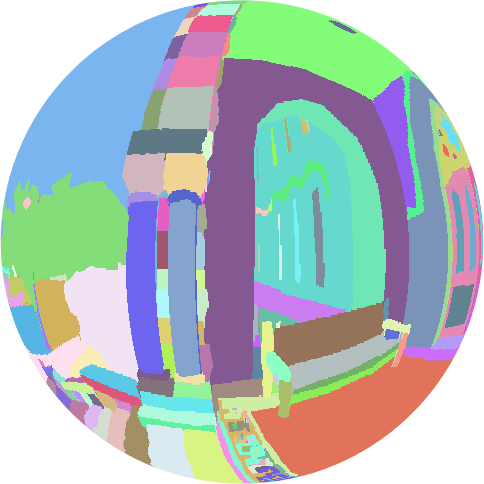}&
\includegraphics[width=\wsev,height=\hsev]{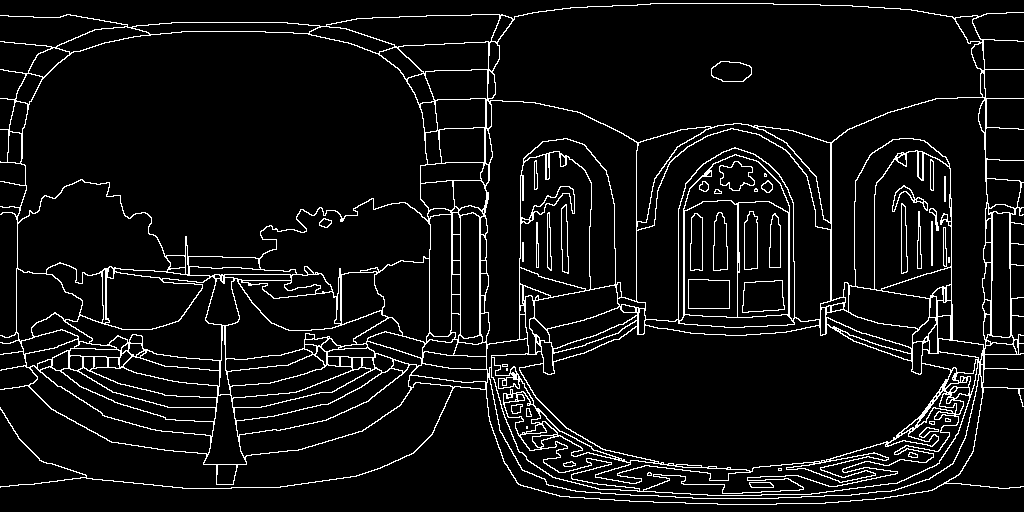}&
\includegraphics[width=\whsev,height=\whsev]{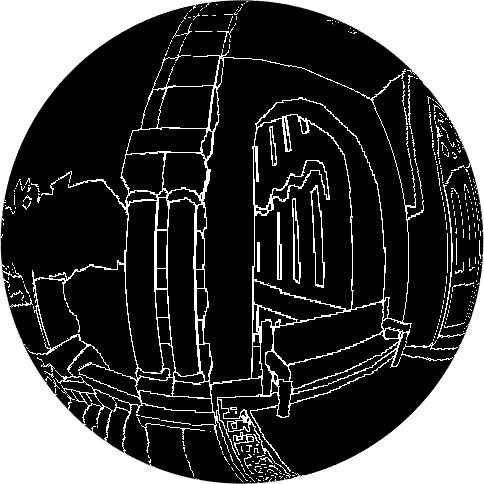}\\
\multicolumn{2}{c}{Image} & \multicolumn{2}{c}{Labels} & \multicolumn{2}{c}{Contours} \\
\end{tabular}
}
\caption{Examples of images from the Panorama Segmentation Dataset (PSD) \cite{wan2018}}
\label{fig:data_ex_sps}
\end{figure*}

To validate our method, we considered images from the
reference 360$^\text{o}$ equirectangular dataset SUN360 \cite{xiao2012recognizing}.
In \cite{wan2018}, 75 images of $512{\times}1024$ pixels from SUN360 are selected
and accurately manually segmented to provide a Panorama Segmentation Dataset (PSD).
These 75 ground-truth segmentations contain
between 118 and 1091 objects having an average size of 1334 pixels.
Examples of PSD images are given in Figure \ref{fig:data_ex_sps}.

\begin{figure*}[t]
{\scriptsize
\begin{tabular}{@{\hspace{0mm}}c@{\hspace{1mm}}c@{\hspace{3mm}}c@{\hspace{1mm}}c@{\hspace{3mm}}c@{\hspace{1mm}}c@{\hspace{0mm}}}
\includegraphics[width=\wsev,height=\hsev]{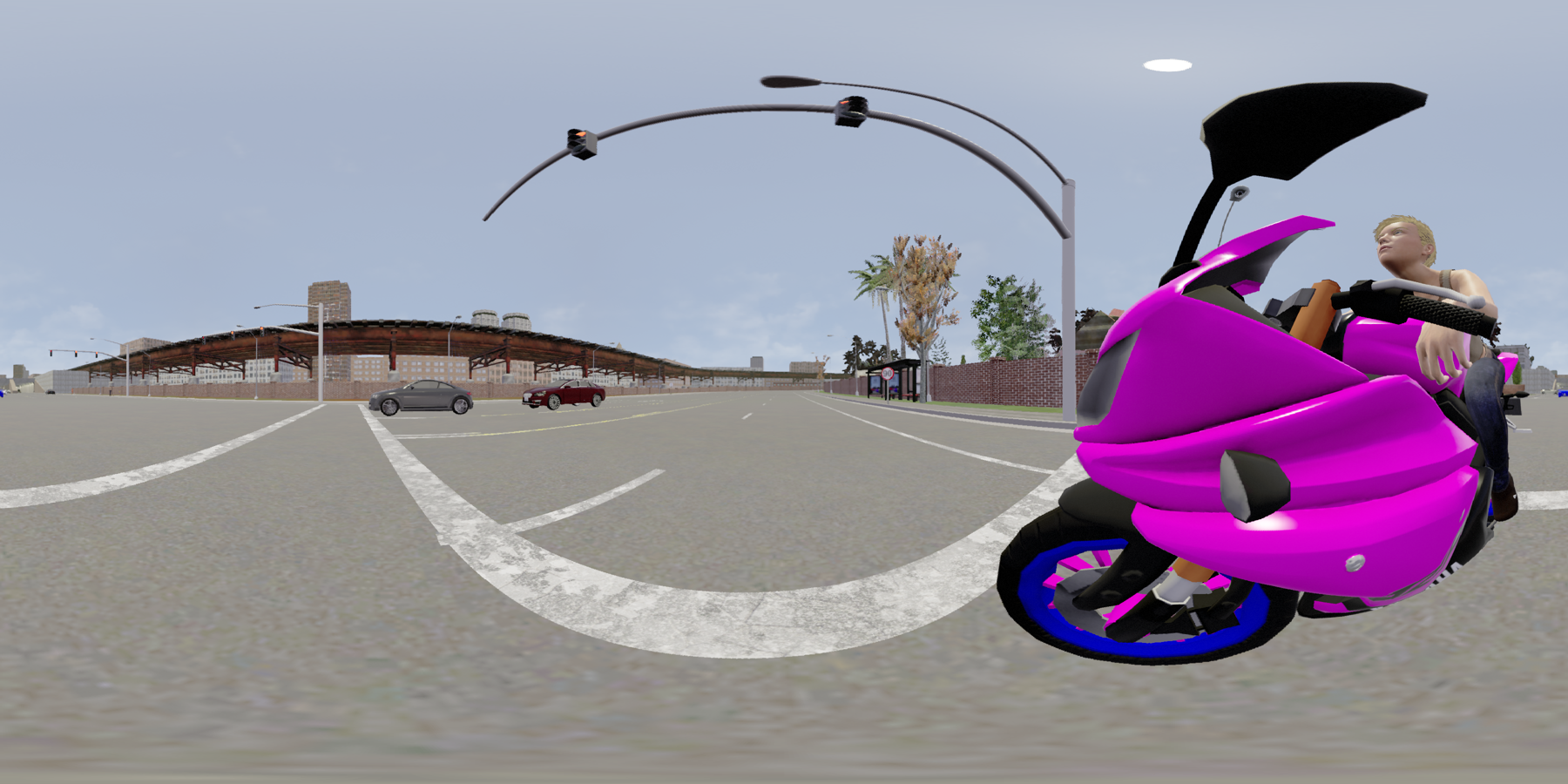}&
\includegraphics[width=\whsev,height=\whsev]{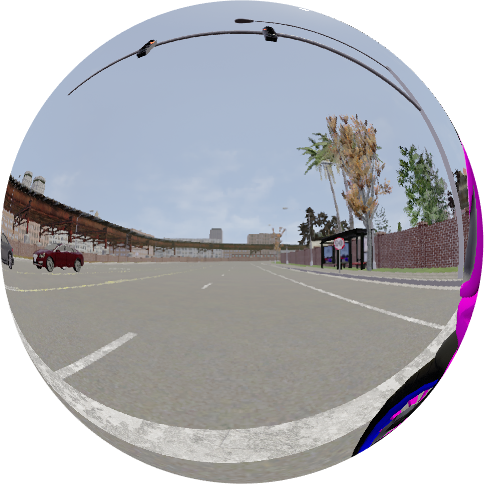}&
\includegraphics[width=\wsev,height=\hsev]{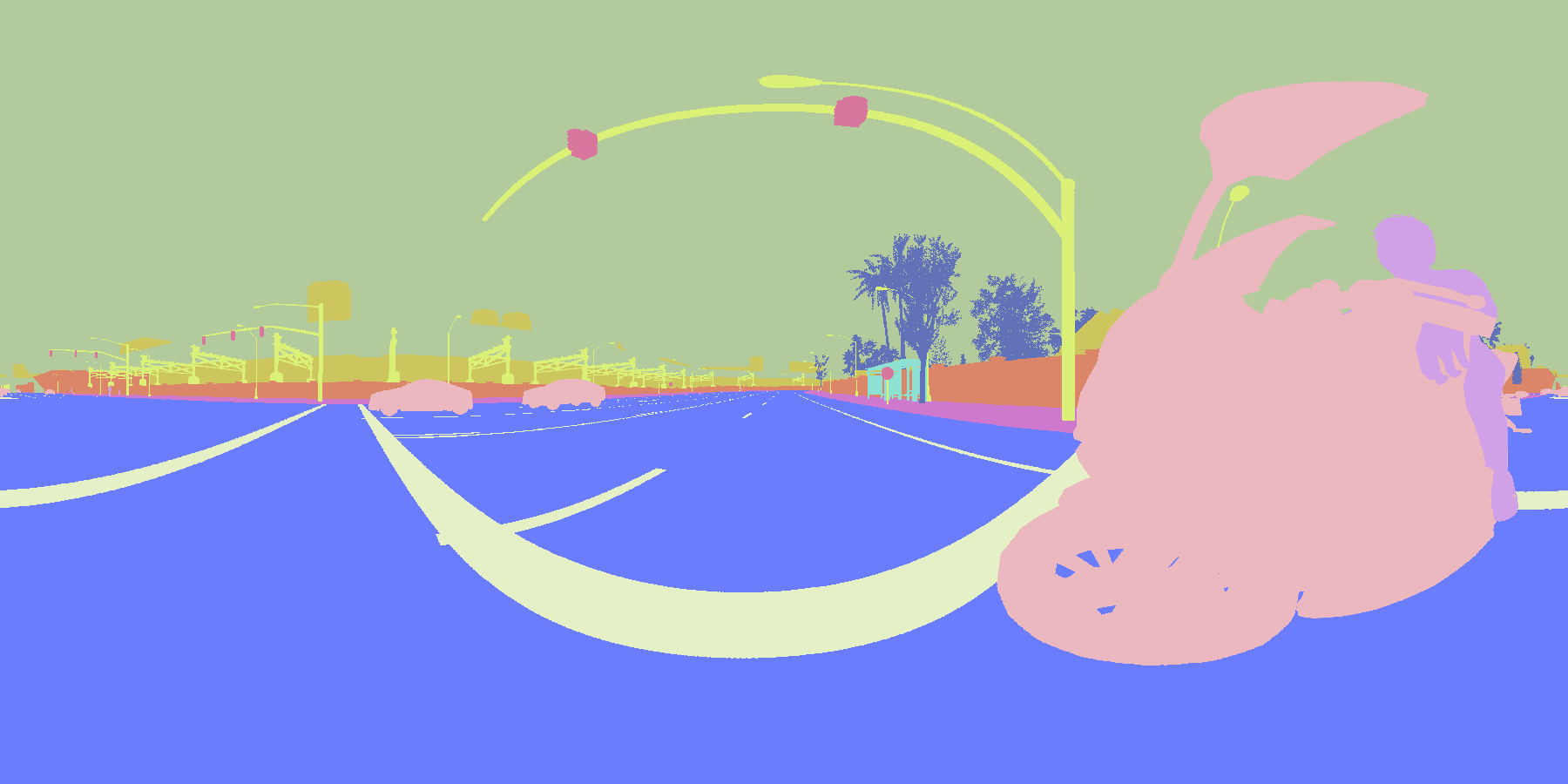}&
\includegraphics[width=\whsev,height=\whsev]{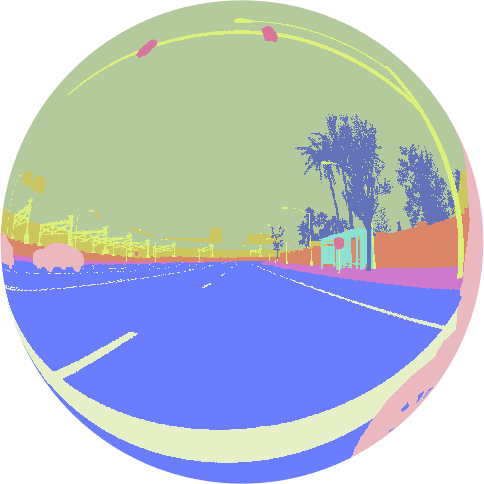}&
\includegraphics[width=\wsev,height=\hsev]{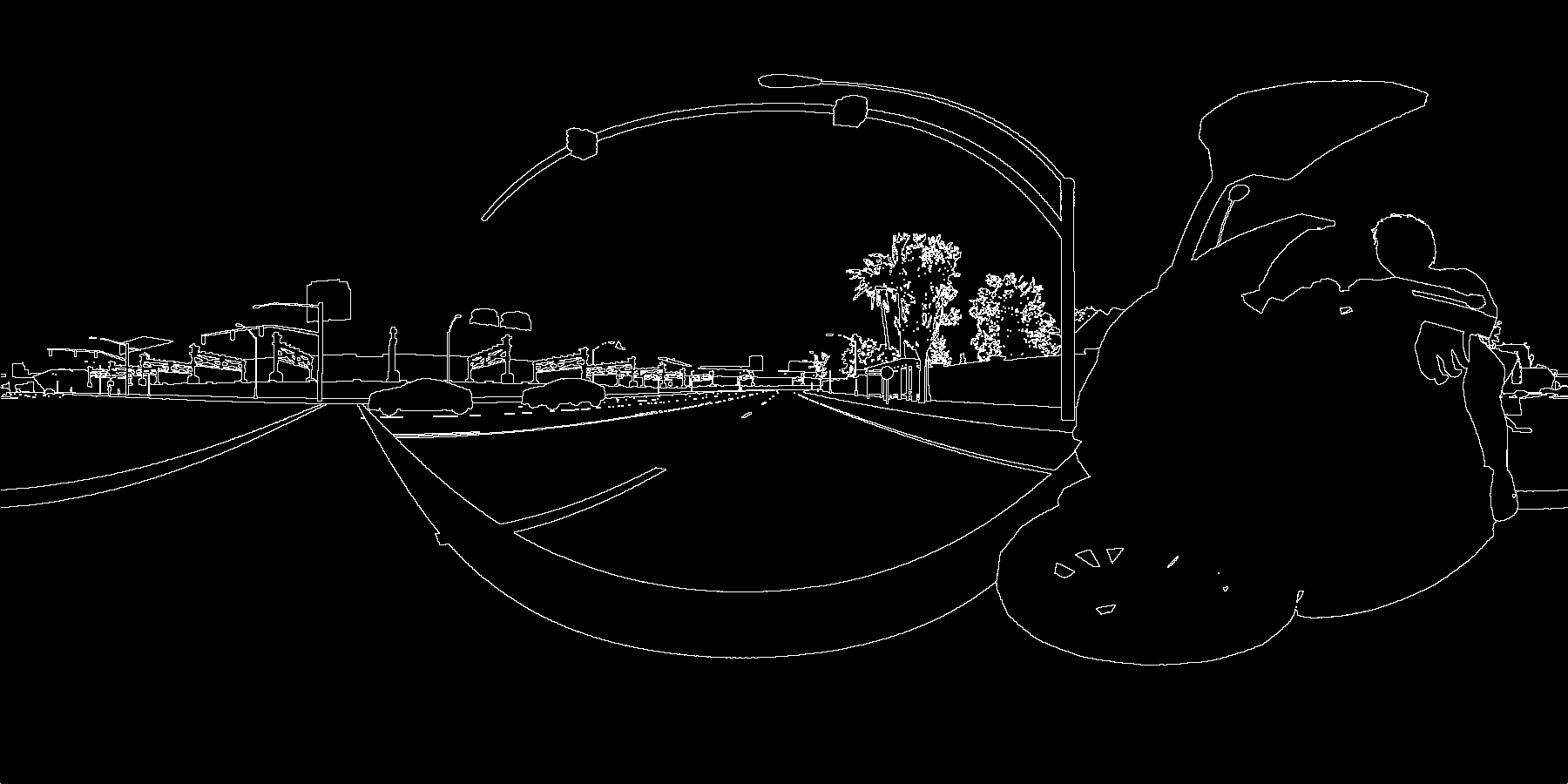}&
\includegraphics[width=\whsev,height=\whsev]{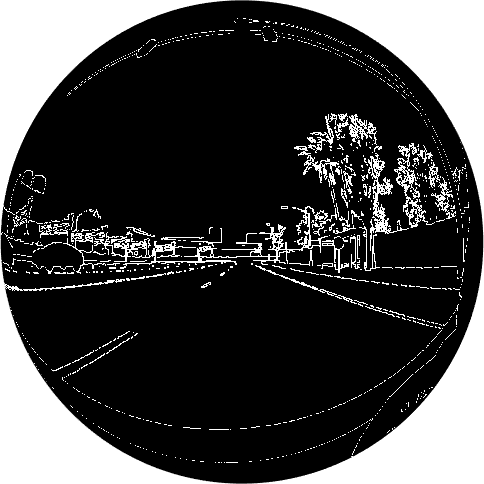}\\
\includegraphics[width=\wsev,height=\hsev]{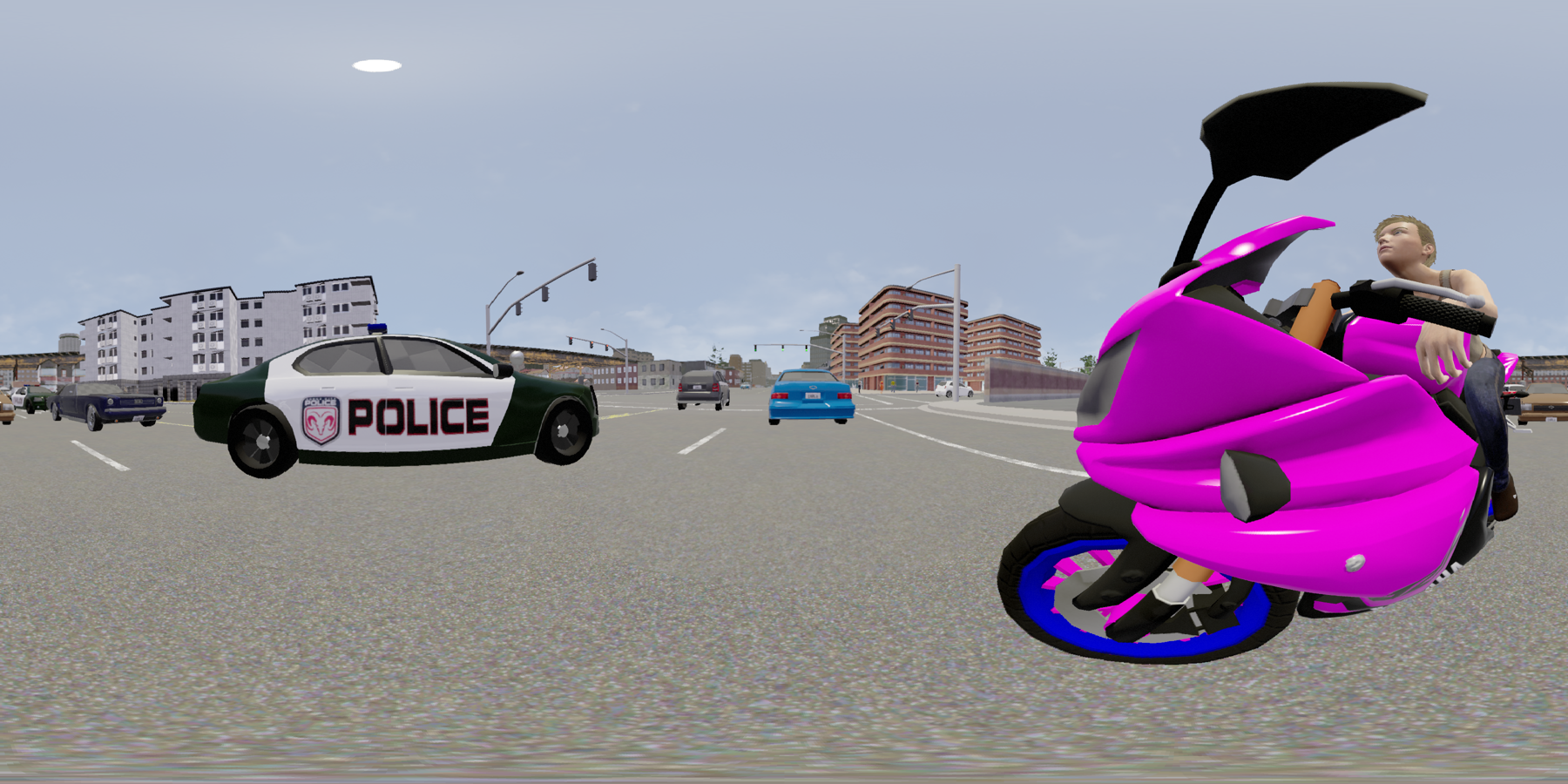}&
\includegraphics[width=\whsev,height=\whsev]{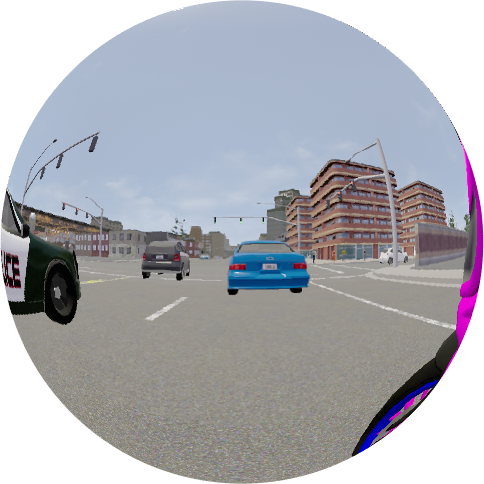}&
\includegraphics[width=\wsev,height=\hsev]{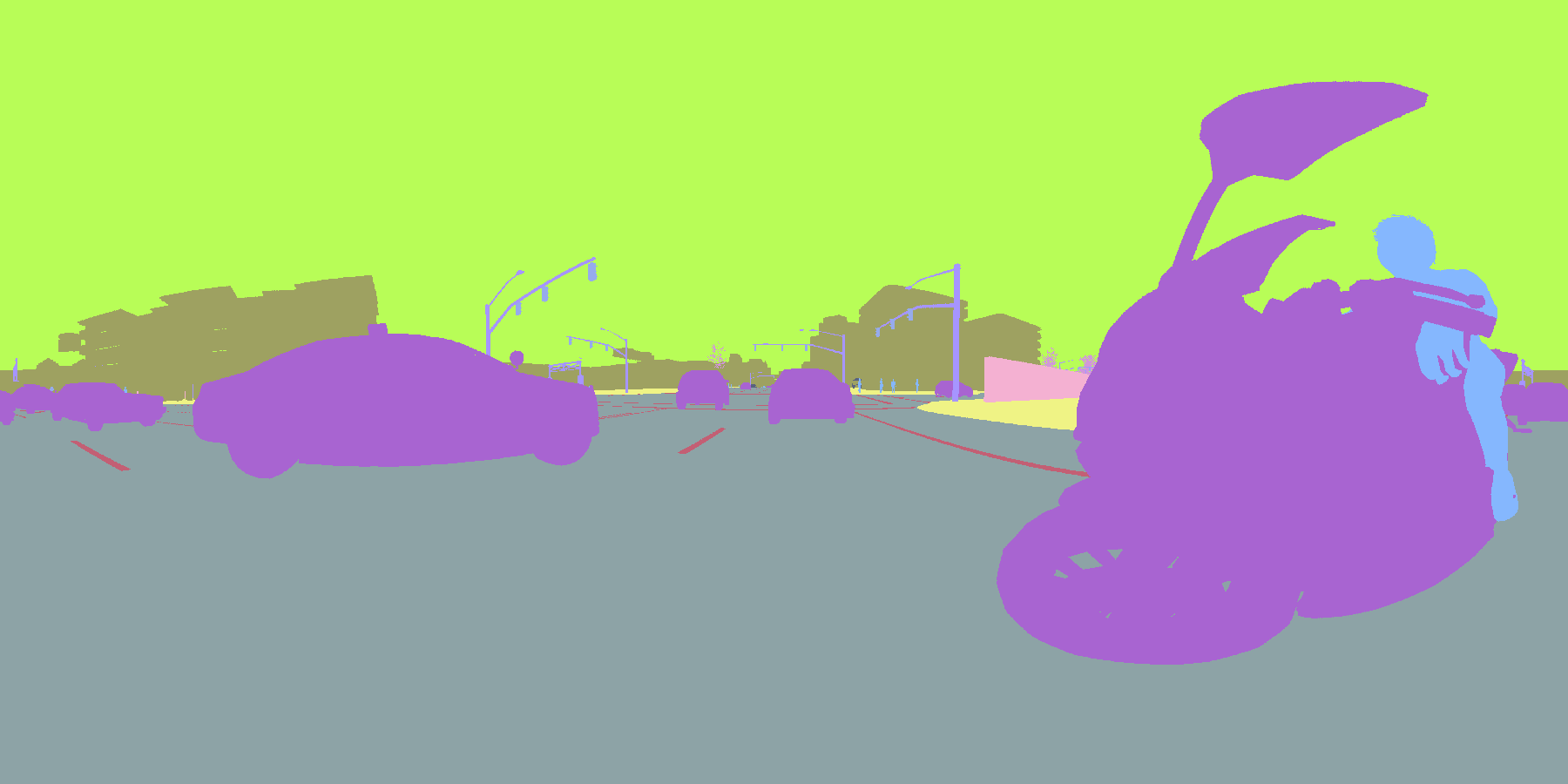}&
\includegraphics[width=\whsev,height=\whsev]{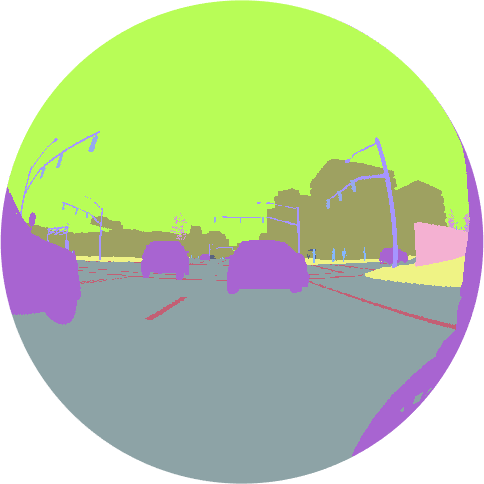}&
\includegraphics[width=\wsev,height=\hsev]{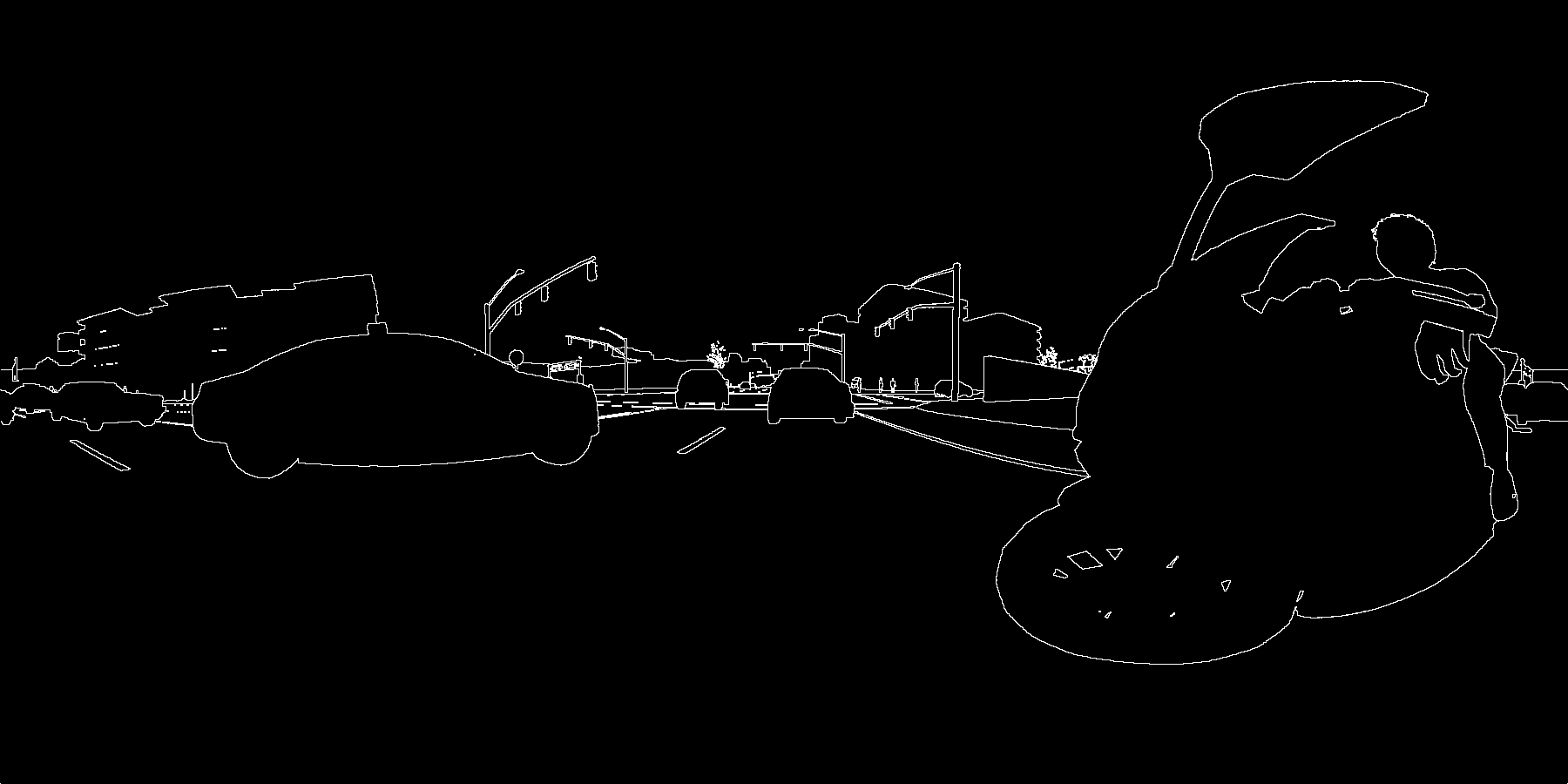}&
\includegraphics[width=\whsev,height=\whsev]{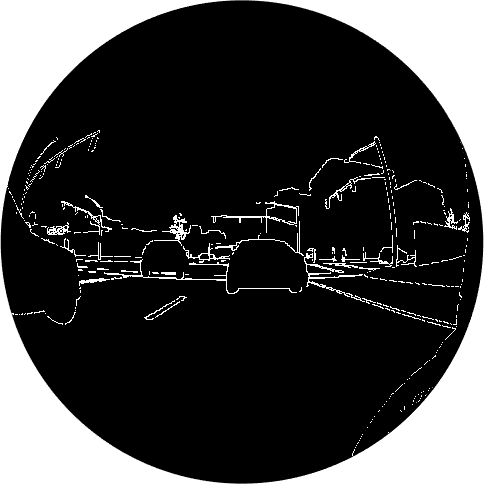}\\
\multicolumn{2}{c}{Image} & \multicolumn{2}{c}{Labels} & \multicolumn{2}{c}{Contours} \\
\end{tabular}
}
\caption{Examples of images from the Omniscape project \cite{sekkat2020omniscape}}
\label{fig:data_ex_omni}
\end{figure*}

{\color{review2}
To further demonstrate the applicability of the SphSPS method,
we also consider for the first time a new dataset of $100$ synthetic omnidirectional road images,
provided in the frame of the Omniscape project \cite{sekkat2020omniscape}.
}
Examples of these images,
of size $900{\times}1800$ pixels and containing up to $13$ classes
are given in Figure \ref{fig:data_ex_omni}.

{\color{review}
Because of their relatively high resolution, and the important number of segmented objects, these datasets are relevant to assess the segmentation performance.
Moreover, the number of validation images is similar to the one usually considered in the standard Berkeley Segmentation Dataset (BSD) containing between 100 and 200 test images of size 321x481 pixels.
}

\smallskip

\subsubsection{\label{subsec:metrics}Metrics}

To compare the performance of the SphSPS method to the ones of the state-of-the-art,
we use the contour detection, object segmentation and regularity metrics recommended in \cite{giraud2017_jei}.

In terms of contour detection, the Boundary-Recall (BR) is a commonly employed metric
evaluating the compliance of the superpixel boundaries $\mathcal{B}(\SSS)$
to the ground truth contours $\mathcal{B(\GG)}$  such that:
%\vspace{-0.15cm}
%
{\eqsize
\begin{equation}
\text{BR}(\SSS,\GG) = \frac{1}{|\mathcal{B}(\GG)|}\sum_{p\in\mathcal{B}(\GG)}\delta[\min_{q\in\mathcal{B}(\SSS)}\|p-q\|< \epsilon]  ,   \label{br}
\end{equation}
}%
\noindent with
 $\epsilon$ a distance threshold set to $2$ pixels \cite{giraud2017_jei},
 and $\delta[a]=1$ when $a$ is true and $0$ otherwise.
With BR, each ground truth pixel is considered as detected, if
a superpixel boundary is at less than an $\epsilon$ distance.
 The issue of the BR measure, is that it does not take into account the number of superpixel borders,
 so fuzzy methods can have high performances \cite{giraud2017_jei}.
Hence, BR results can be compared to  the number of pixels of superpixel borders,
called Contour Density (CD).

To go further in the evaluation of contour detection performance,
the standard Precision-Recall curves can also be represented.
These are computed from a contour probability map $\in [0,1]$
that is generated by averaging the superpixel borders obtained
at different segmentation scales $K\in [50,3000]$. % for a given image.
This normalized boundary map is then thresholded by several
intensities $\in[0,1]$ to get a binary contour map, and for each threshold,
the Precision (PR) (percentage of accurate detection among the superpixel borders)
is computed along with the BR measure.
The performance of each PR/BR curve can be summarized with
the maximum standard F-measure defined as:
%\vspace{-0.075cm}
%
{\eqsize
 \begin{equation}
 \text{F}=\frac{2.\text{PR.BR}}{\text{PR}+\text{BR}} . \label{fmeasure}
 \end{equation}}%

Since the dataset provides segmented regions,
object segmentation performance can also be evaluated.
To do so, we use
the standard Achievable Segmentation Accuracy (ASA) measure \cite{liu2011},
which has been proven highly correlated to the Undersegmentation Error %(UE)
\cite{neubert2012} in \cite{giraud2017_jei}.
The ASA evaluates the overlap of the superpixel segmentation $\SSS$
with the objects of
the ground truth segmentation, denoted $\GG$, such as:
%\vspace{-0.10cm}
%
{\eqsize
\begin{equation}
 \text{ASA}(\SSS,\GG) = \frac{1}{\sum\limits_{S_i \in \SSS}|S_i|}\sum_{S_i}\underset{G_j\in \GG}{\max}|S_i\cap G_j|.  \label{asa}
\end{equation}
}

Finally, the regularity aspect is robustly evaluated
in the acquisition space with the proposed G-GR metric \eqref{grs}.
To demonstrate its relevance, we also compare its results to the ones of the
spherical COM measure proposed in \cite{zhao2018} in Annex A.

\subsubsection{Parameter settings}

The SphSPS method was implemented in MATLAB using C-MEX code,
on a Linux computer having 12 cores at 2.6 GHz with 64GB of RAM.
SphSPS is based on the spherical $K$-means framework of \cite{zhao2018},
with $5$ iterations,
with a seed initialization following a Hammersley sampling \cite{wong1997sampling},
and a cosine dissimilarity spatial distance $d_s$ \eqref{ds_sph}.
It uses the 6 dimension CIELab color space of \cite{chen2017}, and
includes the features of neighboring pixels \cite{giraud2018_scalp}
in its color distance $d_c$.
In its spherical shortest path, by default $N=15$ pixels are considered \eqref{sph_path}.
The color distance trade-off
parameter $\lambda$ \eqref{path_color},
is set to $0.5$ as in \cite{giraud2018_scalp},
and when used, \emph{i.e.}, $\gamma \neq 0$,
the contour prior is computed from
\cite{xie2015holistically} and $\gamma$ is set to $10$ \eqref{path_contour}.
Finally, the regularity parameter $m$ \eqref{newdist} is empirically set to $0.12$ to
provide a satisfying trade-off between the spatial regularity of superpixels and their segmentation performance.

\subsection{Impact of Contributions\label{subsec:param_res}}

In this section, we show the impact of the contributions introduced in the SphSPS method
on contour detection performance using PR/BR curves with maximum F-measure \eqref{fmeasure},
on segmentation accuracy using ASA metric \eqref{asa},
and on regularity with the proposed G-GR metric \eqref{grs}.
First, in Figure \ref{fig:init_res}, we quantitatively evaluate the
impact of the initialization strategies represented in Figure \ref{fig:init}.
%
%with the maximum F-measure \eqref{fmeasure},
The spherical seed samplings logically provide much higher accuracy,
since superpixels are regularly placed over the acquisition space
and can accurately follow object contours.
For instance, for a low number of superpixels, planar initialization would set
a large number of seeds on top and bottom of the image where
very few objects generally are.
After a few iterations, we observe that
the spherical distance of the SphSPS method is able to
provide a spherically regular decomposition with any input method.
Nevertheless, with a high number of superpixels,
the two most regular spherical initialization methods, Hammersley and Geodesic clearly
obtain the highest regularity.
In the following, due to the complexity and lack of exact control of number of superpixels with the geodesic approach, we use the Hammersley sampling in the SphSPS method.

\newcommand{\wwh}{0.32\textwidth}
\newcommand{\hhh}{0.26\textwidth}

\begin{figure*}[t!]
\centering
{\scriptsize
\begin{tabular}{@{\hspace{0mm}}c@{\hspace{1mm}}c@{\hspace{1mm}}c@{\hspace{0mm}}}
\includegraphics[width=\wwh,height=\hhh]{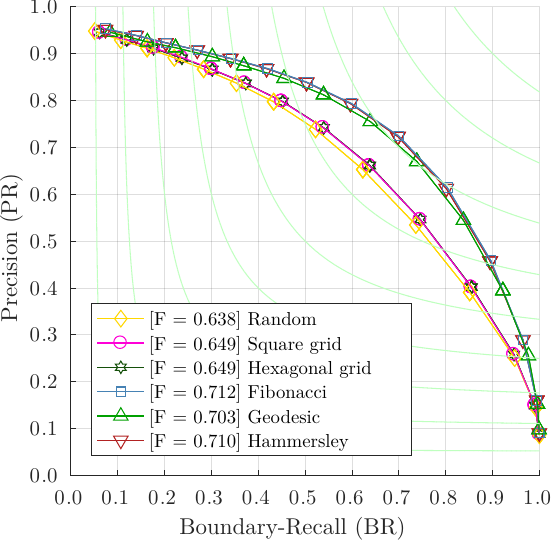}&
\includegraphics[width=\wwh,height=\hhh]{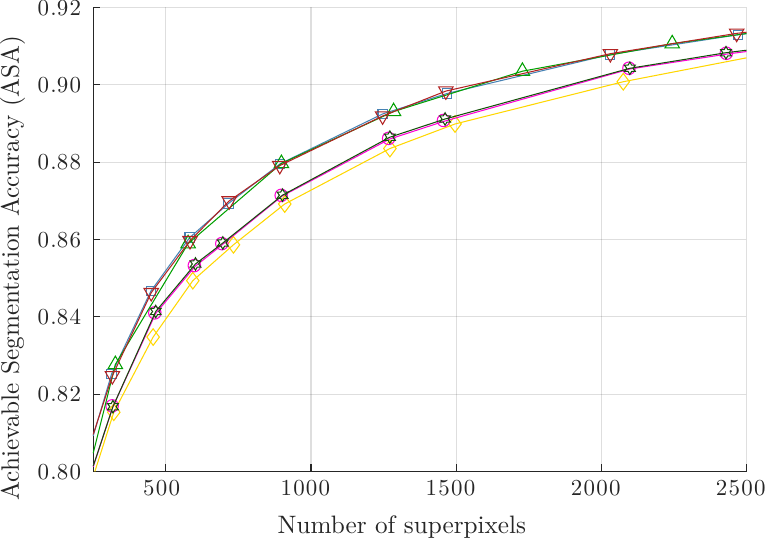}&
\includegraphics[width=\wwh,height=\hhh]{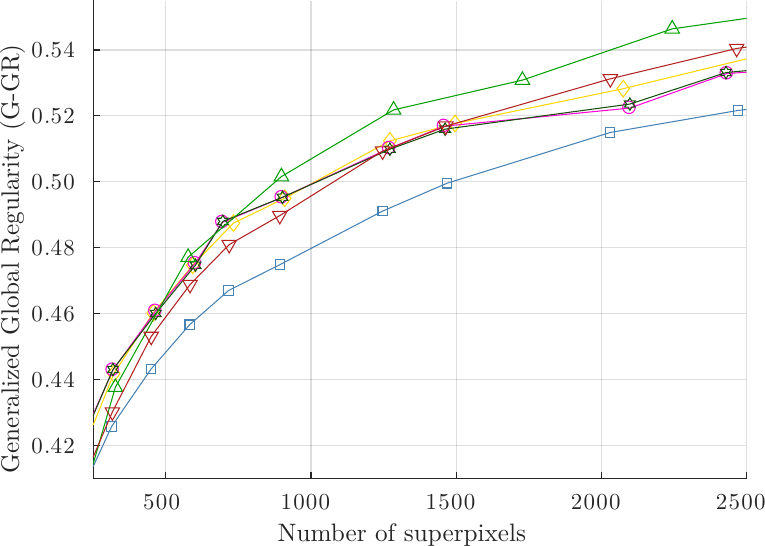}
\end{tabular}}
\caption{Impact of the seed initialization strategy on the performance of SphSPS evaluated on the PSD dataset \cite{wan2018}.
The different sampling strategies are represented in Figure \ref{fig:init}}%
\label{fig:init_res}
\end{figure*}

\begin{figure*}[t!]
\centering
{\scriptsize
\begin{tabular}{@{\hspace{0mm}}c@{\hspace{1mm}}c@{\hspace{1mm}}c@{\hspace{0mm}}}
\includegraphics[width=\wwh,height=\hhh]{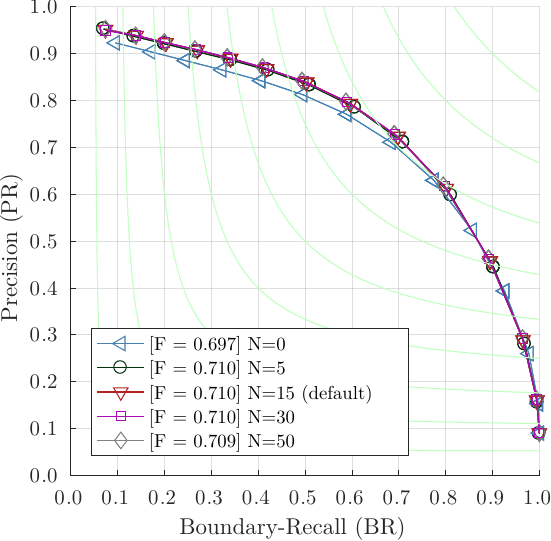}&&
\includegraphics[width=\wwh,height=\hhh]{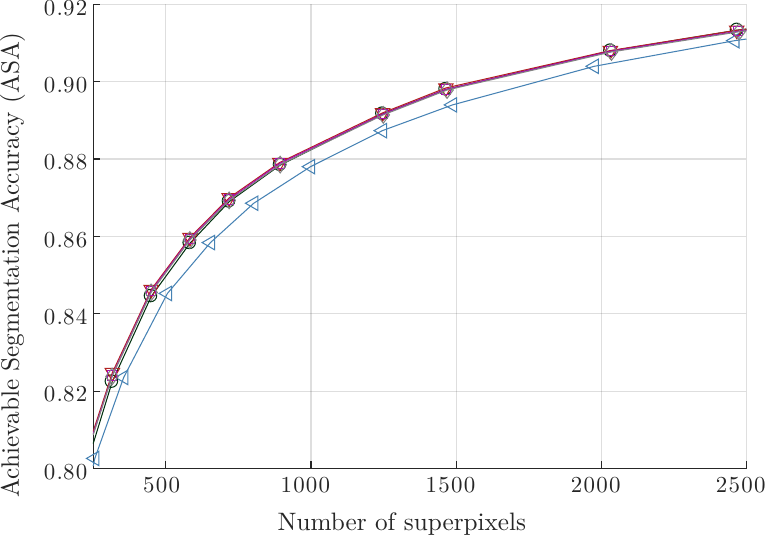}
\includegraphics[width=\wwh,height=\hhh]{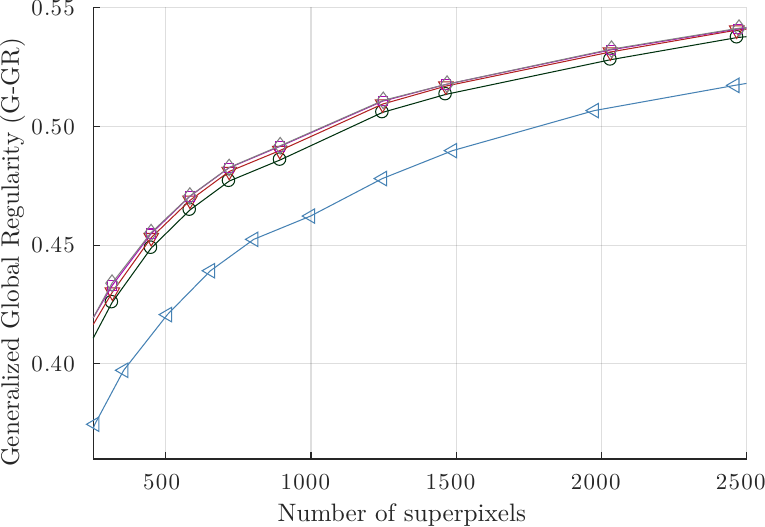}
\end{tabular}}
\caption{Impact of the number of points $N$ in the spherical shortest path  \eqref{sph_path} on the performance of SphSPS evaluated on the PSD dataset \cite{wan2018}}%
\label{fig:N_path_res}
\end{figure*}

Then, in Figure \ref{fig:N_path_res},
we report the influence of the number of points in the shortest path \eqref{sph_path}.
More sampling points enable to more accurately follow the spherical shortest path
to capture homogeneous colors, so the shape regularity is enforced with the number of points, and reaches a plateau around $N=15$.
The same behavior can be observed for the PR curves where a plateau is obtained with a low number of points thanks to our recursive implementation (see Section \ref{subsubsec:gen_path}).
In the following we use a trade-off number of $N=15$ points in the shortest path.

\begin{figure*}[t!]
\centering
{\scriptsize
\begin{tabular}{@{\hspace{0mm}}c@{\hspace{1mm}}c@{\hspace{1mm}}c@{\hspace{0mm}}}
\includegraphics[width=\wwh,height=\hhh]{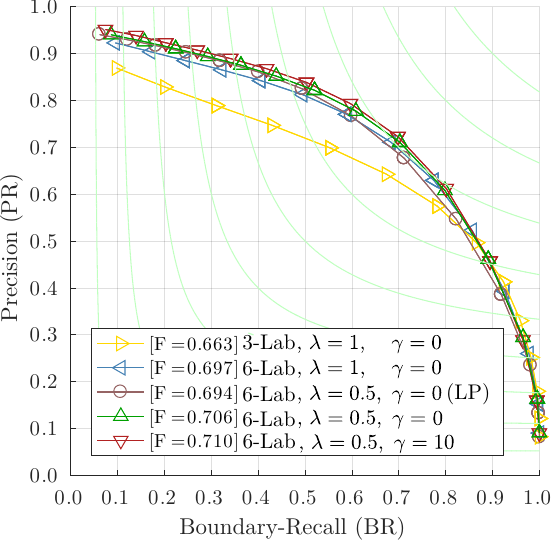}&
\includegraphics[width=\wwh,height=\hhh]{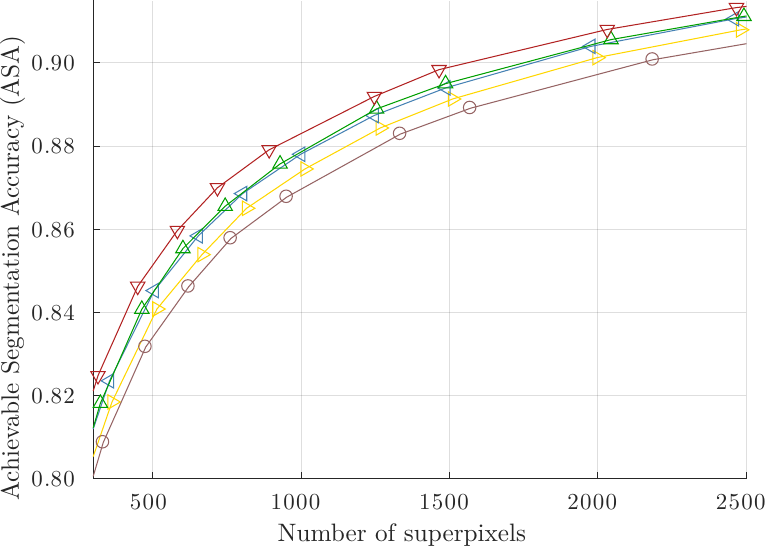}&
\includegraphics[width=\wwh,height=\hhh]{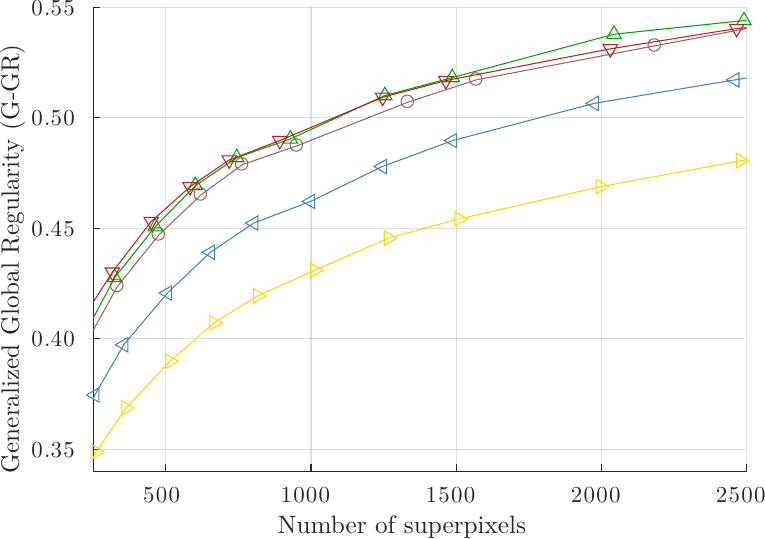}
\end{tabular}}
\caption{Impact of the distance parameters on the performance of SphSPS evaluated on the PSD dataset \cite{wan2018}.
The contributions enable to significantly improve the
accuracy and regularity performances}%
\label{fig:sps_param_curves}
\end{figure*}

\begin{figure*}[t!]
\newcommand{\gwwh}{0.26\textwidth}
\newcommand{\whhh}{0.2\textwidth}
\centering
{\scriptsize
\begin{tabular}{@{\hspace{0mm}}c@{\hspace{2mm}}c@{\hspace{0mm}}}
\includegraphics[width=\gwwh,height=\whhh]{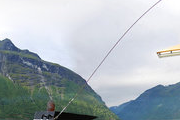}&
\includegraphics[width=\gwwh,height=\whhh]{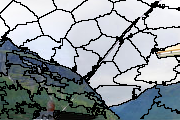}\\
 (a) Initial image & (b) 3-Lab, $\lambda$=$1,\gamma$=$0$\\[2ex]
\end{tabular}
\begin{tabular}{@{\hspace{0mm}}c@{\hspace{2mm}}c@{\hspace{2mm}}c@{\hspace{0mm}}}
\includegraphics[width=\gwwh,height=\whhh]{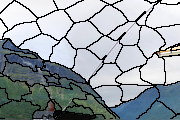}&
\includegraphics[width=\gwwh,height=\whhh]{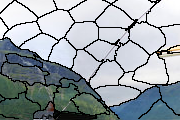}&
\includegraphics[width=\gwwh,height=\whhh]{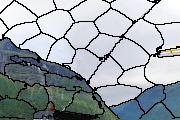}\\
 (c) \textbf{6-Lab}$,\lambda$=$1,\gamma$=$0$ &
(d) {6-Lab}$, $ {\color{black}\textbf{$\mathbf{\lambda}$=$\mathbf{0.5}$}}$,\gamma$=$0$ &
(e) {6-Lab}$,\lambda$=$0.5,$ {\color{black}\textbf{$\mathbf{\gamma}$=$\mathbf{10}$}}\\
\end{tabular}}%
\caption{Visual impact of  SphSPS parameters. Each contribution
relevantly increases the regularity and $\gamma=10$
integrates the contour prior information
}%
\label{fig:sps_param_ex}
\end{figure*}

{\color{review}
%We report for
We show the impact of the different distance settings in SphSPS
on PR/BR, ASA and G-GR curves
in Figure \ref{fig:sps_param_curves}.
With a 3 feature dimension space (3-Lab, $\lambda$=$1$, and $\gamma$=$0$),
SphSPS reduces to an implementation of the spherical SLIC algorithm \cite{zhao2018}.
With the 6 dimension space (6-Lab) SphSPS uses the CIELab color features of \cite{chen2017}
with the information of the neighboring pixels as in \cite{giraud2018_scalp},
increasing accuracy and regularity.
With $\lambda$ set to $0.5$,
SphSPS considers in its model the average color distance on the path \eqref{path_color},
and with $\gamma=10$, the maximum contour information \eqref{path_contour}.
We observe that each contribution within the SphSPS method improves the segmentation performance.
We also compare the results obtained with the SphSPS method using a shortest linear path (LP), represented in Figure \ref{fig:spherical_path}.
These results demonstrate the interest of using our spherical path,
since the exact same method using the initial 2D planar shortest path \cite{giraud2018_scalp} reports lower performance.

Finally, the joint contributions of the different terms in SphSPS
are illustrated on a zoom on a segmentation example in Figure \ref{fig:sps_param_ex}.
First, the larger 6-Lab dimension space helps to reduce the noise at the superpixel borders (Figure \ref{fig:sps_param_ex}(c)).
Then, we see that the color distance on the shortest path ($\lambda=0.5$),
strengthens the superpixel convexity and homogeneity (Figure \ref{fig:sps_param_ex}(d)),
to provide much more regular superpixels in a relevant manner
that also increases the segmentation accuracy.
Finally, the contour term ($\gamma=10$), enables to capture thin contours to improve segmentation accuracy (Figure \ref{fig:sps_param_ex}(e)).

}

\subsection{\label{subsec:soa}Comparison with the State-of-the-Art Methods}

The performance
of the proposed SphSPS method
are compared to the ones of
state-of-the-art approaches
on the standard PSD dataset \cite{wan2018}
and also on a set of synthetic images extracted from the Omniscape project \cite{sekkat2020omniscape}
on the metrics presented in Section \ref{subsec:metrics}.
We consider the recent planar methods
SLIC \cite{achanta2012},
ETPS \cite{yao2015},
LSC \cite{chen2017},
SNIC \cite{achanta2017superpixels},
SCALP \cite{giraud2018_scalp},
and
GMMSP \cite{Ban18}.
We also compare our SphSPS to
the spherical approach SphSLIC \cite{zhao2018}, in its two different settings, \emph{i.e.},
using an Euclidean spatial distance (SphSLIC-Euc)
and the same Cosine (SphSLIC-Cos) distance as SphSPS \eqref{ds_sph}.
For further comparison to spherical methods, we also implement the natural extension of the LSC method \cite{chen2017} to spherical images (SphLSC),
\emph{i.e.}, using the SphSLIC framework \cite{zhao2018} and the 6 dimension features of \cite{chen2017}.
% %
To ensure fair comparison between methods,
results are reported using the default regularity setting recommended by the authors,
providing a good trade-off between accuracy and regularity.
Note that only for the SphSLIC-Cos method \cite{zhao2018},
no default parameter is provided so we use the one
optimizing the segmentation accuracy.

\begin{figure*}[t!]
\centering
{\scriptsize
\begin{tabular}{@{\hspace{0mm}}c@{\hspace{1mm}}c@{\hspace{1mm}}c@{\hspace{0mm}}}
\includegraphics[width=\wwh,height=\hhh]{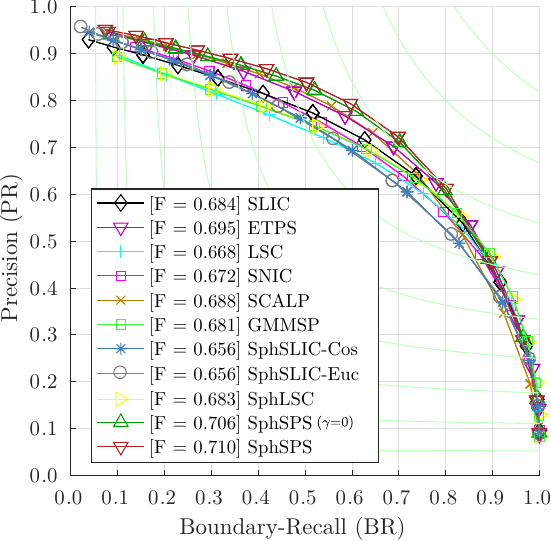}&
\includegraphics[width=\wwh,height=\hhh]{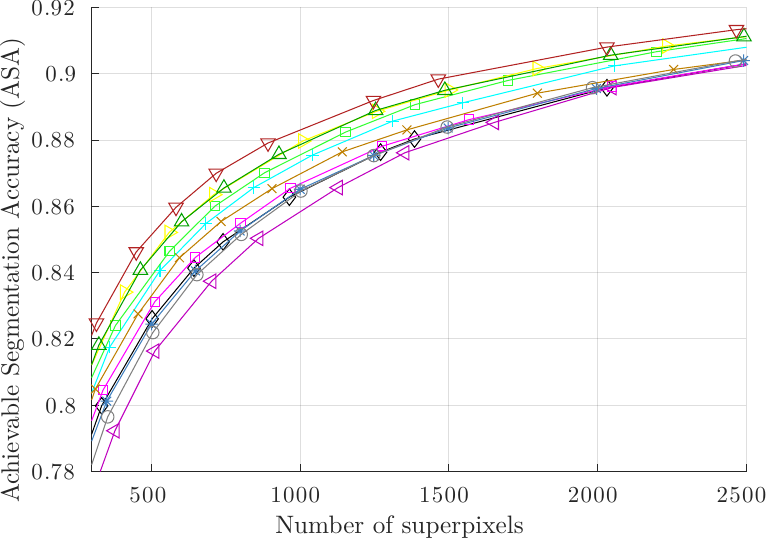}&
\includegraphics[width=\wwh,height=\hhh]{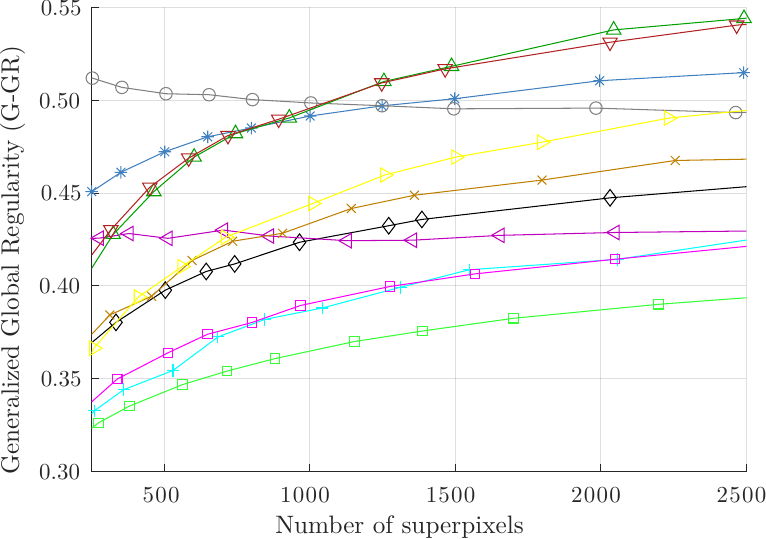}\\
% (a) P-R (F-measure) & (b) BR vs CD & (c) ASA & (d) GR \\
\end{tabular}
}
\caption{
Quantitative comparison on the PSD images \cite{wan2018}, on PR/BR, %BR/CD,
ASA and G-GR metrics
of the proposed SphSPS method to the state-of-the-art ones}%
\label{fig:sps_soa}
\end{figure*}

\begin{figure*}[t!]
\centering
{\scriptsize
\begin{tabular}{@{\hspace{0mm}}c@{\hspace{1mm}}c@{\hspace{1mm}}c@{\hspace{0mm}}}
\includegraphics[width=\wwh,height=\hhh]{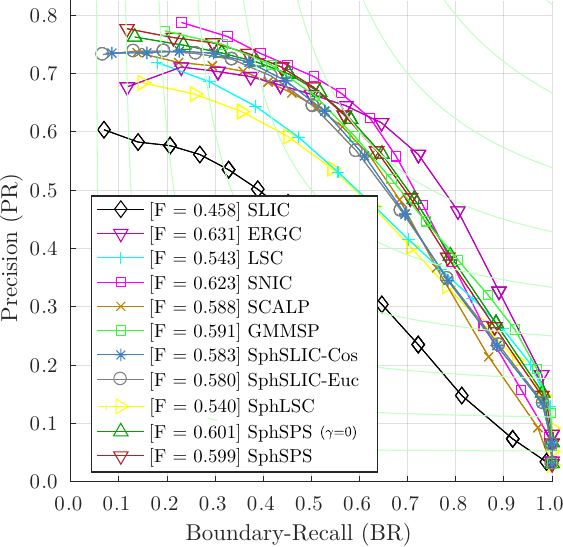}&
\includegraphics[width=\wwh,height=\hhh]{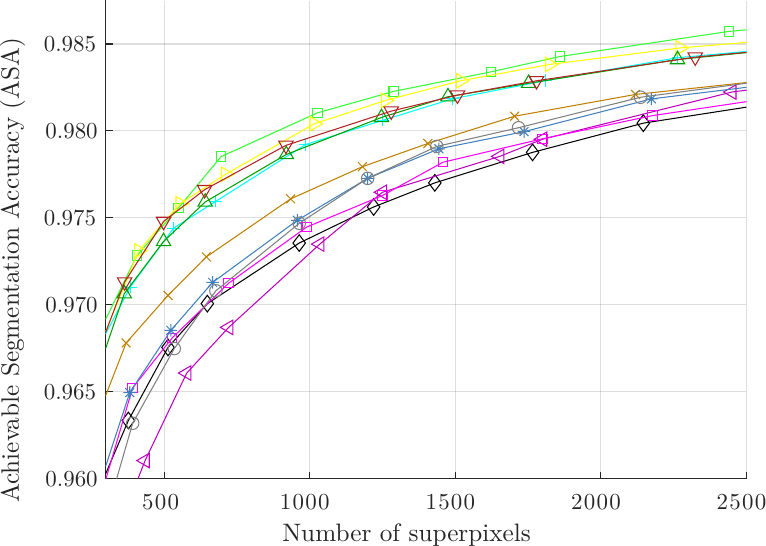}&
\includegraphics[width=\wwh,height=\hhh]{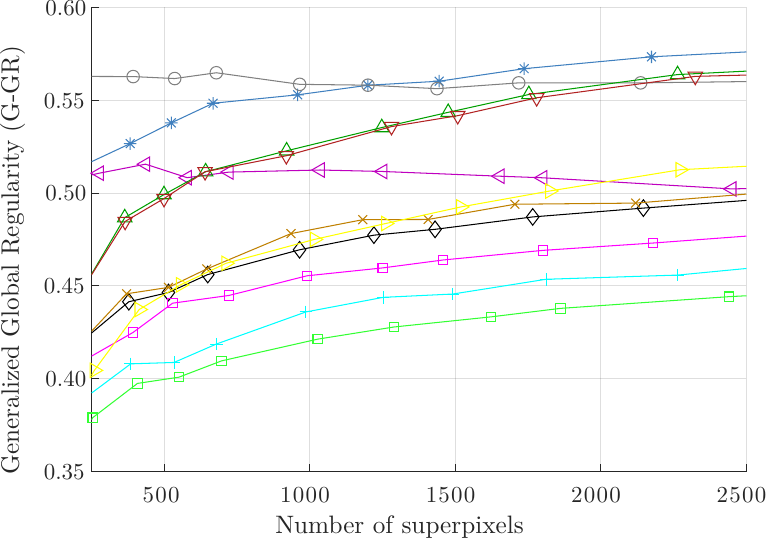}\\
% (a) P-R (F-measure) & (b) BR vs CD & (c) ASA & (d) GR \\
\end{tabular}
}
\caption{
Quantitative comparison on the synthetic Omniscape images \cite{sekkat2020omniscape}
of the proposed SphSPS method to the state-of-the-art ones}%
\label{fig:sps_soa_omni}
\end{figure*}

We report the quantitative performance for several numbers of superpixels
of SphSPS and the compared state-of-the-art methods
in Figures \ref{fig:sps_soa} and \ref{fig:sps_soa_omni}.
% %
First, on the PSD dataset (Figure \ref{fig:sps_soa}), we observe that SphSPS obtains the best segmentation and contour detection performance with %, the lowest BR/CD,
the highest ASA \eqref{asa} and the highest F-measure \eqref{fmeasure}
of $\text{F}=0.710$,
%on the PSD dataset,
even without using a contour prior ($\gamma=0$), \emph{i.e.}, only using
color information on the shortest path ($\text{F}=0.706$).
A significant improvement is obtained over the other spherical methods SphLSC and SphSLIC
using both distances, while producing more regular superpixels.
SphSPS indeed succeeds in producing both accurate and spherically regular superpixels.
On the synthetic road images (Figure \ref{fig:sps_soa_omni}), SphSPS also offers the best trade-off
on all metrics. While SphSPS is ranked 3rd of F-measure, the two better methods on this criteria
(ETPS \cite{yao2015} and SNIC \cite{achanta2017superpixels}) poorly segment the image objects according to the ASA metric.
On this last criteria, SphSPS is ranked 3rd, after the GMMSP method \cite{Ban18} that is the less regular among all methods,
and the SphLSC method (\cite{chen2017,zhao2018}) which also produces very fuzzy superpixels and has a low F-measure.

The proposed G-GR \eqref{grs} provides a relevant regularity measure
able to differentiate planar and spherical methods.
In particular, with a representative number of superpixels, no planar methods have higher regularity than spherical ones
for a given number of superpixels,
contrary to the results of the COM metric \cite{zhao2018} reported in Annex A.
G-GR results are also better correlated to the evolution of the superpixel scale,
which is relevant since, to a certain extent, smaller superpixels are more likely to have regular shapes.

Examples of segmentation on the 360$^\text{o}$ equirectangular images,
and projected on the unit sphere, with superpixel borders and labels,
for the proposed SphSPS and the best state-of-the-art methods
are represented in Figures \ref{fig:sps_soa_img_1} and
\ref{fig:sps_soa_img_2} for the PSD dataset,
and in Figure \ref{fig:sps_soa_img_omni} for the Omniscape images.
On both image type, we observe that SphSPS generates very regular superpixels in
the spherical space while it is able to accurately capture the image objects compared to other methods.

\begin{figure*}[ht!]
\centering
\newcommand{\wwx}{0.26\textwidth}
\newcommand{\pppx}{0.13\textwidth}
{\scriptsize
\begin{tabular}{@{\hspace{1mm}}c@{\hspace{1mm}}c@{\hspace{1mm}}c@{\hspace{1mm}}c@{\hspace{1mm}}c@{\hspace{1mm}}c@{\hspace{1mm}}c@{\hspace{0mm}}}
\multicolumn{2}{@{\hspace{1mm}}c@{\hspace{1mm}}}{\includegraphics[width=\wwx,height=\pppx]{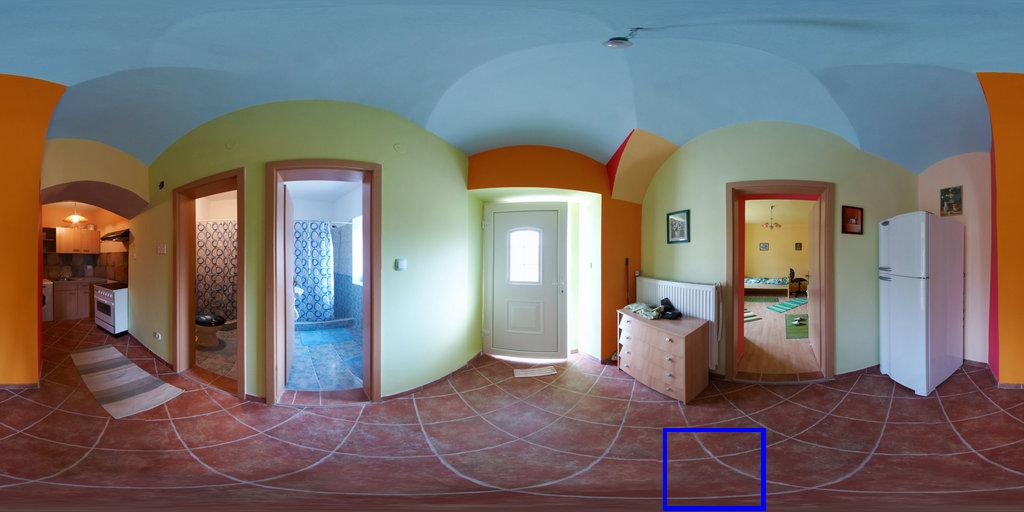}}&
\multicolumn{2}{@{\hspace{1mm}}c@{\hspace{1mm}}}{\includegraphics[width=\wwx,height=\pppx]{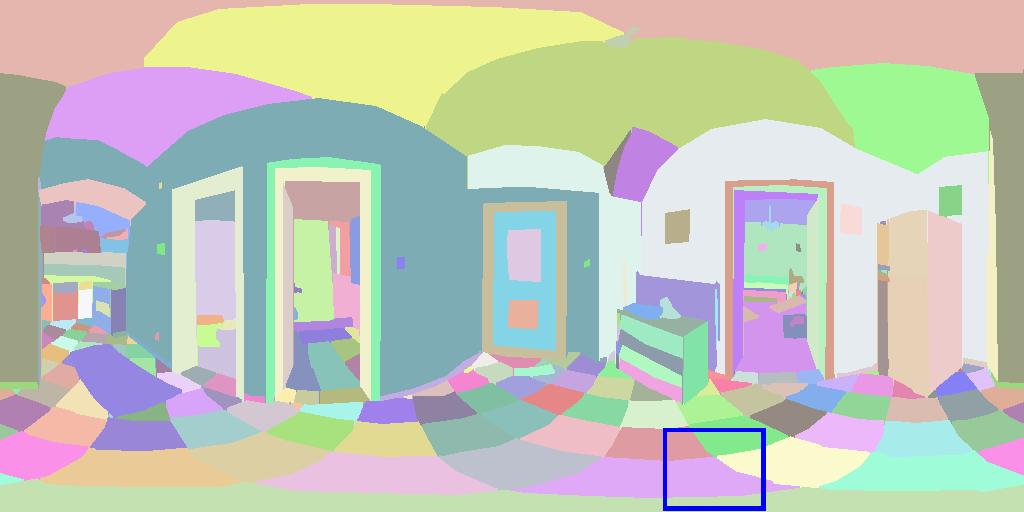}}&
\includegraphics[width=\pppx,height=\pppx]{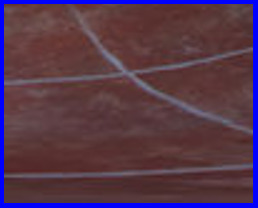}&
\includegraphics[width=\pppx,height=\pppx]{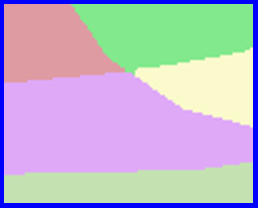}&
\includegraphics[width=\pppx,height=\pppx]{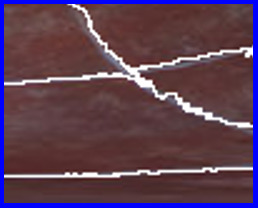}\\
\multicolumn{2}{c}{Image} & \multicolumn{2}{c}{Labels} & Image & Labels & \textbf{SphSPS}\\[1ex]
\includegraphics[width=\pppx,height=\pppx]{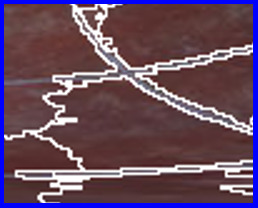}&
\includegraphics[width=\pppx,height=\pppx]{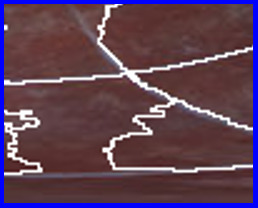}&
\includegraphics[width=\pppx,height=\pppx]{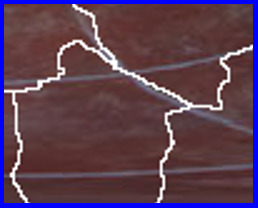}&
\includegraphics[width=\pppx,height=\pppx]{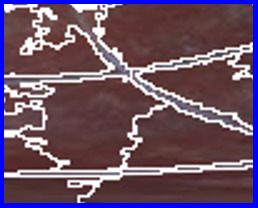}&
\includegraphics[width=\pppx,height=\pppx]{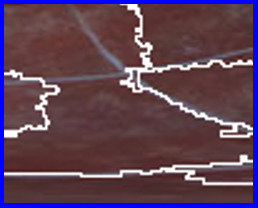}&
\includegraphics[width=\pppx,height=\pppx]{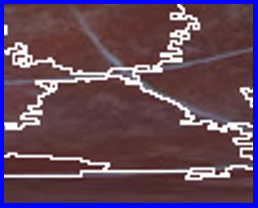}&
\includegraphics[width=\pppx,height=\pppx]{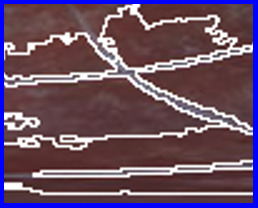}\\[-0.5ex]
{LSC \cite{chen2017}}&
{SNIC \cite{achanta2017superpixels}}&
{SCALP \cite{giraud2018_scalp}} &
GMMSP \cite{Ban18} &
{{SphSLIC-Euc \cite{zhao2018}}}&
{{SphSLIC-Cos \cite{zhao2018}}} &
{SphLSC \cite{chen2017,zhao2018}}\\[2ex]
\multicolumn{2}{@{\hspace{1mm}}c@{\hspace{1mm}}}{\includegraphics[width=\wwx,height=\pppx]{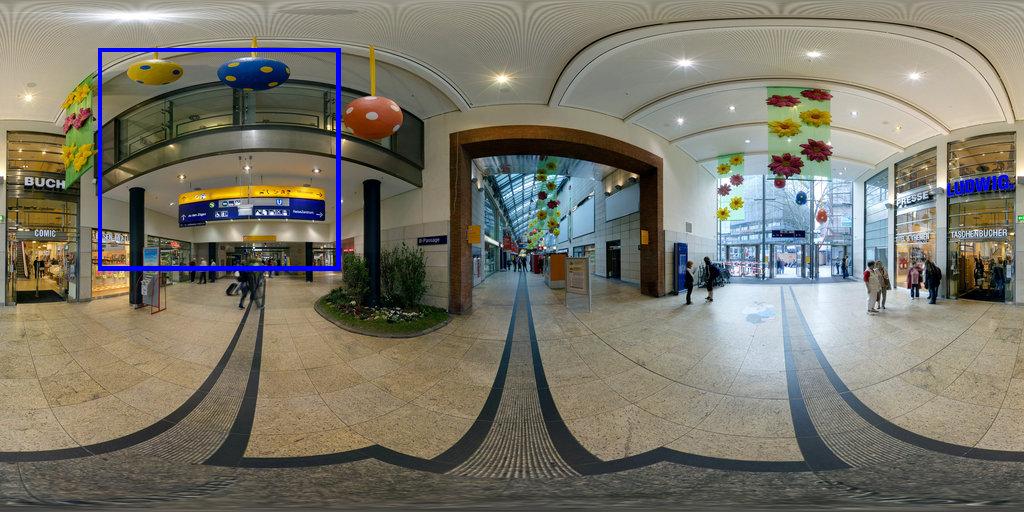}}&
\multicolumn{2}{@{\hspace{1mm}}c@{\hspace{1mm}}}{\includegraphics[width=\wwx,height=\pppx]{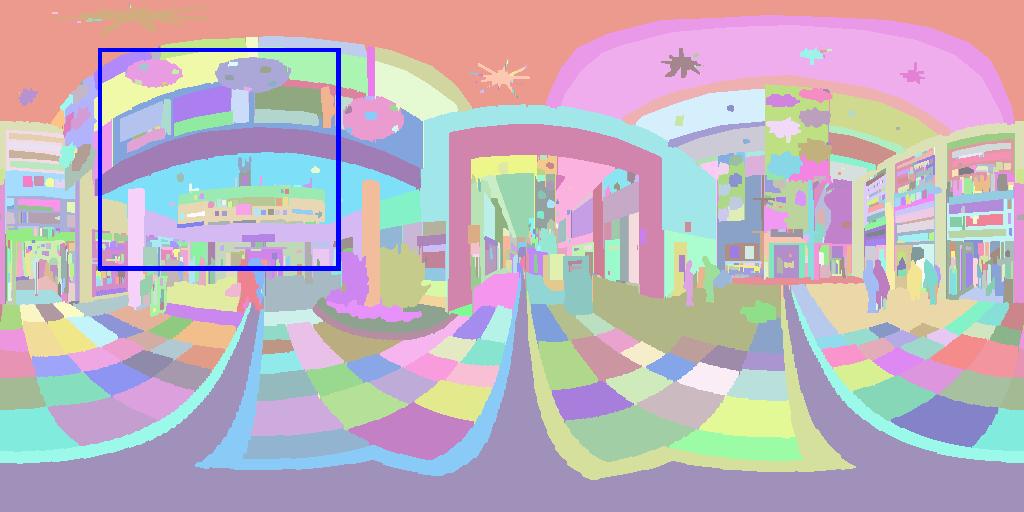}}&
\includegraphics[width=\pppx,height=\pppx]{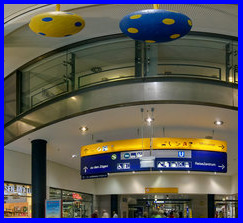}&
\includegraphics[width=\pppx,height=\pppx]{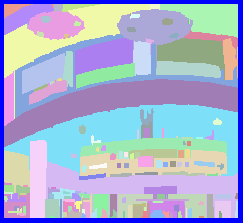}&
\includegraphics[width=\pppx,height=\pppx]{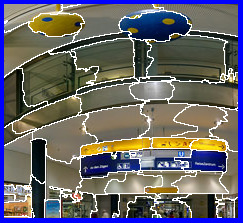}\\
\multicolumn{2}{c}{Image} & \multicolumn{2}{c}{Labels} & Image & Labels & \textbf{SphSPS}\\[1ex]
\includegraphics[width=\pppx,height=\pppx]{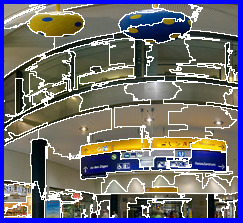}&
\includegraphics[width=\pppx,height=\pppx]{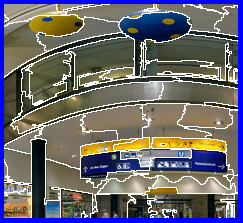}&
\includegraphics[width=\pppx,height=\pppx]{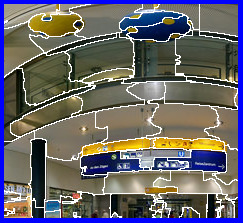}&
\includegraphics[width=\pppx,height=\pppx]{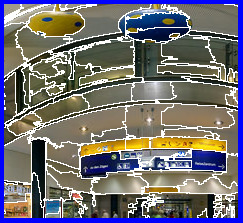}&
\includegraphics[width=\pppx,height=\pppx]{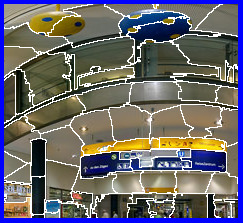}&
\includegraphics[width=\pppx,height=\pppx]{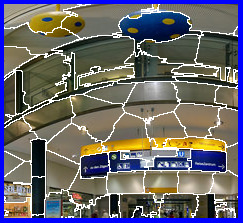}&
\includegraphics[width=\pppx,height=\pppx]{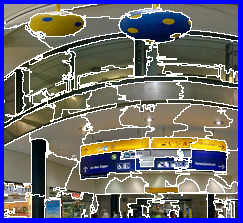}\\[-0.5ex]
{LSC \cite{chen2017}}&
{SNIC \cite{achanta2017superpixels}}&
{SCALP \cite{giraud2018_scalp}} &
GMMSP \cite{Ban18} &
{{SphSLIC-Euc \cite{zhao2018}}}&
{{SphSLIC-Cos \cite{zhao2018}}} &
{SphLSC \cite{chen2017,zhao2018}}\\
\end{tabular}
}
\caption{{\color{review}Failures examples on the PSD dataset for $K=400$ superpixels.
Most errors for all methods are due to thin object contours (top).
Note that SphSPS explicitly considers a contour term in its model preventing from high failures on this aspect.
Poor performance may also be due to a high resolution ground truth segmentation containing small objects that cannot be captured by the used superpixel scale (bottom).}
}%
\label{fig:errors}
\end{figure*}

{\color{review}
We also display in Figure \ref{fig:errors} segmentation results with relatively poor accuracy.
Failures are mainly due to standard segmentation issues when
object contours do not match color intensity shifts, or are delimited by  thin contours (Figure \ref{fig:errors}(top)).
Contrary to most methods, SphSPS explicitly considers a contour term in its model, preventing from high failures on this aspect.
Finally, a lower accuracy may be due to the ground truth segmentation of very thin objects that cannot be captured by the used superpixel scale (Figure \ref{fig:errors}(bottom)).
}

\newcommand{\ww}{0.235\textwidth}
\newcommand{\ppp}{0.1175\textwidth}
\begin{figure*}[ht!]
{\scriptsize
\begin{tabular}{@{\hspace{1mm}}c@{\hspace{1mm}}c@{\hspace{1mm}}c@{\hspace{3mm}}c@{\hspace{1mm}}c@{\hspace{1mm}}c@{\hspace{0mm}}}
\multicolumn{3}{c}{\small \textbf{Planar methods}} &
\multicolumn{3}{c}{\small \textbf{Spherical methods}} \\[1.5ex]
\includegraphics[width=\ww,height=\ppp]{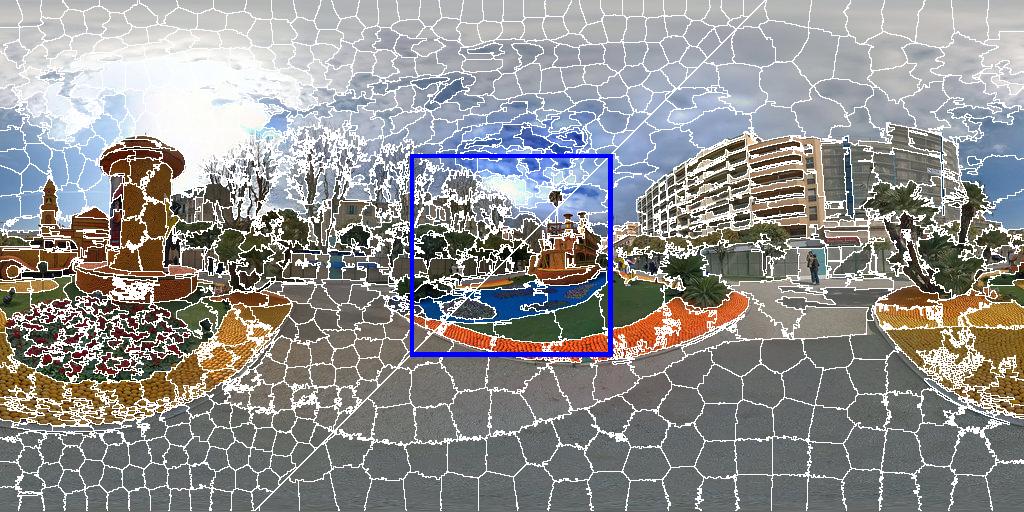}&
\includegraphics[width=\ppp,height=\ppp]{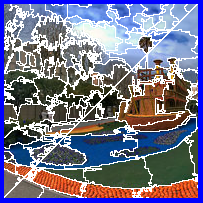}&
\includegraphics[width=\ppp,height=\ppp]{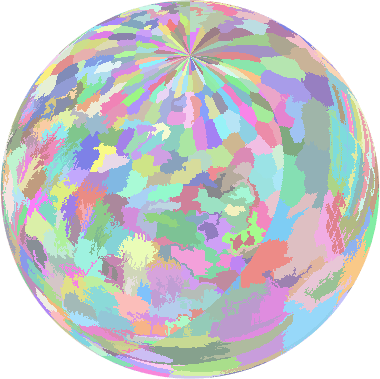}&
\includegraphics[width=\ww,height=\ppp]{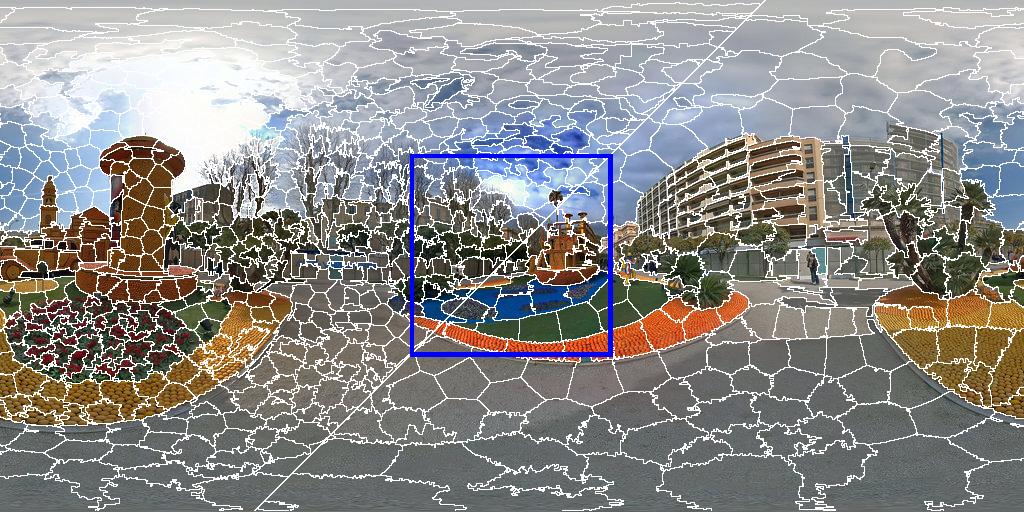}&
\includegraphics[width=\ppp,height=\ppp]{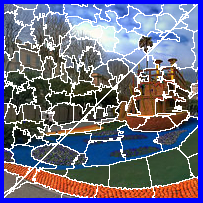}&
\includegraphics[width=\ppp,height=\ppp]{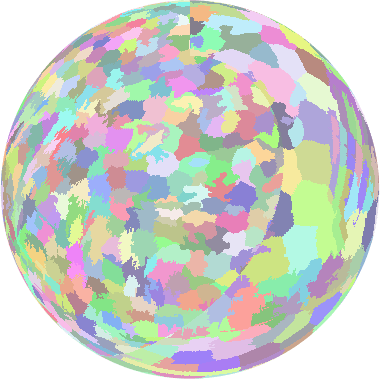}\\[-0.5ex]
\multicolumn{3}{c}{LSC \cite{chen2017}}&
\multicolumn{3}{c}{{SphSLIC-Euc \cite{zhao2018}}} \\[0.75ex]
\includegraphics[width=\ww,height=\ppp]{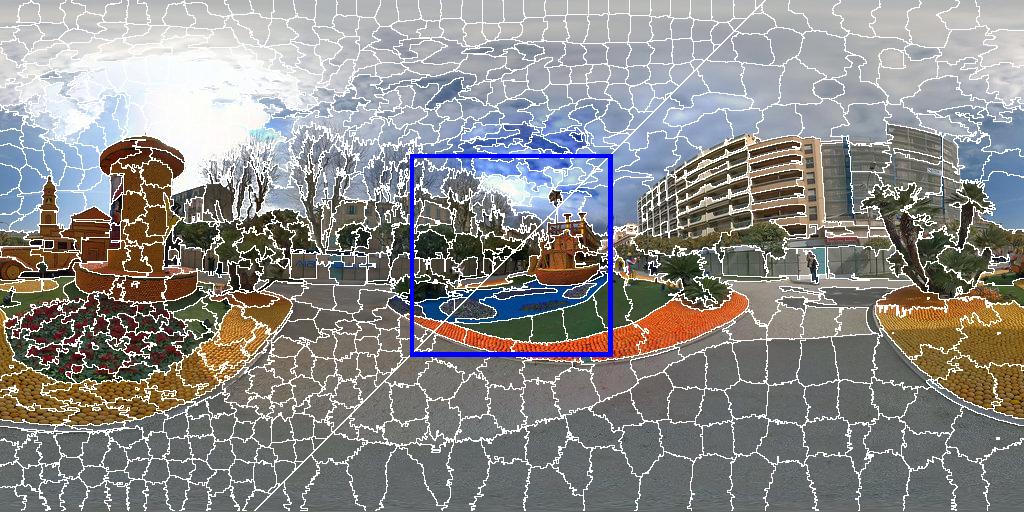}&
\includegraphics[width=\ppp,height=\ppp]{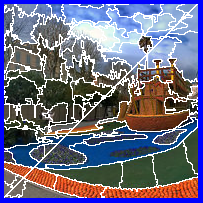}&
\includegraphics[width=\ppp,height=\ppp]{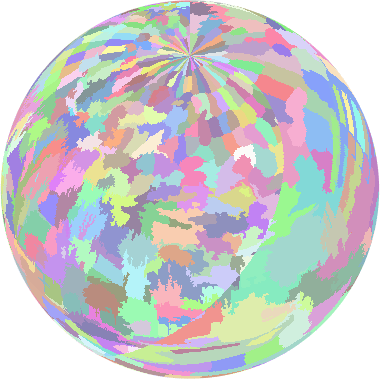}&
\includegraphics[width=\ww,height=\ppp]{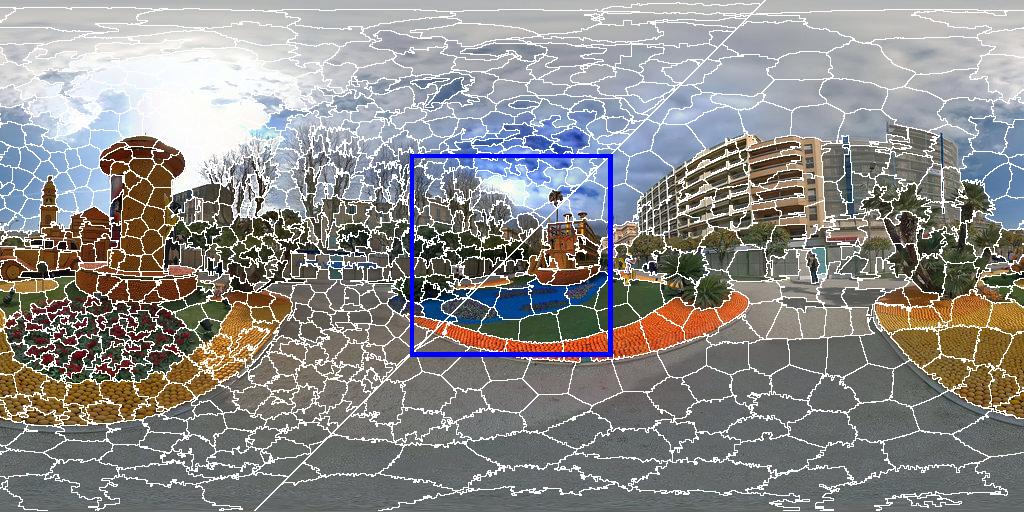}&
\includegraphics[width=\ppp,height=\ppp]{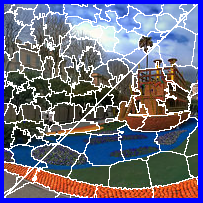}&
\includegraphics[width=\ppp,height=\ppp]{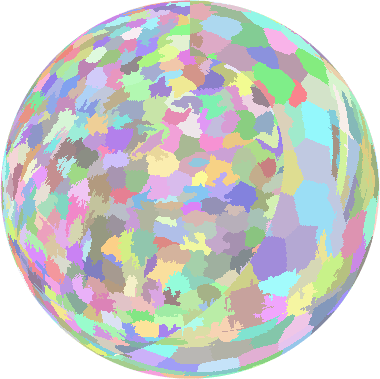}\\[-0.5ex]
\multicolumn{3}{c}{SNIC \cite{achanta2017superpixels}}&
\multicolumn{3}{c}{{SphSLIC-Cos \cite{zhao2018}}} \\[0.75ex]
\includegraphics[width=\ww,height=\ppp]{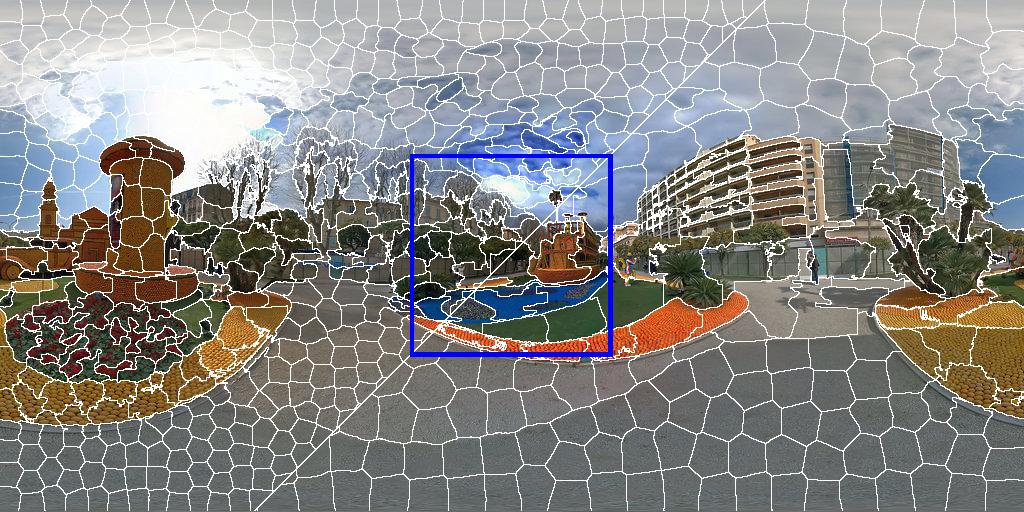}&
\includegraphics[width=\ppp,height=\ppp]{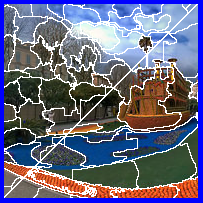}&
\includegraphics[width=\ppp,height=\ppp]{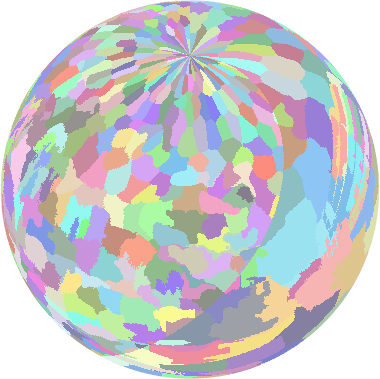}&
\includegraphics[width=\ww,height=\ppp]{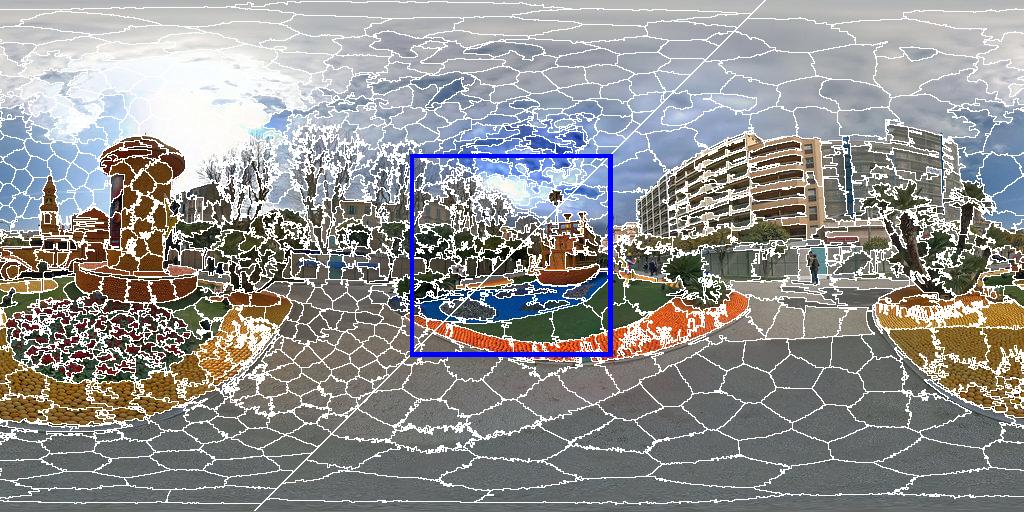}&
\includegraphics[width=\ppp,height=\ppp]{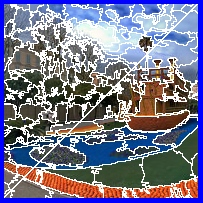}&
\includegraphics[width=\ppp,height=\ppp]{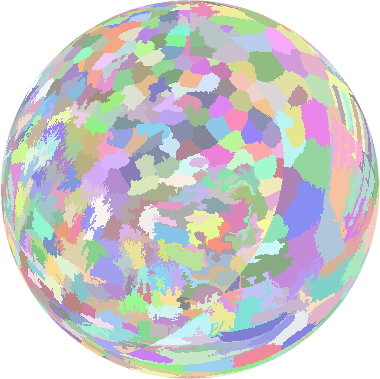}\\[-0.5ex]
\multicolumn{3}{c}{SCALP \cite{giraud2018_scalp}} &
\multicolumn{3}{c}{SphLSC \cite{chen2017,zhao2018}} \\[0.75ex]
\includegraphics[width=\ww,height=\ppp]{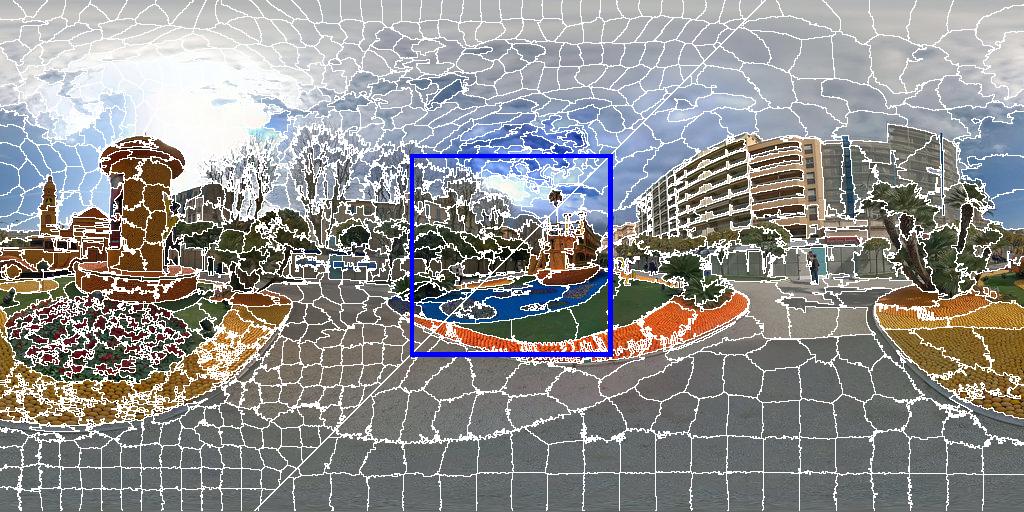}&
\includegraphics[width=\ppp,height=\ppp]{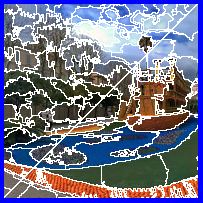}&
\includegraphics[width=\ppp,height=\ppp]{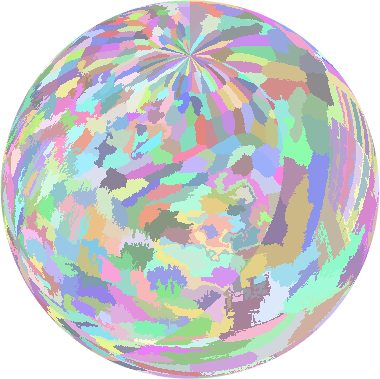}&
\includegraphics[width=\ww,height=\ppp]{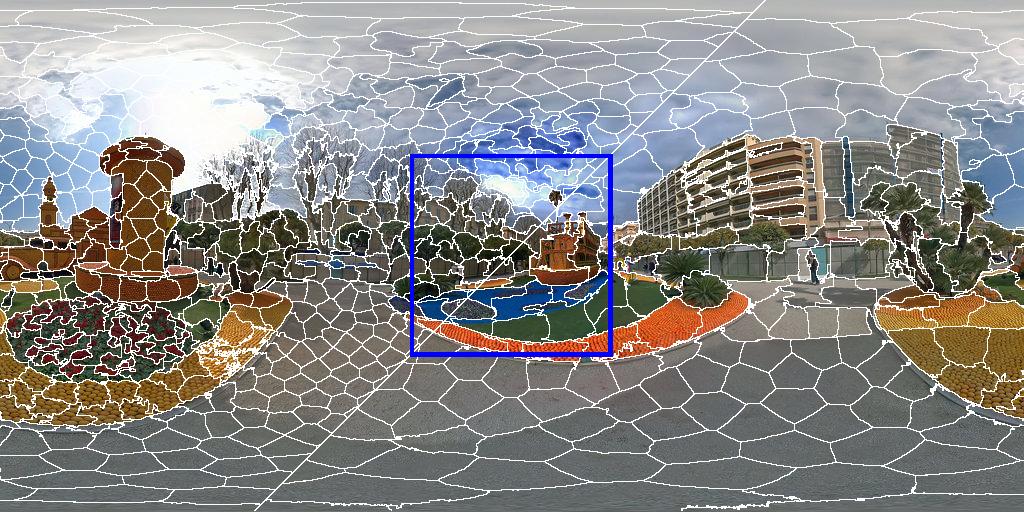}&
\includegraphics[width=\ppp,height=\ppp]{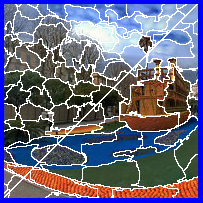}&
\includegraphics[width=\ppp,height=\ppp]{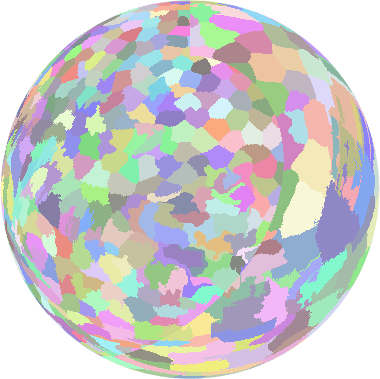}\\[-0.5ex]
\multicolumn{3}{c}{GMMSP \cite{Ban18}} &
\multicolumn{3}{c}{\textbf{SphSPS}}\\[2ex]
%
% \rotatebox{90}{\hspace{0.35cm} (Euclidean)}&
\includegraphics[width=\ww,height=\ppp]{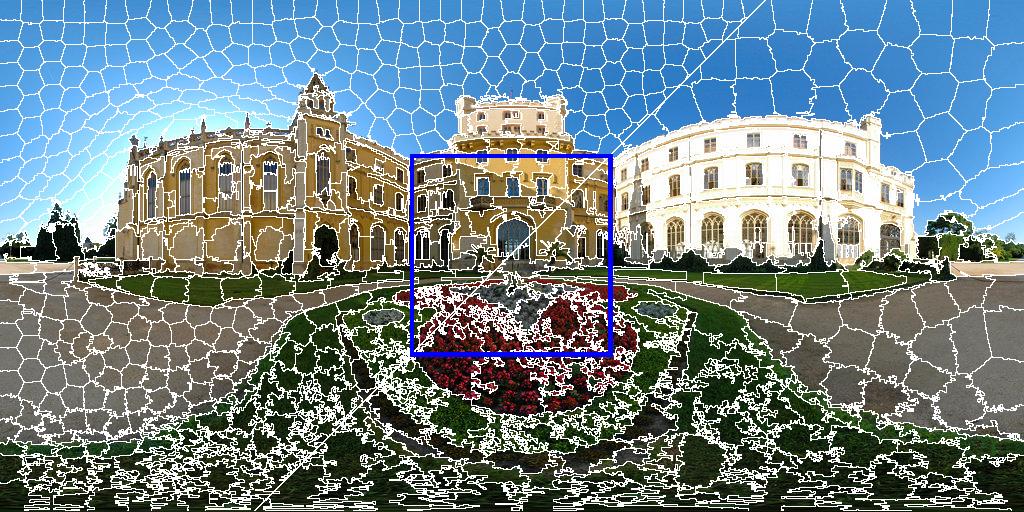}&
\includegraphics[width=\ppp,height=\ppp]{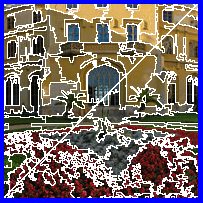}&
\includegraphics[width=\ppp,height=\ppp]{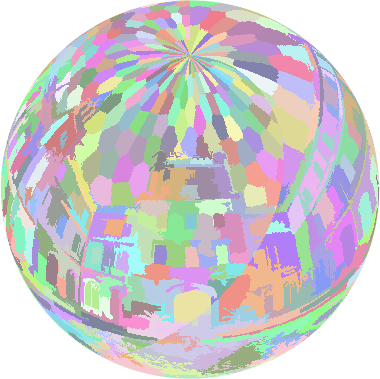}&
% \rotatebox{90}{\hspace{0.35cm} (Euclidean)}&
\includegraphics[width=\ww,height=\ppp]{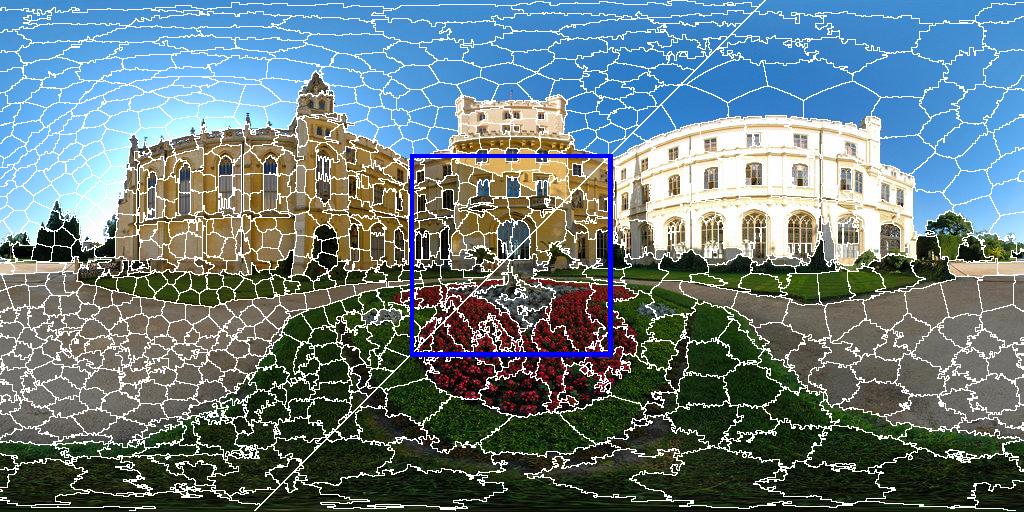}&
\includegraphics[width=\ppp,height=\ppp]{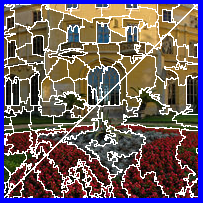}&
\includegraphics[width=\ppp,height=\ppp]{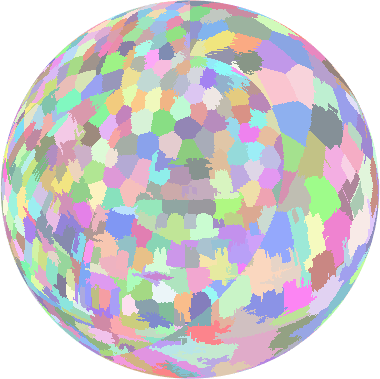}\\[-0.5ex]
\multicolumn{3}{c}{LSC \cite{chen2017}}&
\multicolumn{3}{c}{{SphSLIC-Euc \cite{zhao2018}}} \\[0.75ex]
\includegraphics[width=\ww,height=\ppp]{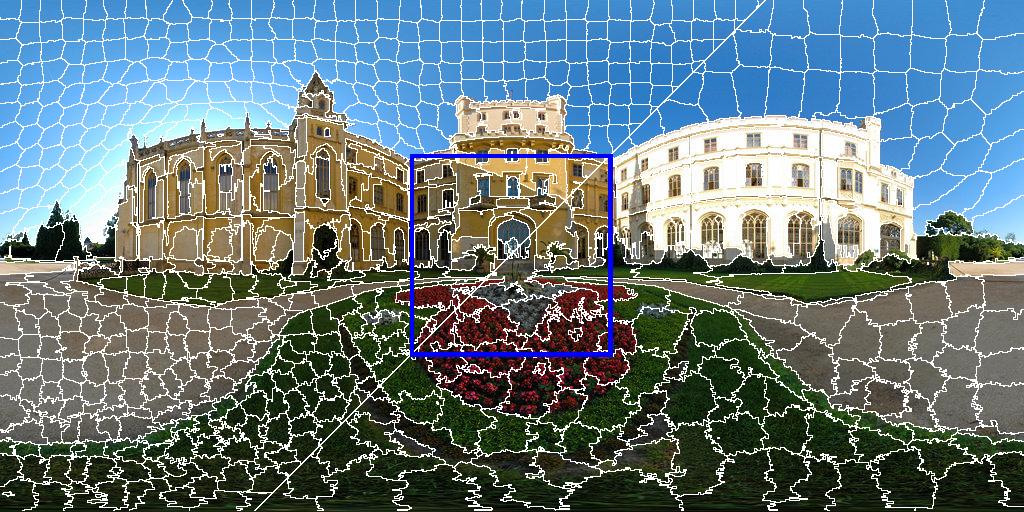}&
\includegraphics[width=\ppp,height=\ppp]{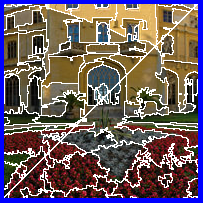}&
\includegraphics[width=\ppp,height=\ppp]{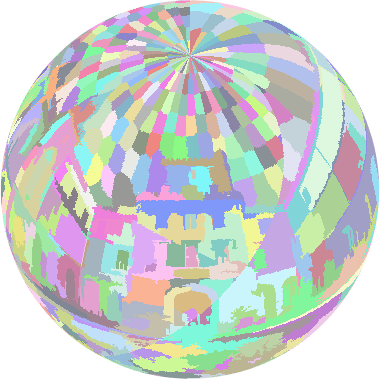}&
\includegraphics[width=\ww,height=\ppp]{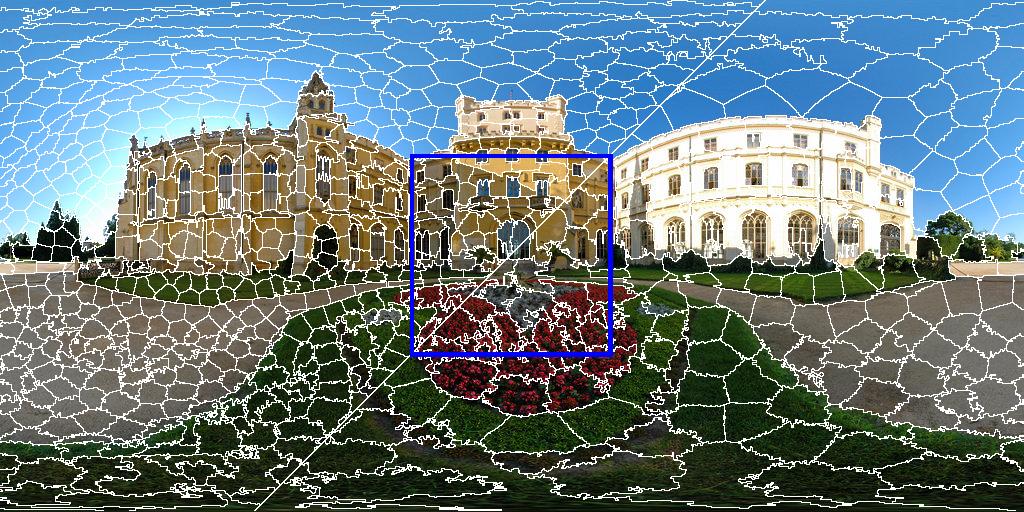}&
\includegraphics[width=\ppp,height=\ppp]{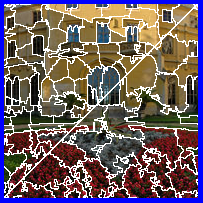}&
\includegraphics[width=\ppp,height=\ppp]{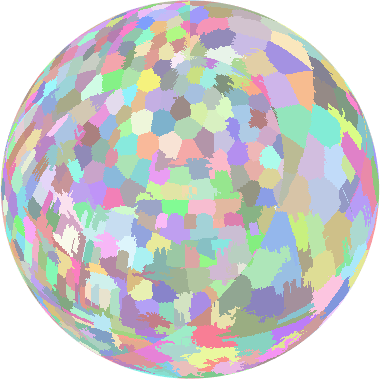}\\[-0.5ex]
\multicolumn{3}{c}{SNIC \cite{achanta2017superpixels}}&
\multicolumn{3}{c}{{SphSLIC-Cos \cite{zhao2018}}} \\[0.75ex]
\includegraphics[width=\ww,height=\ppp]{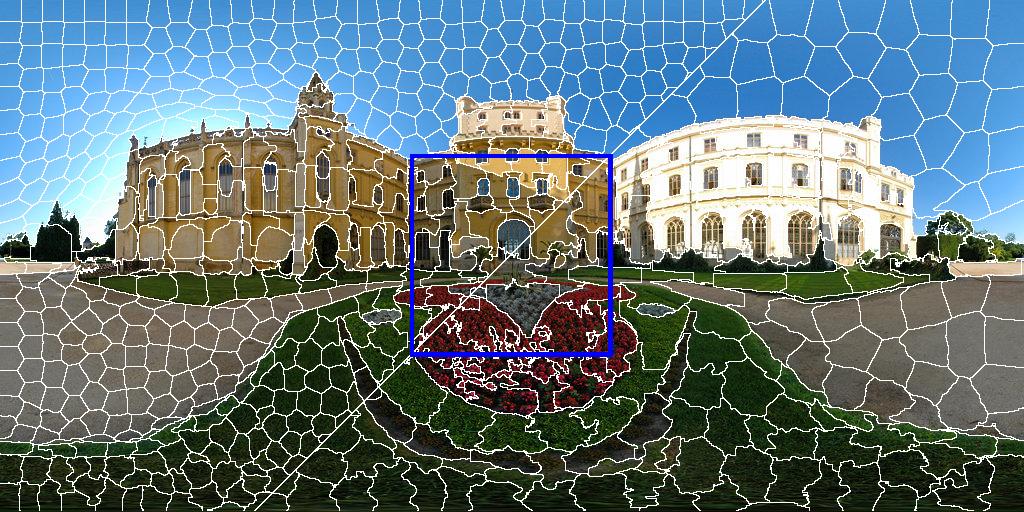}&
\includegraphics[width=\ppp,height=\ppp]{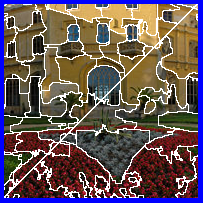}&
\includegraphics[width=\ppp,height=\ppp]{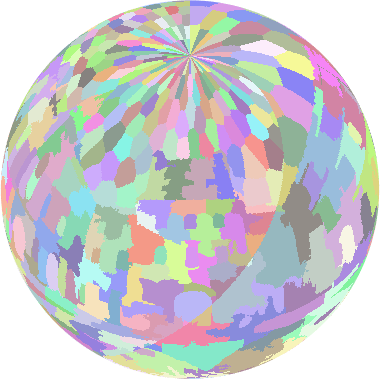}&
\includegraphics[width=\ww,height=\ppp]{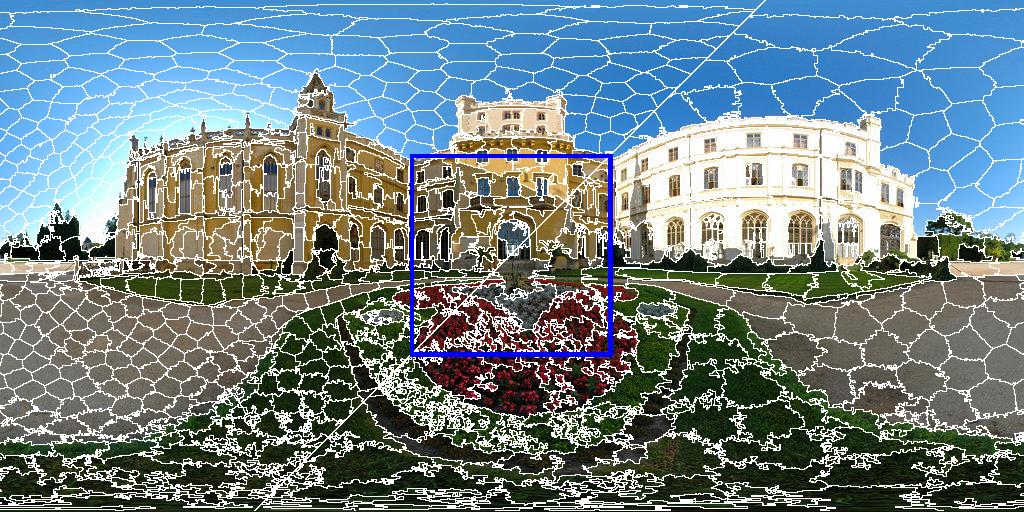}&
\includegraphics[width=\ppp,height=\ppp]{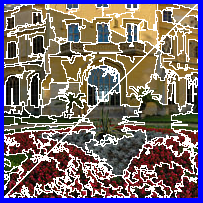}&
\includegraphics[width=\ppp,height=\ppp]{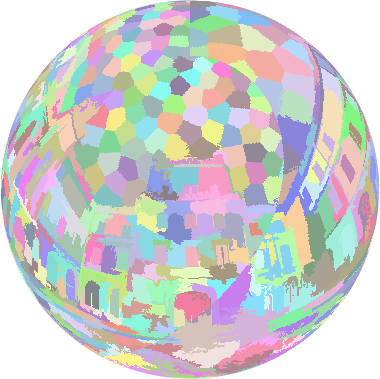}\\[-0.5ex]
\multicolumn{3}{c}{SCALP \cite{giraud2018_scalp}} &
\multicolumn{3}{c}{SphLSC \cite{chen2017,zhao2018}}\\[0.75ex]
\includegraphics[width=\ww,height=\ppp]{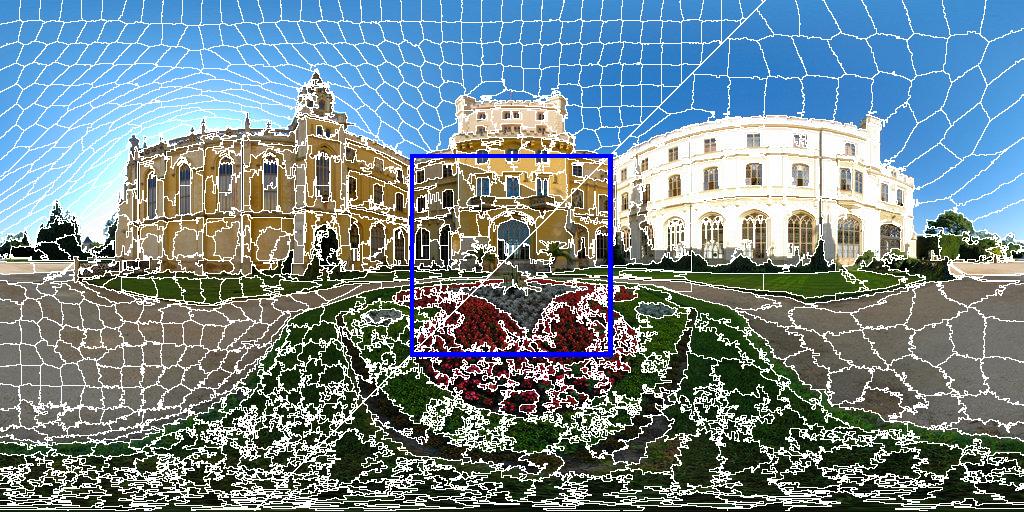}&
\includegraphics[width=\ppp,height=\ppp]{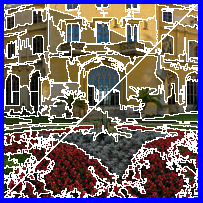}&
\includegraphics[width=\ppp,height=\ppp]{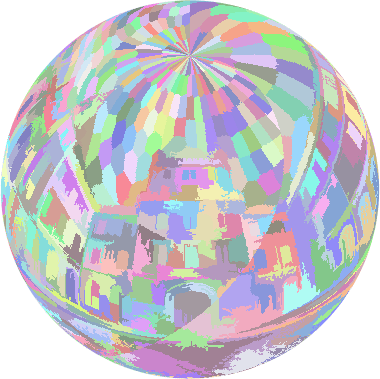}&
\includegraphics[width=\ww,height=\ppp]{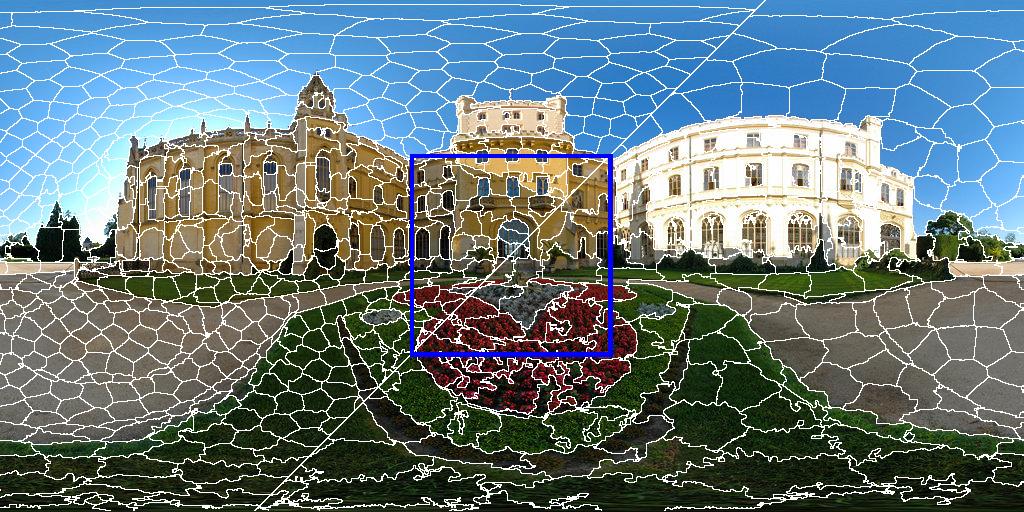}&
\includegraphics[width=\ppp,height=\ppp]{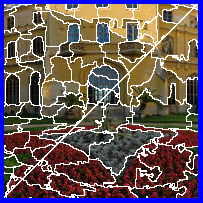}&
\includegraphics[width=\ppp,height=\ppp]{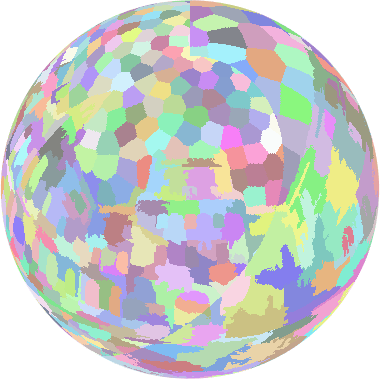}\\[-0.5ex]
\multicolumn{3}{c}{GMMSP \cite{Ban18}} &
\multicolumn{3}{c}{\textbf{SphSPS}}\\
\end{tabular}
}
\caption{
Visual comparison between SphSPS and the
best planar and spherical (underlined) state-of-the-art
methods on PSD images, for two superpixel numbers $K=1200$ (top-left) and $K=400$ (bottom right).
  SphSPS produces regular spherical superpixels with smooth boundaries
  that adhere well to the image contours}%
\label{fig:sps_soa_img_1}
\end{figure*}

\begin{figure*}[ht!]
\centering
{\scriptsize
\newcommand{\pppr}{0.115\textwidth}
\begin{tabular}{@{\hspace{1mm}}c@{\hspace{1mm}}c@{\hspace{1mm}}c@{\hspace{3mm}}c@{\hspace{1mm}}c@{\hspace{1mm}}c@{\hspace{0mm}}}
\multicolumn{3}{c}{\small \textbf{Planar methods}} &
\multicolumn{3}{c}{\small \textbf{Spherical methods}} \\[0.5ex]
\includegraphics[width=\ww,height=\pppr]{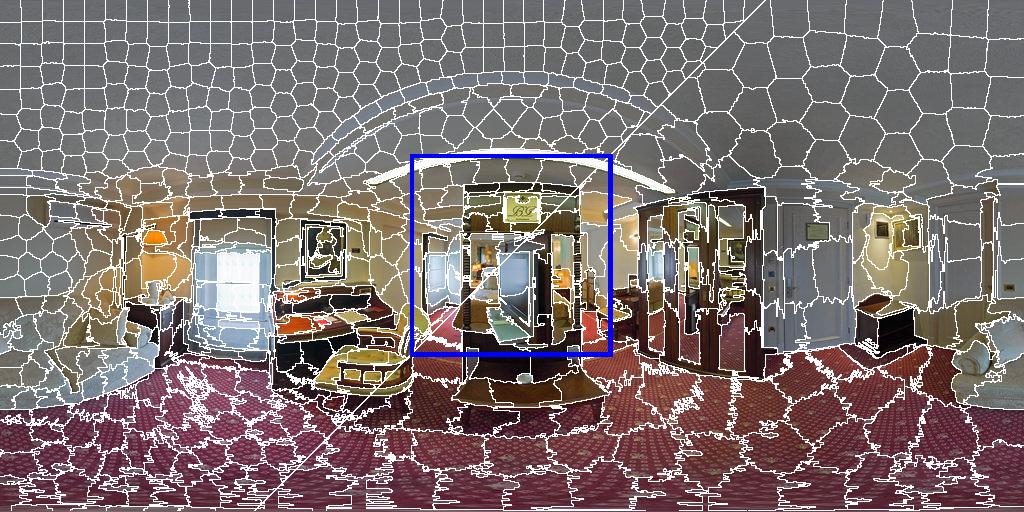}&
\includegraphics[width=\ppp,height=\pppr]{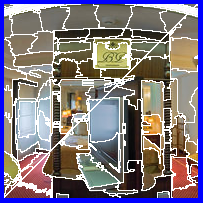}&
\includegraphics[width=\ppp,height=\pppr]{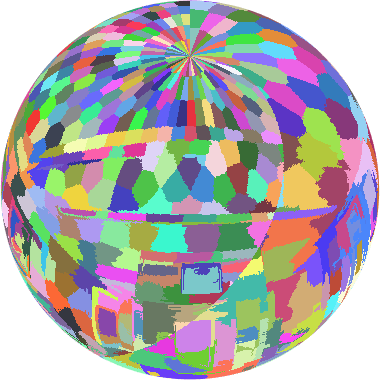}&
% \rotatebox{90}{\hspace{0.35cm} (Euclidean)}&
\includegraphics[width=\ww,height=\pppr]{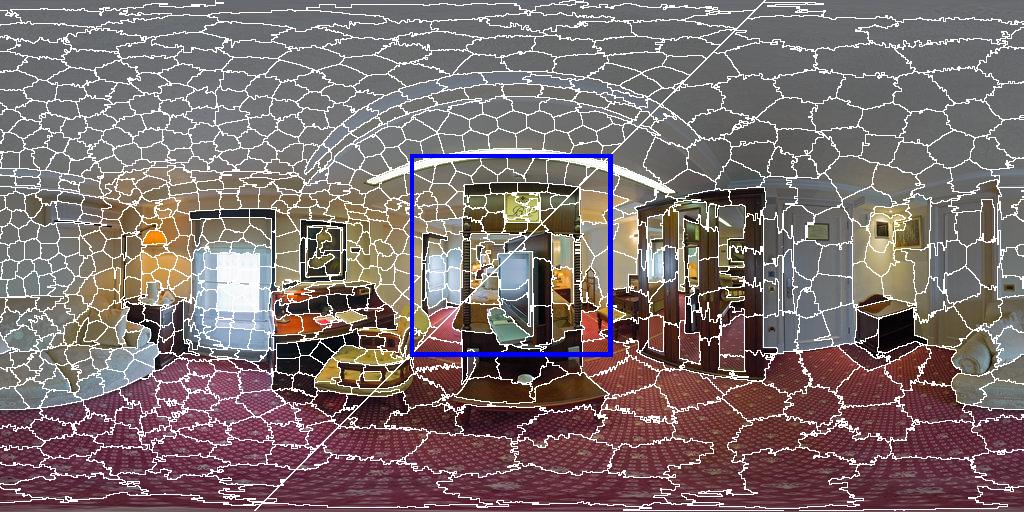}&
\includegraphics[width=\ppp,height=\pppr]{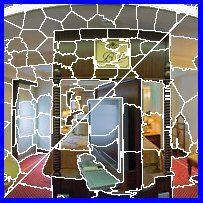}&
\includegraphics[width=\ppp,height=\pppr]{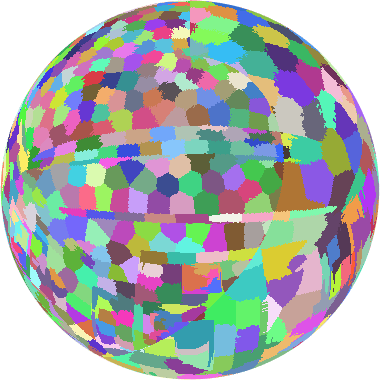}\\[-0.5ex]
\multicolumn{3}{c}{LSC \cite{chen2017}}&
\multicolumn{3}{c}{{SphSLIC-Euc \cite{zhao2018}}} \\[0.75ex]
\includegraphics[width=\ww,height=\pppr]{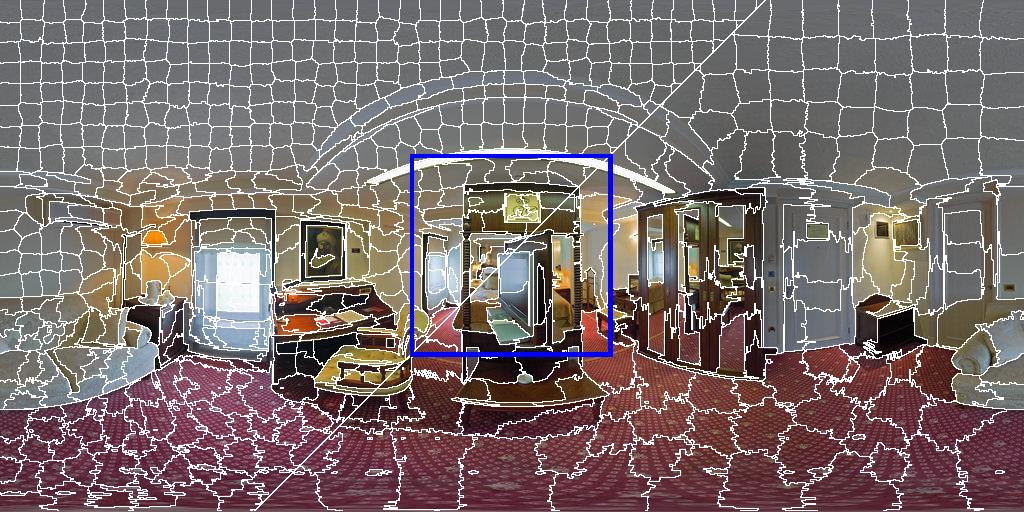}&
\includegraphics[width=\ppp,height=\pppr]{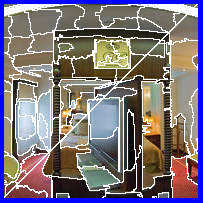}&
\includegraphics[width=\ppp,height=\pppr]{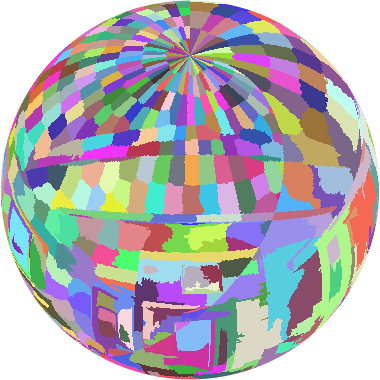}&
% \rotatebox{90}{\hspace{0.35cm} (Cosine-opt)}&
\includegraphics[width=\ww,height=\pppr]{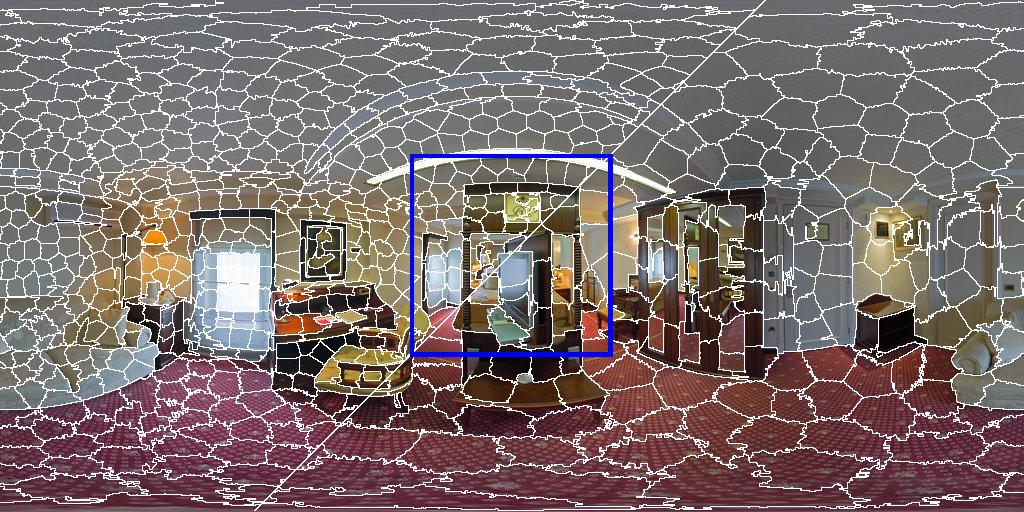}&
\includegraphics[width=\ppp,height=\pppr]{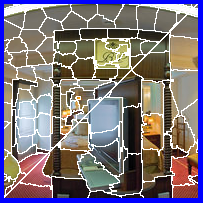}&
\includegraphics[width=\ppp,height=\pppr]{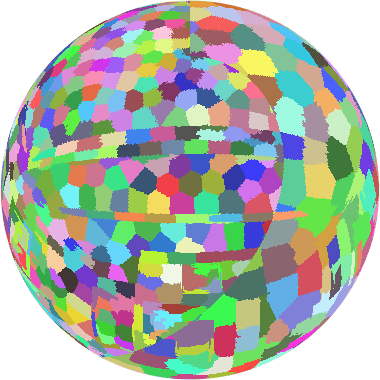}\\[-0.5ex]
\multicolumn{3}{c}{SNIC \cite{achanta2017superpixels}}&
\multicolumn{3}{c}{{SphSLIC-Cos \cite{zhao2018}}} \\[0.75ex]
\includegraphics[width=\ww,height=\pppr]{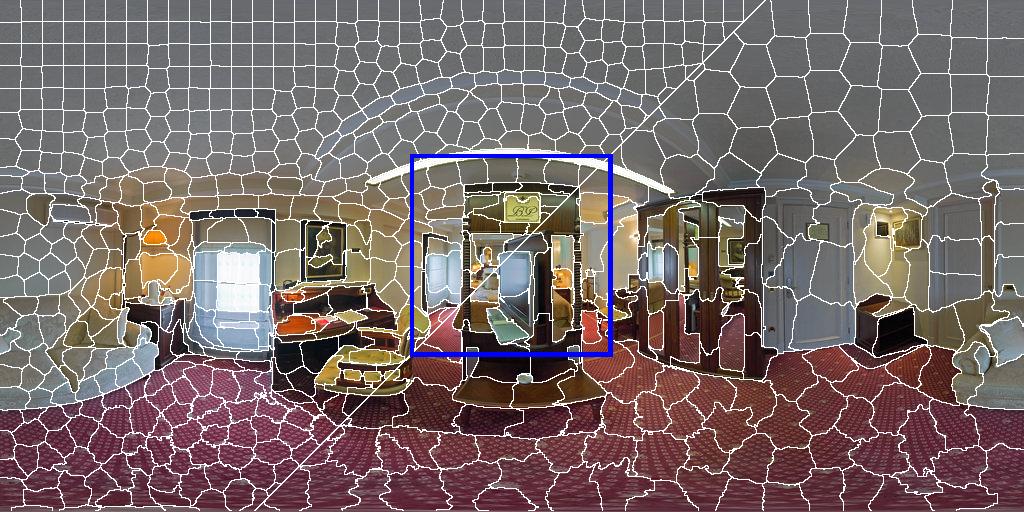}&
\includegraphics[width=\ppp,height=\pppr]{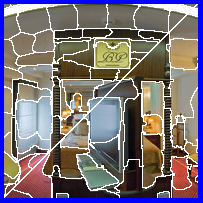}&
\includegraphics[width=\ppp,height=\pppr]{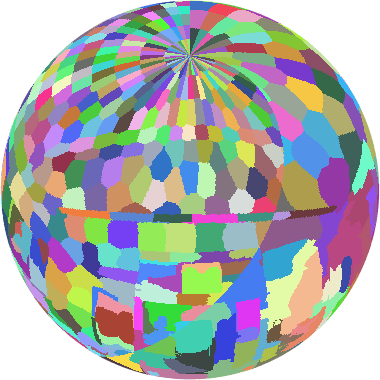}&
\includegraphics[width=\ww,height=\pppr]{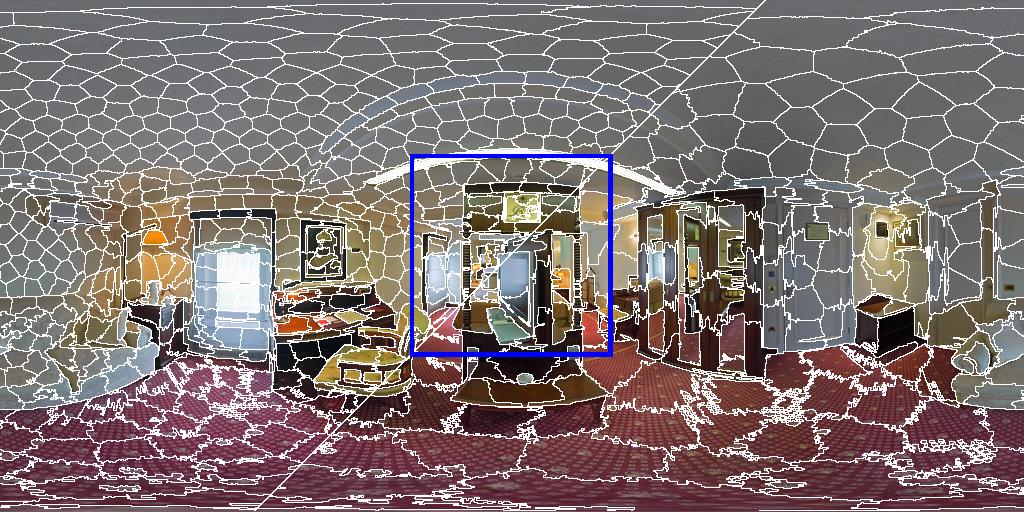}&
\includegraphics[width=\ppp,height=\pppr]{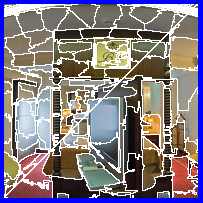}&
\includegraphics[width=\ppp,height=\pppr]{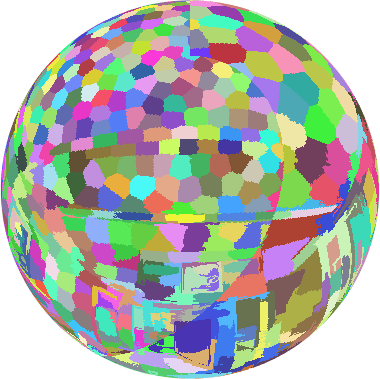}\\[-0.5ex]
\multicolumn{3}{c}{SCALP \cite{giraud2018_scalp}}&
\multicolumn{3}{c}{SphLSC \cite{chen2017,zhao2018}} \\[0.75ex]
\includegraphics[width=\ww,height=\pppr]{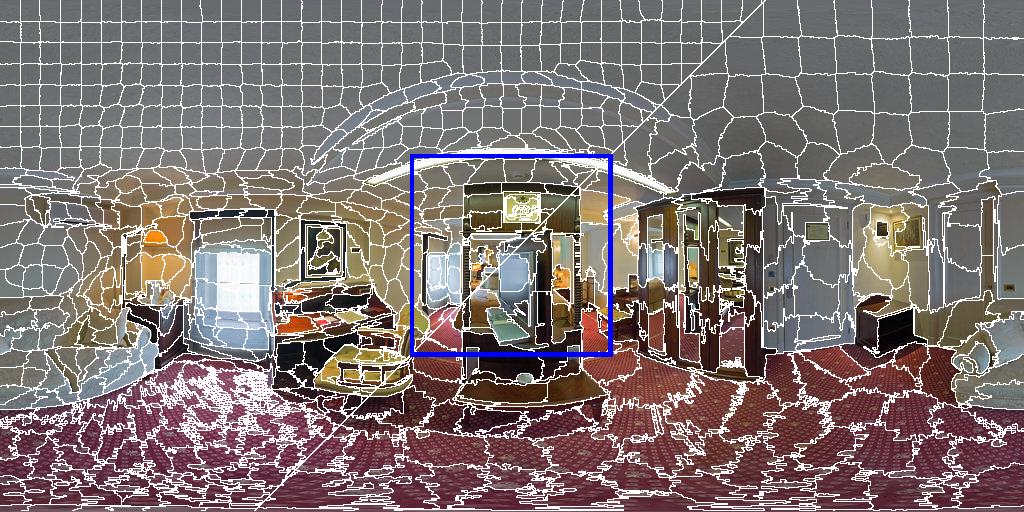}&
\includegraphics[width=\ppp,height=\pppr]{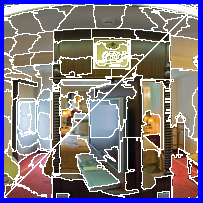}&
\includegraphics[width=\ppp,height=\pppr]{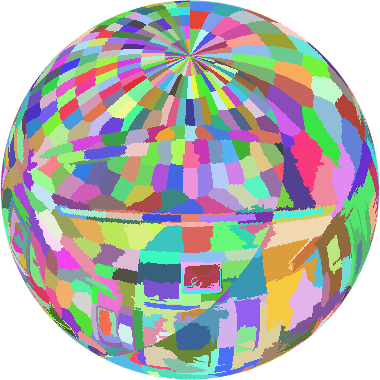}&
\includegraphics[width=\ww,height=\pppr]{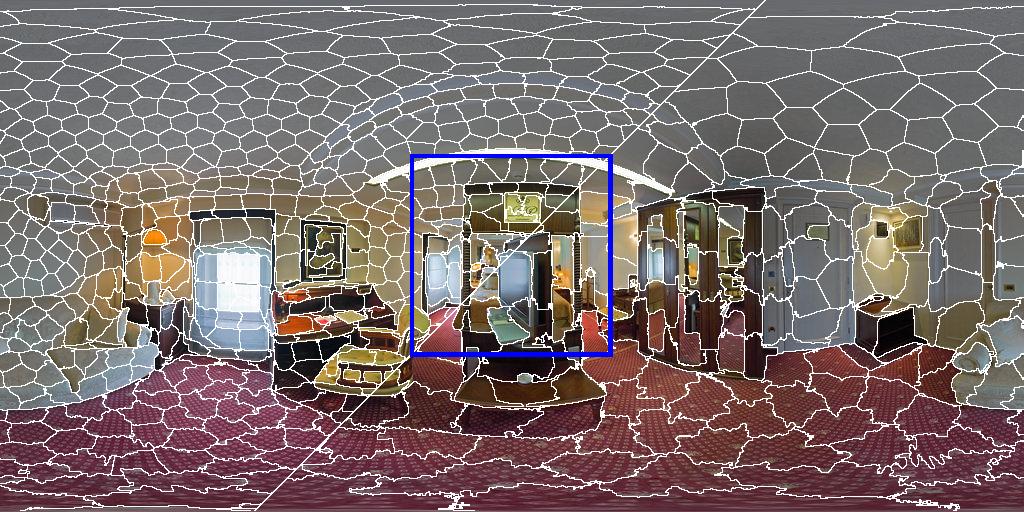}&
\includegraphics[width=\ppp,height=\pppr]{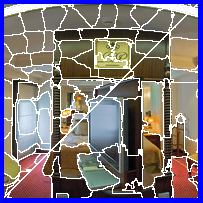}&
\includegraphics[width=\ppp,height=\pppr]{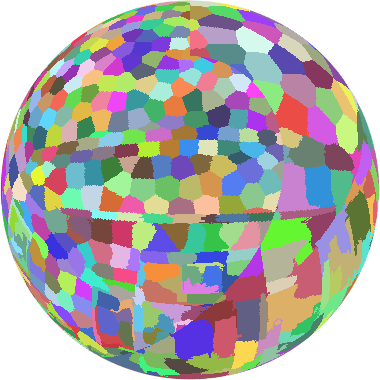}\\[-0.5ex]
\multicolumn{3}{c}{GMMSP \cite{Ban18}}&
\multicolumn{3}{c}{{\textbf{SphSPS}}} \\[2ex]
\includegraphics[width=\ww,height=\pppr]{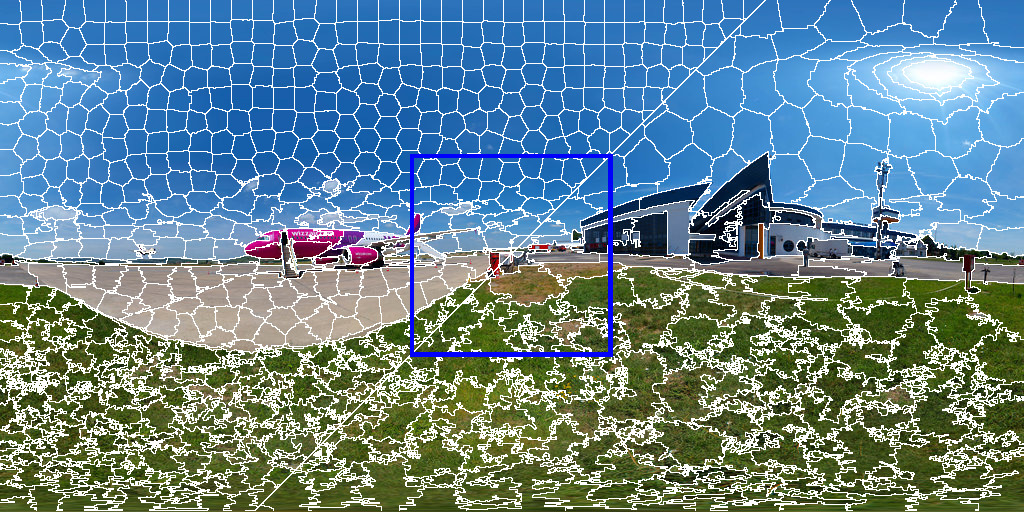}&
\includegraphics[width=\ppp,height=\pppr]{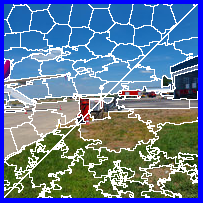}&
\includegraphics[width=\ppp,height=\pppr]{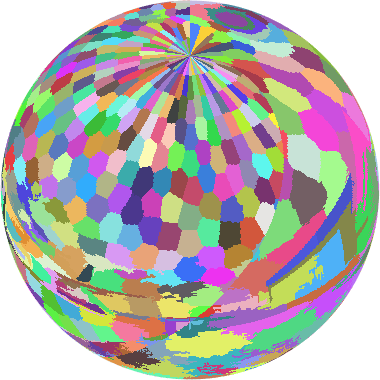}&
% \rotatebox{90}{\hspace{0.35cm} (Euclidean)}&
\includegraphics[width=\ww,height=\pppr]{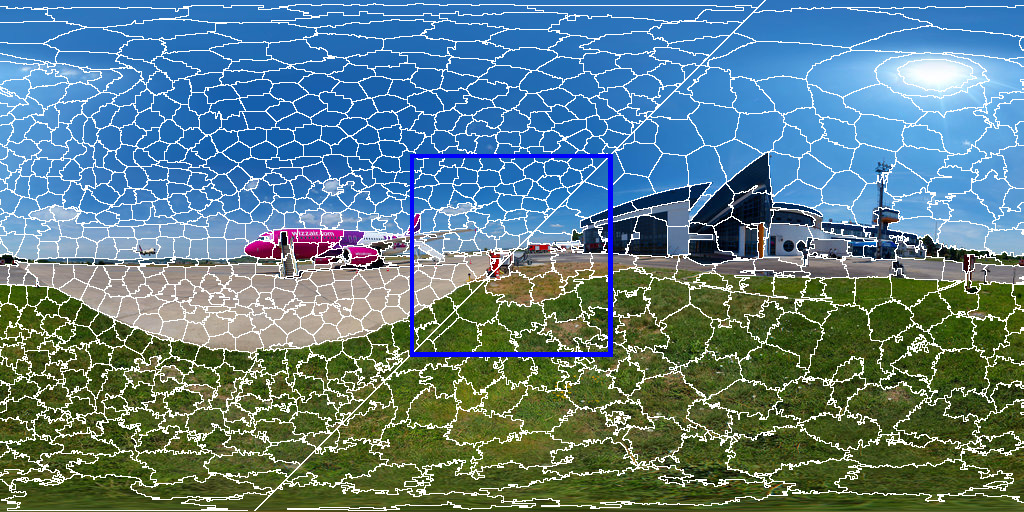}&
\includegraphics[width=\ppp,height=\pppr]{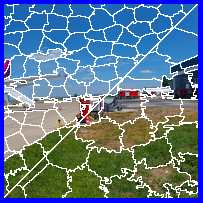}&
\includegraphics[width=\ppp,height=\pppr]{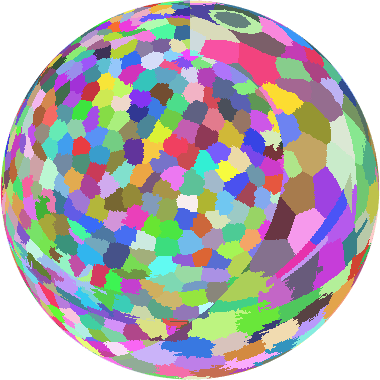}\\[-0.5ex]
\multicolumn{3}{c}{LSC \cite{chen2017}}&
\multicolumn{3}{c}{{SphSLIC-Euc \cite{zhao2018}}} \\[0.75ex]
\includegraphics[width=\ww,height=\pppr]{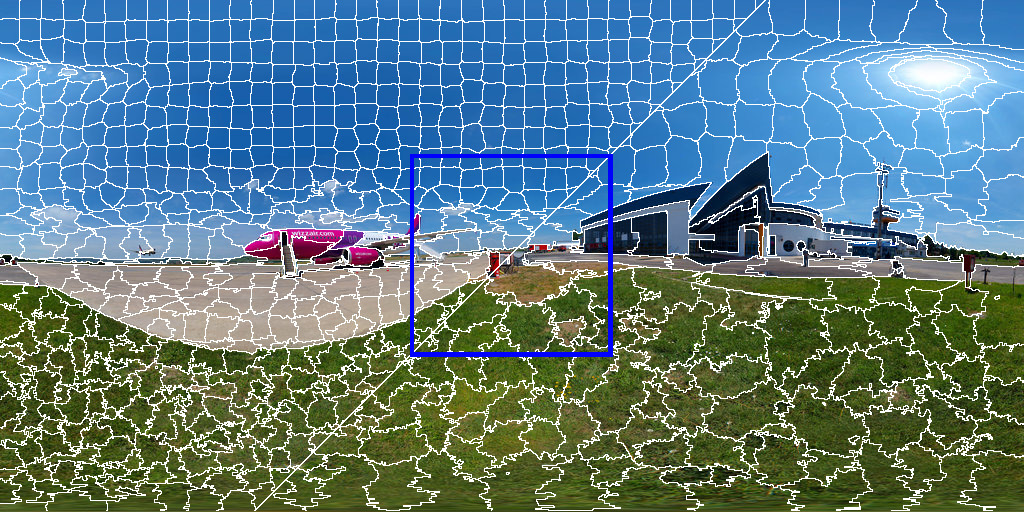}&
\includegraphics[width=\ppp,height=\pppr]{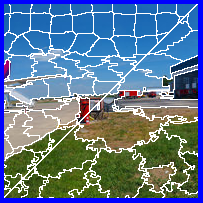}&
\includegraphics[width=\ppp,height=\pppr]{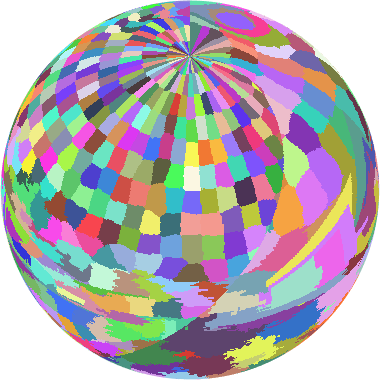}&
% \rotatebox{90}{\hspace{0.35cm} (Cosine-opt)}&
\includegraphics[width=\ww,height=\pppr]{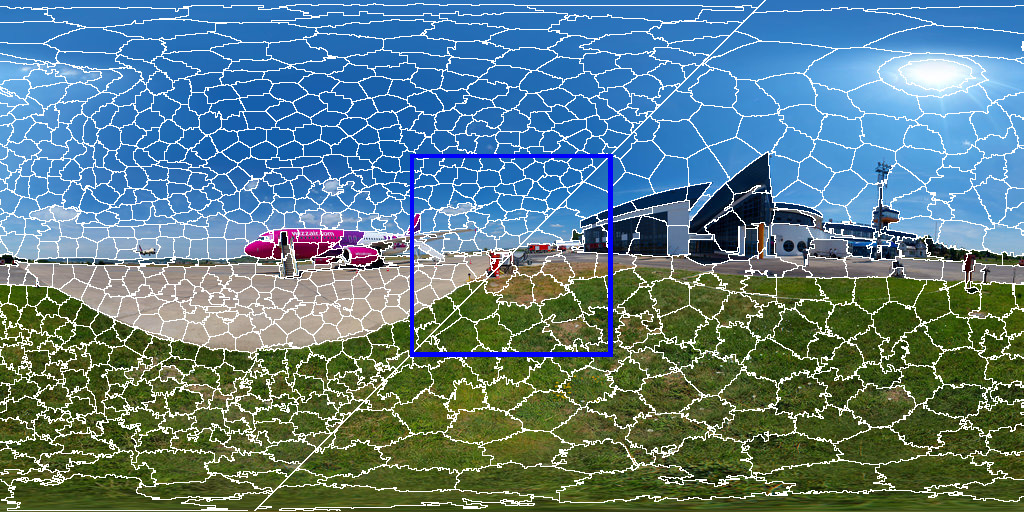}&
\includegraphics[width=\ppp,height=\pppr]{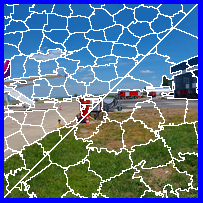}&
\includegraphics[width=\ppp,height=\pppr]{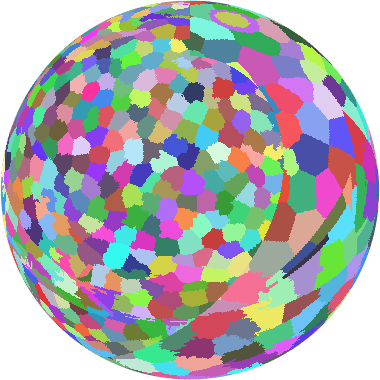}\\[-0.5ex]
\multicolumn{3}{c}{SNIC \cite{achanta2017superpixels}}&
\multicolumn{3}{c}{{SphSLIC-Cos \cite{zhao2018}}} \\[0.75ex]
\includegraphics[width=\ww,height=\pppr]{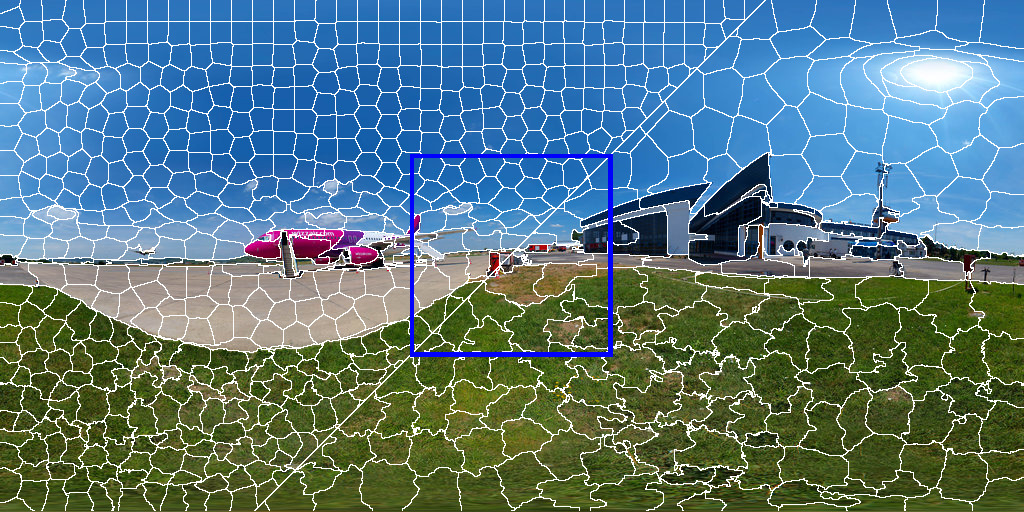}&
\includegraphics[width=\ppp,height=\pppr]{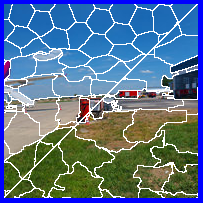}&
\includegraphics[width=\ppp,height=\pppr]{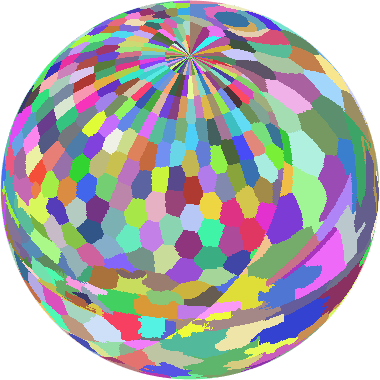}&
\includegraphics[width=\ww,height=\pppr]{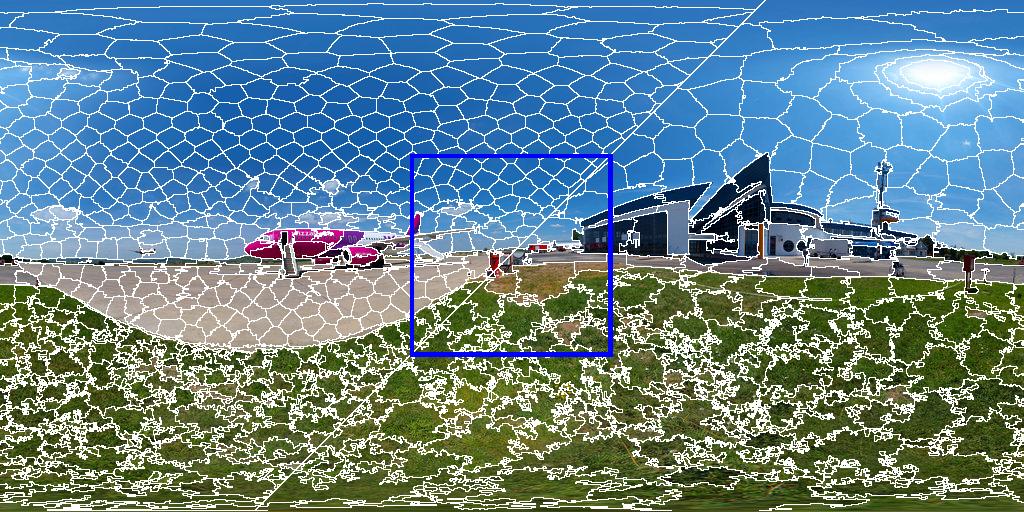}&
\includegraphics[width=\ppp,height=\pppr]{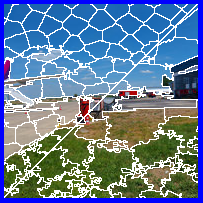}&
\includegraphics[width=\ppp,height=\pppr]{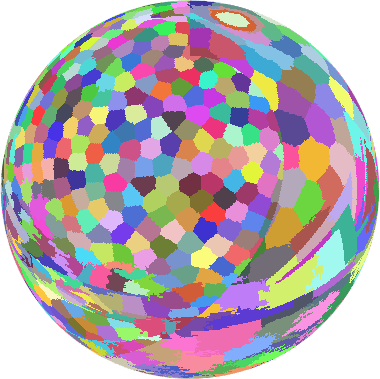}\\[-0.5ex]
\multicolumn{3}{c}{SCALP \cite{giraud2018_scalp}} &
\multicolumn{3}{c}{SphLSC \cite{chen2017,zhao2018}}\\[0.75ex]
\includegraphics[width=\ww,height=\pppr]{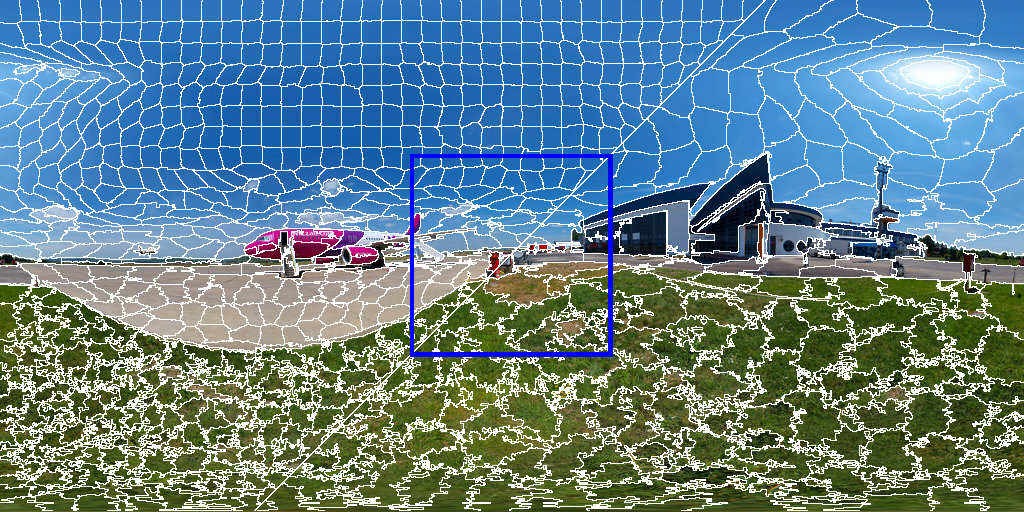}&
\includegraphics[width=\ppp,height=\pppr]{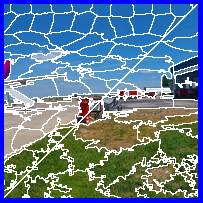}&
\includegraphics[width=\ppp,height=\pppr]{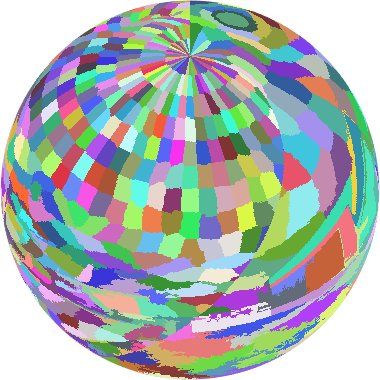}&
\includegraphics[width=\ww,height=\pppr]{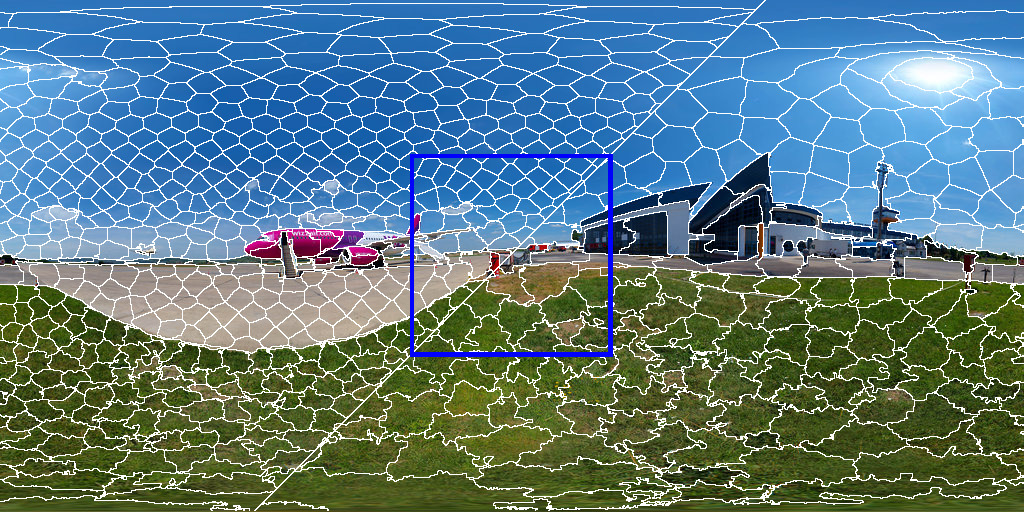}&
\includegraphics[width=\ppp,height=\pppr]{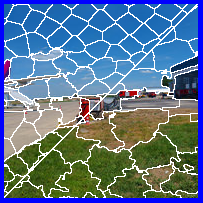}&
\includegraphics[width=\ppp,height=\pppr]{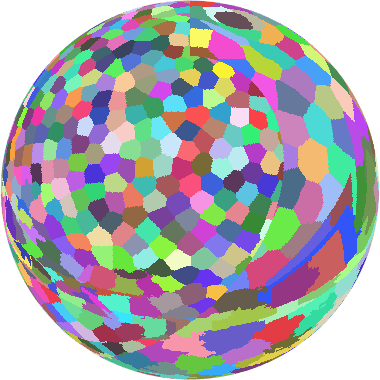}\\[-0.5ex]
\multicolumn{3}{c}{GMMSP \cite{Ban18}}  &
\multicolumn{3}{c}{\textbf{SphSPS}}\\
\end{tabular}}
\caption{Visual comparison between SphSPS and the
best planar and spherical (underlined) state-of-the-art
methods on PSD images, for two superpixel numbers $K=1200$ (top-left) and $K=400$ (bottom right).
  SphSPS produces regular spherical superpixels with smooth boundaries
  that adhere well to the image contours} \vspace{-0.1cm}
\label{fig:sps_soa_img_2}
\end{figure*}

\begin{figure*}[ht!]
{\scriptsize
\begin{tabular}{@{\hspace{1mm}}c@{\hspace{1mm}}c@{\hspace{1mm}}c@{\hspace{3mm}}c@{\hspace{1mm}}c@{\hspace{1mm}}c@{\hspace{0mm}}}
\multicolumn{3}{c}{\small \textbf{Planar methods}} &
\multicolumn{3}{c}{\small \textbf{Spherical methods}} \\[1.5ex]
\includegraphics[width=\ww,height=\ppp]{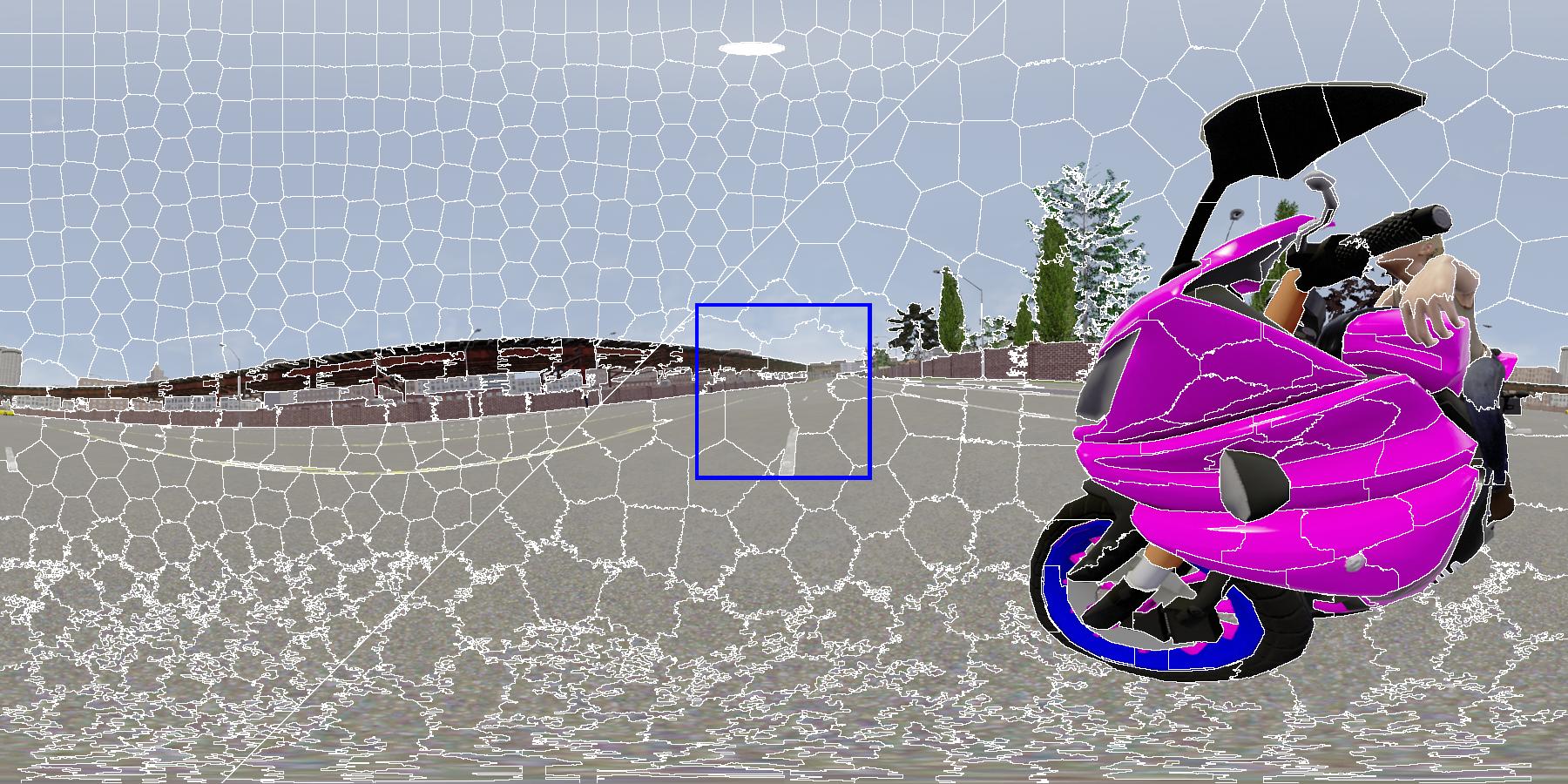}&
\includegraphics[width=\ppp,height=\ppp]{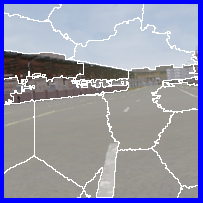}&
\includegraphics[width=\ppp,height=\ppp]{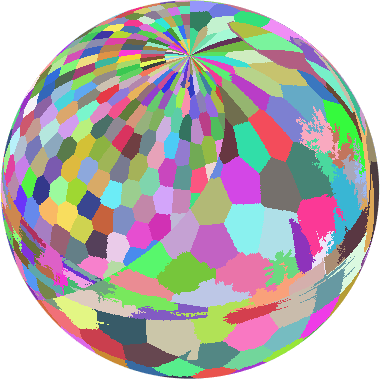}&
\includegraphics[width=\ww,height=\ppp]{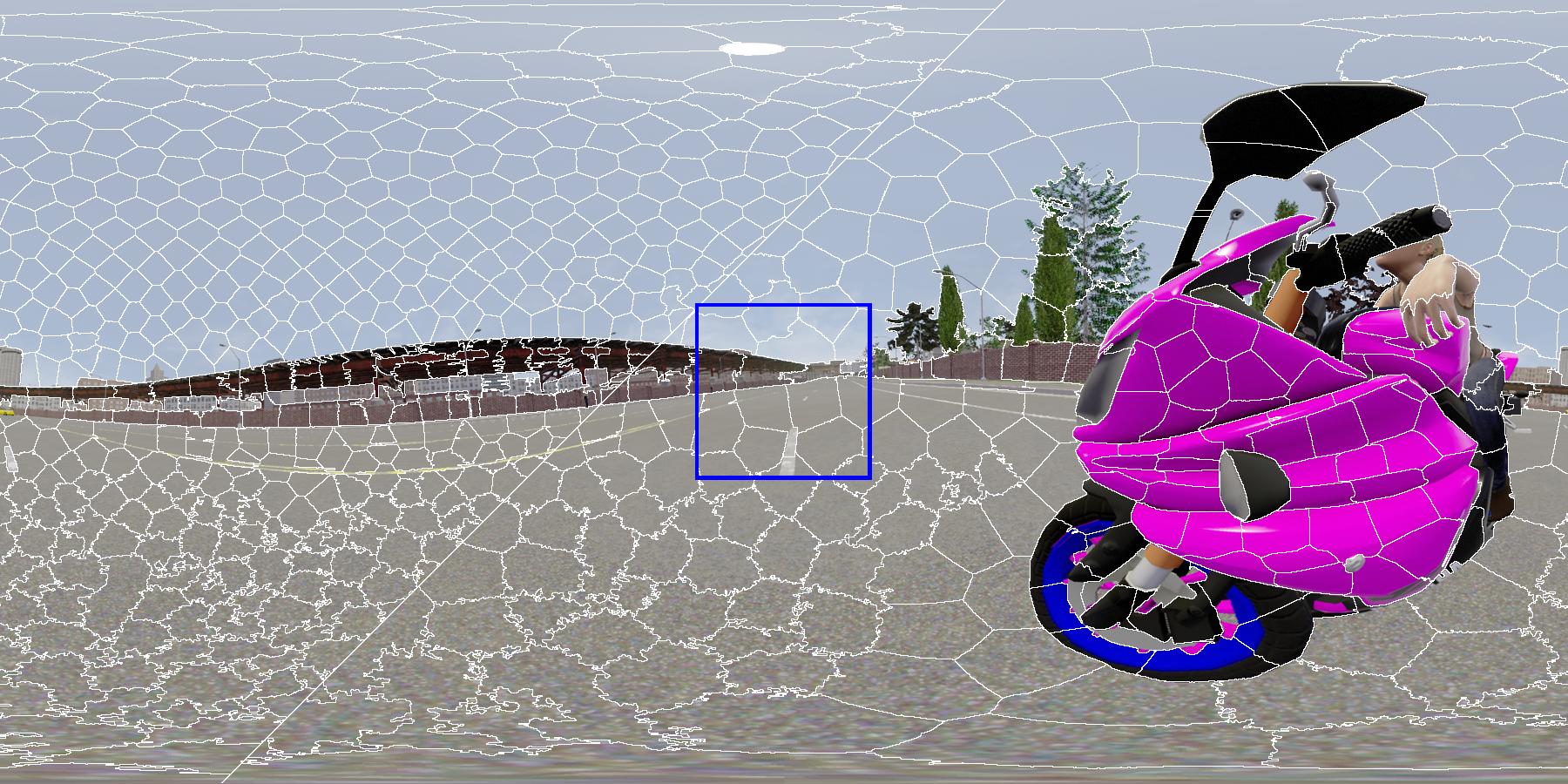}&
\includegraphics[width=\ppp,height=\ppp]{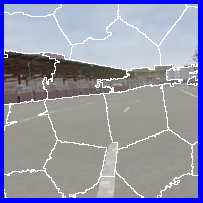}&
\includegraphics[width=\ppp,height=\ppp]{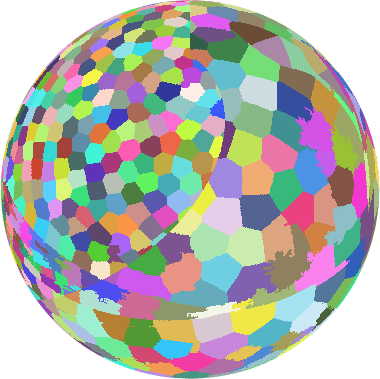}\\[-0.5ex]
\multicolumn{3}{c}{LSC \cite{chen2017}}&
\multicolumn{3}{c}{{SphSLIC-Euc \cite{zhao2018}}} \\[0.75ex]
\includegraphics[width=\ww,height=\ppp]{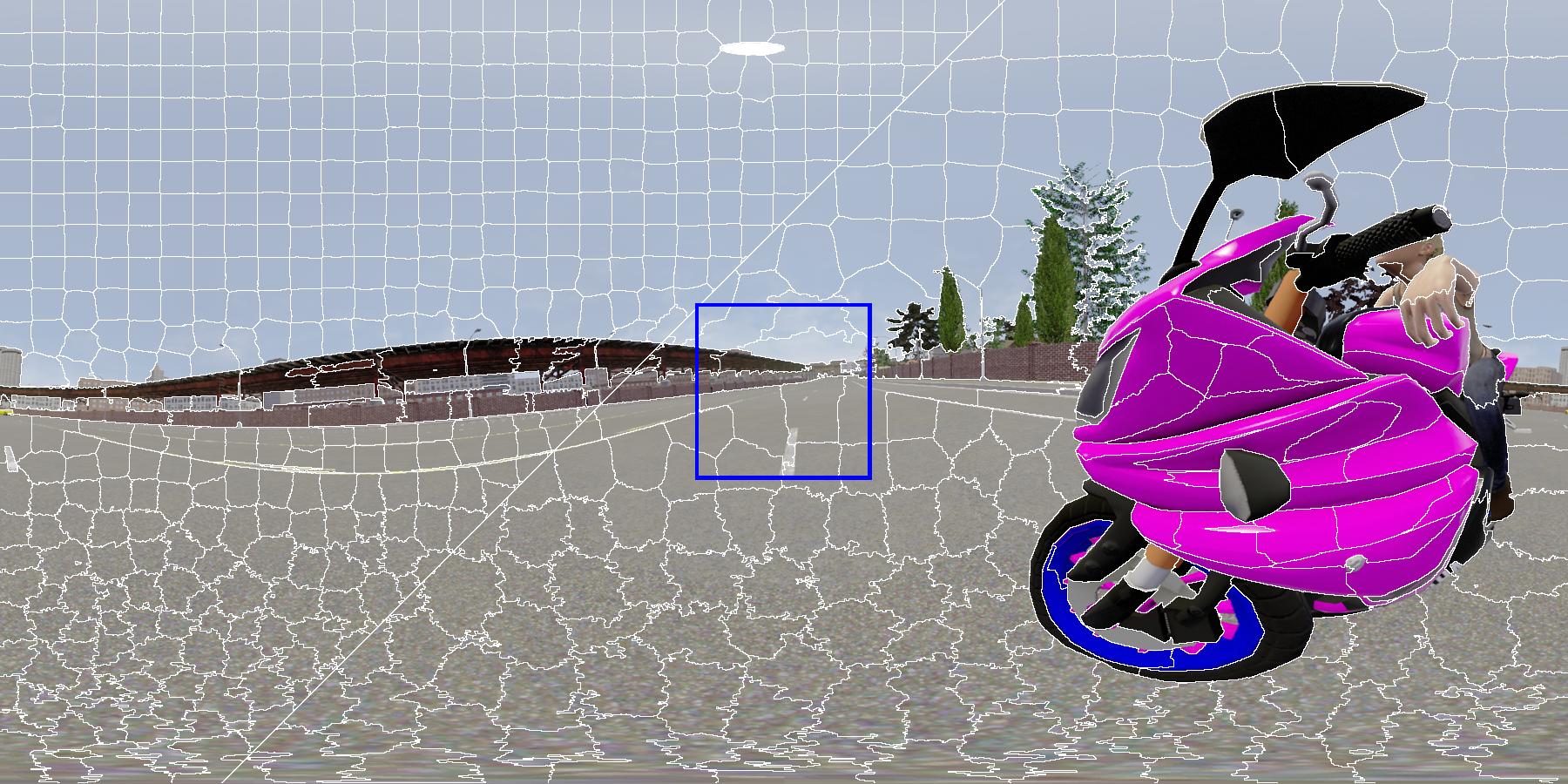}&
\includegraphics[width=\ppp,height=\ppp]{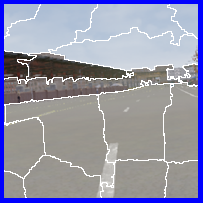}&
\includegraphics[width=\ppp,height=\ppp]{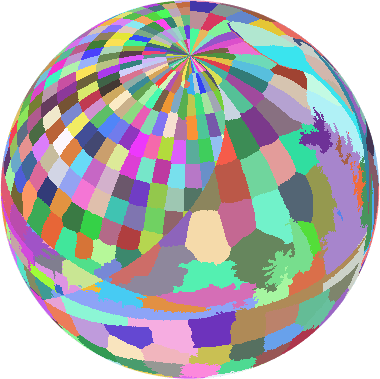}&
\includegraphics[width=\ww,height=\ppp]{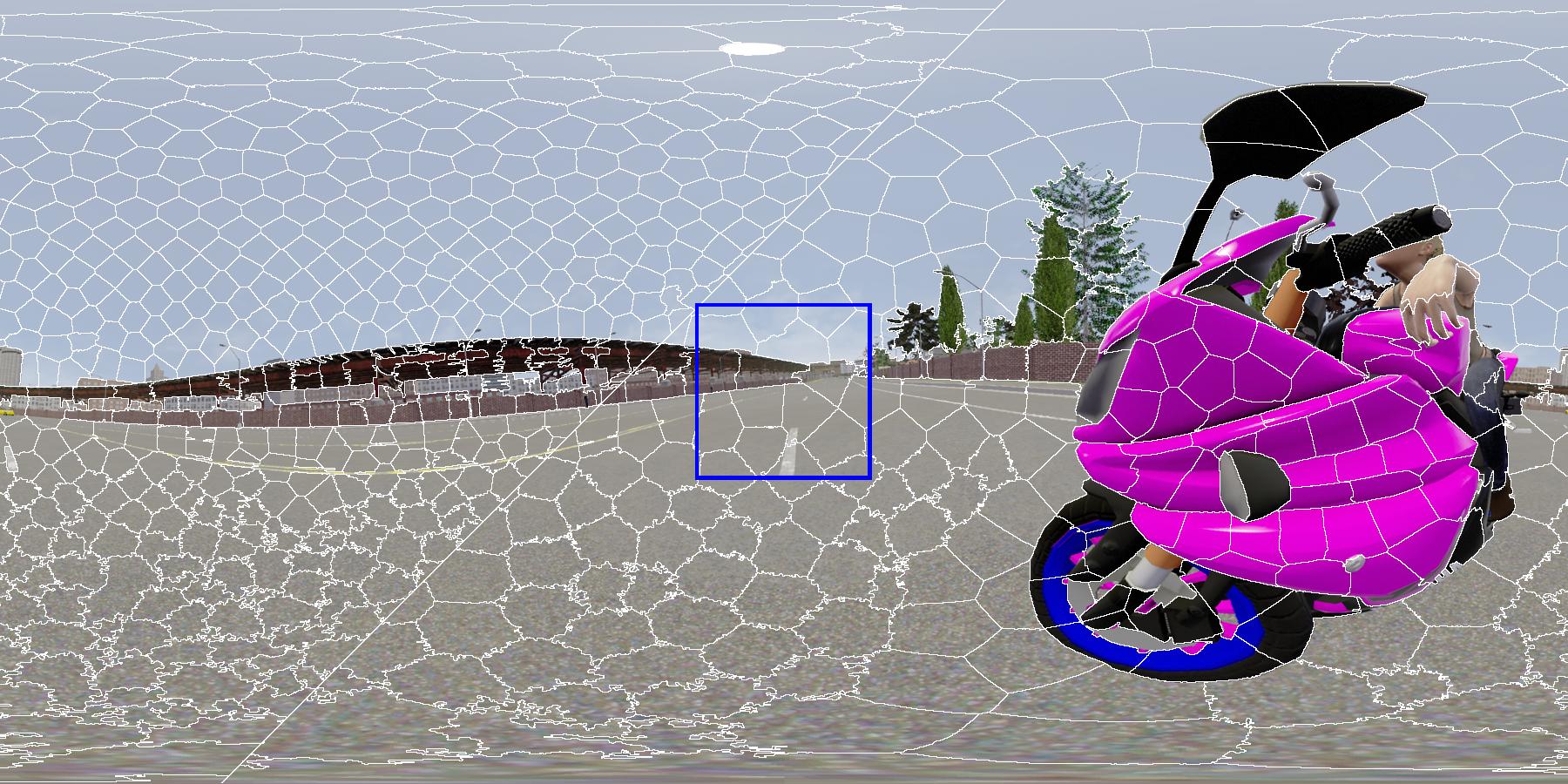}&
\includegraphics[width=\ppp,height=\ppp]{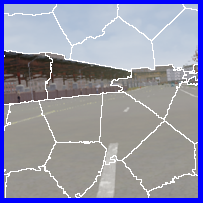}&
\includegraphics[width=\ppp,height=\ppp]{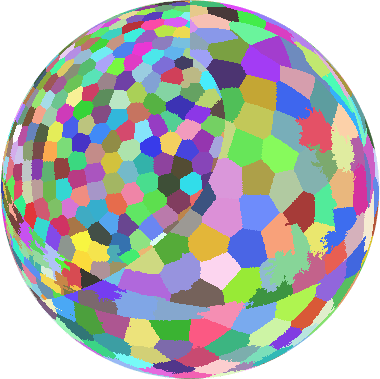}\\[-0.5ex]
\multicolumn{3}{c}{SNIC \cite{achanta2017superpixels}}&
\multicolumn{3}{c}{{SphSLIC-Cos \cite{zhao2018}}} \\[0.75ex]
\includegraphics[width=\ww,height=\ppp]{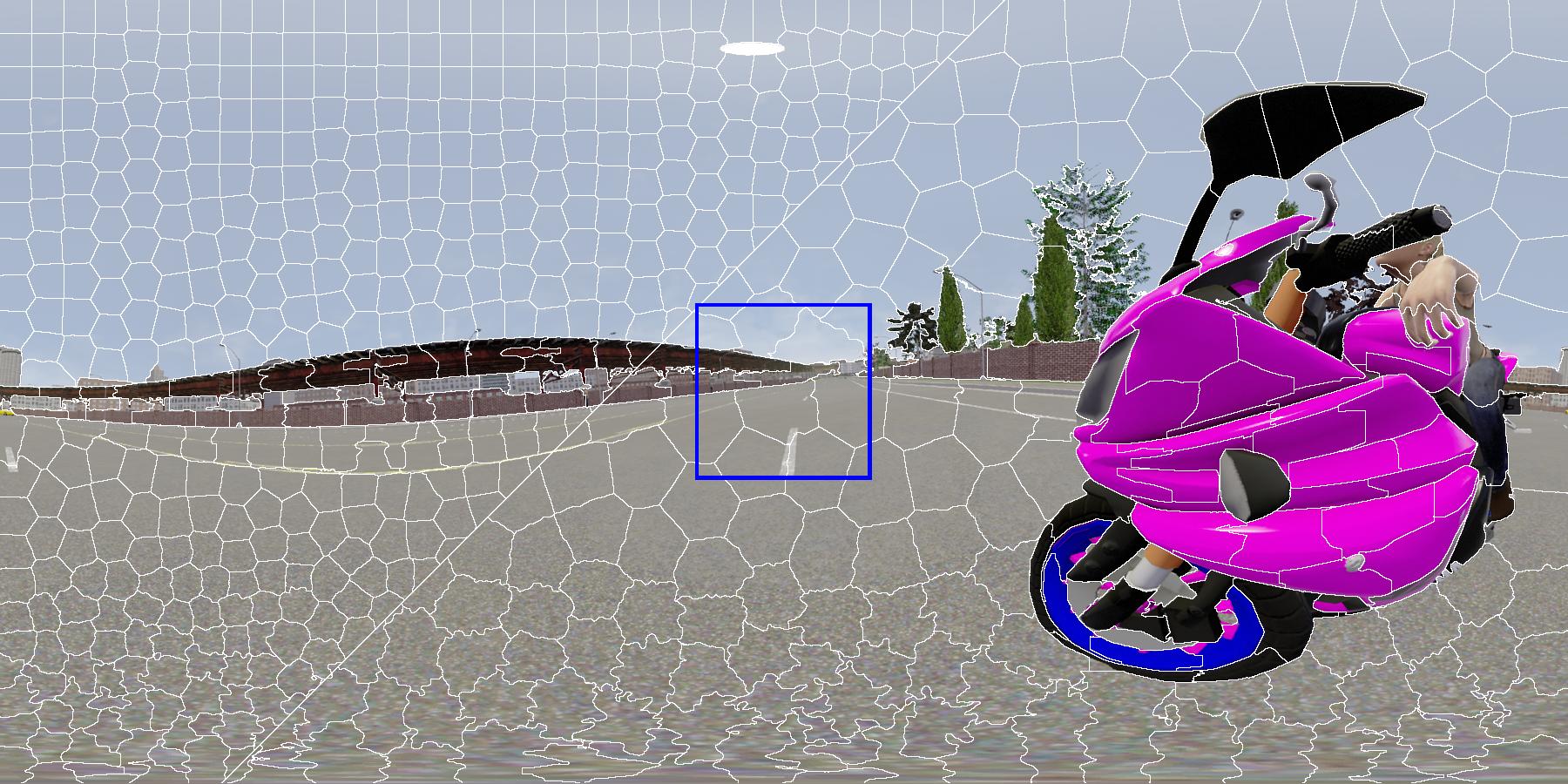}&
\includegraphics[width=\ppp,height=\ppp]{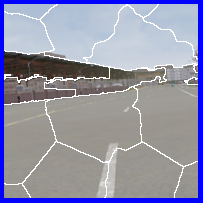}&
\includegraphics[width=\ppp,height=\ppp]{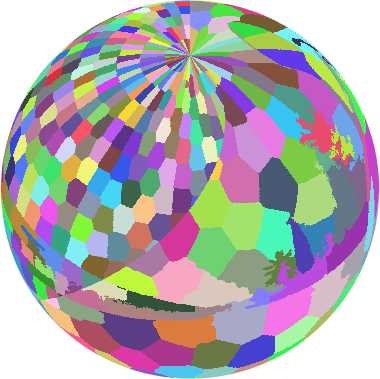}&
\includegraphics[width=\ww,height=\ppp]{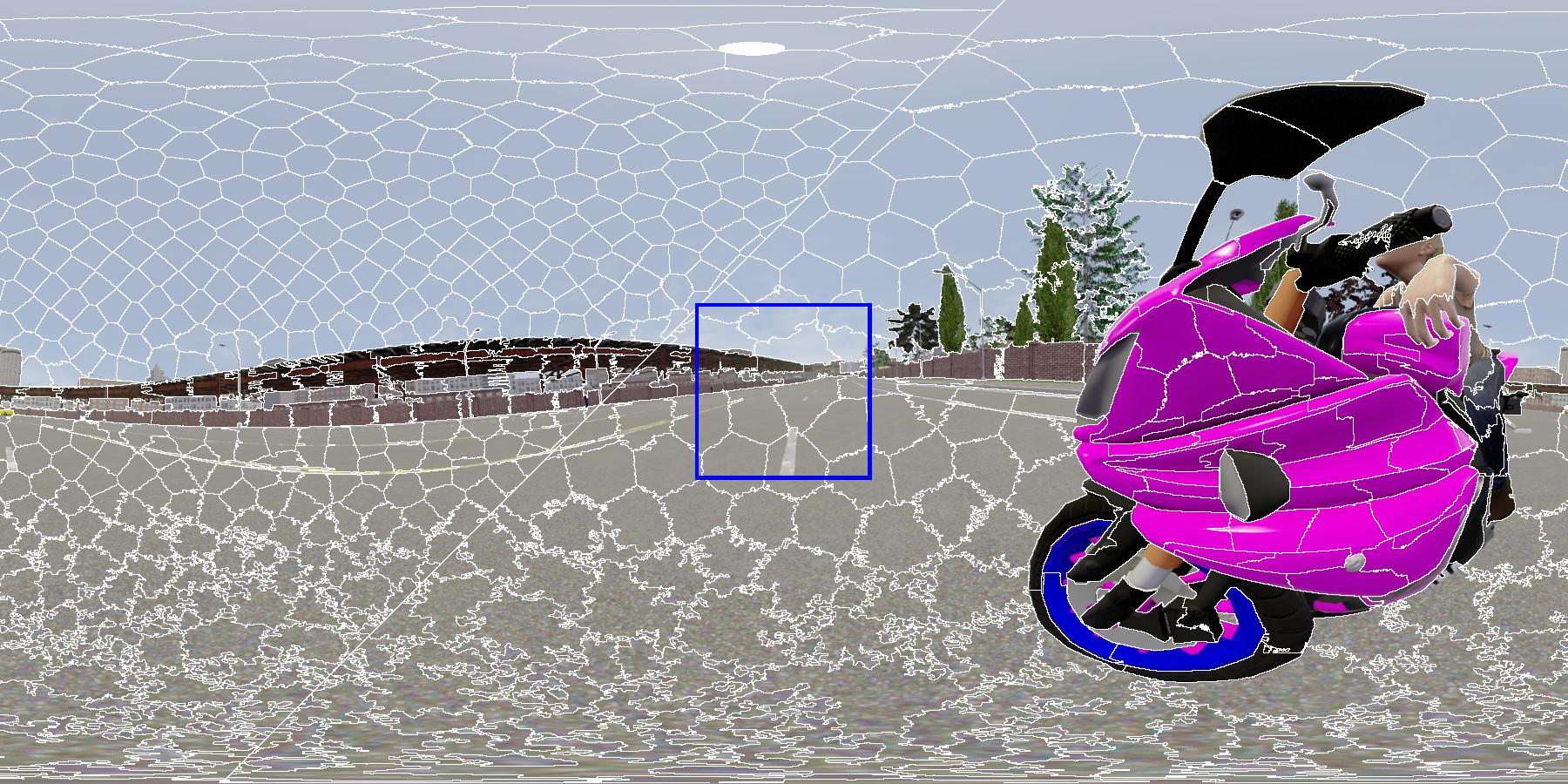}&
\includegraphics[width=\ppp,height=\ppp]{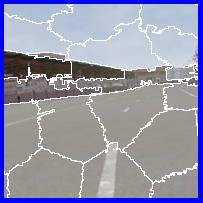}&
\includegraphics[width=\ppp,height=\ppp]{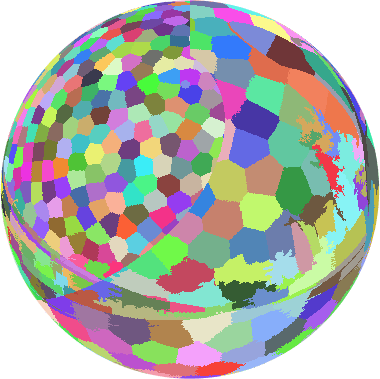}\\[-0.5ex]
\multicolumn{3}{c}{SCALP \cite{giraud2018_scalp}} &
\multicolumn{3}{c}{SphLSC \cite{chen2017,zhao2018}}\\[0.75ex]
\includegraphics[width=\ww,height=\ppp]{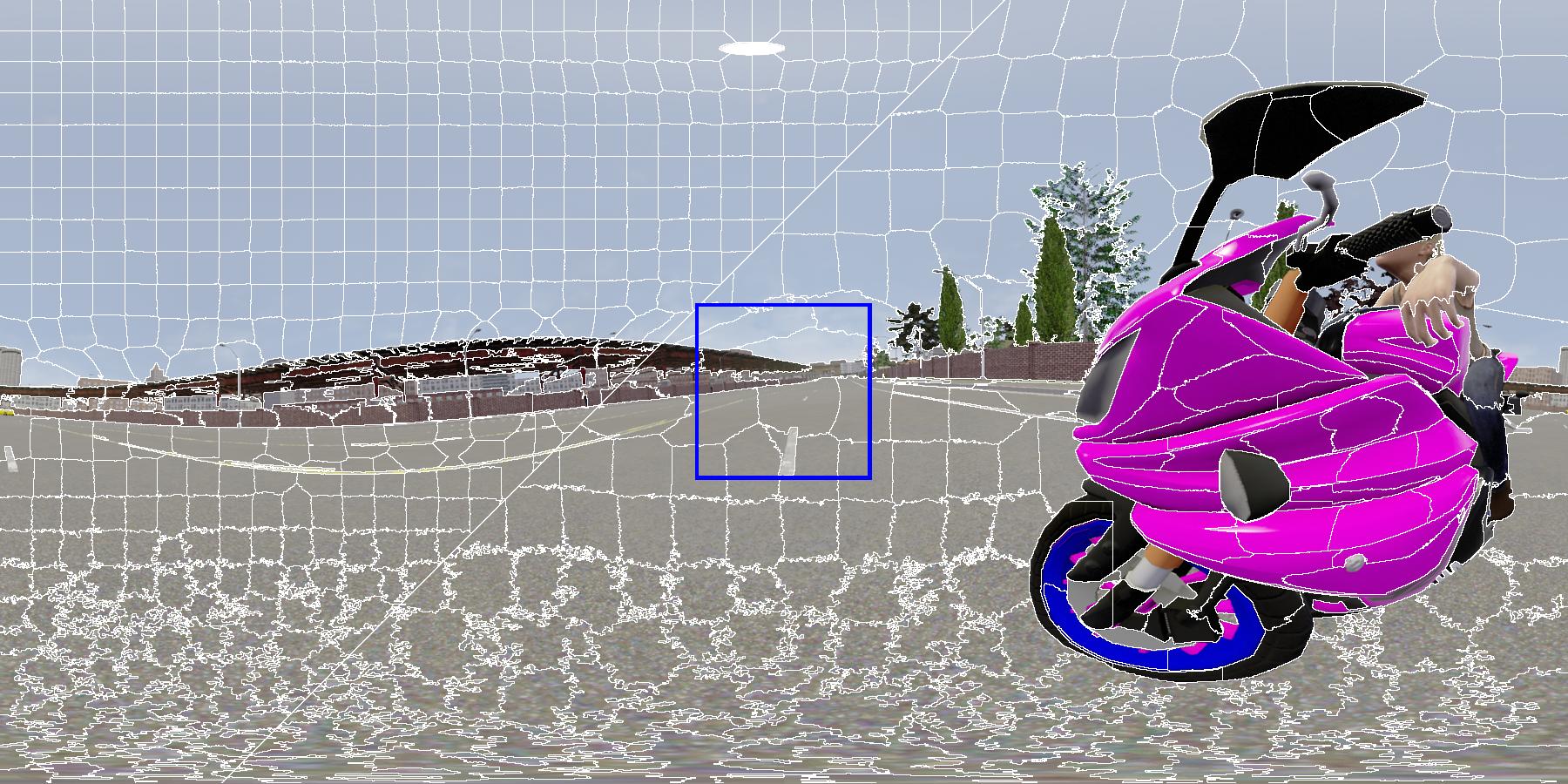}&
\includegraphics[width=\ppp,height=\ppp]{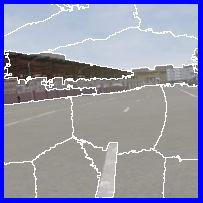}&
\includegraphics[width=\ppp,height=\ppp]{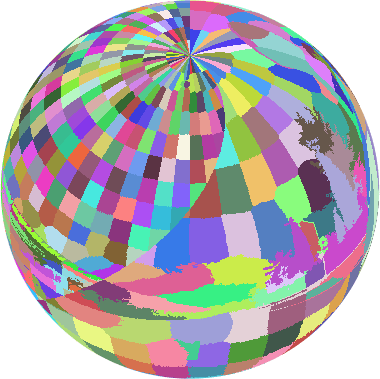}&
\includegraphics[width=\ww,height=\ppp]{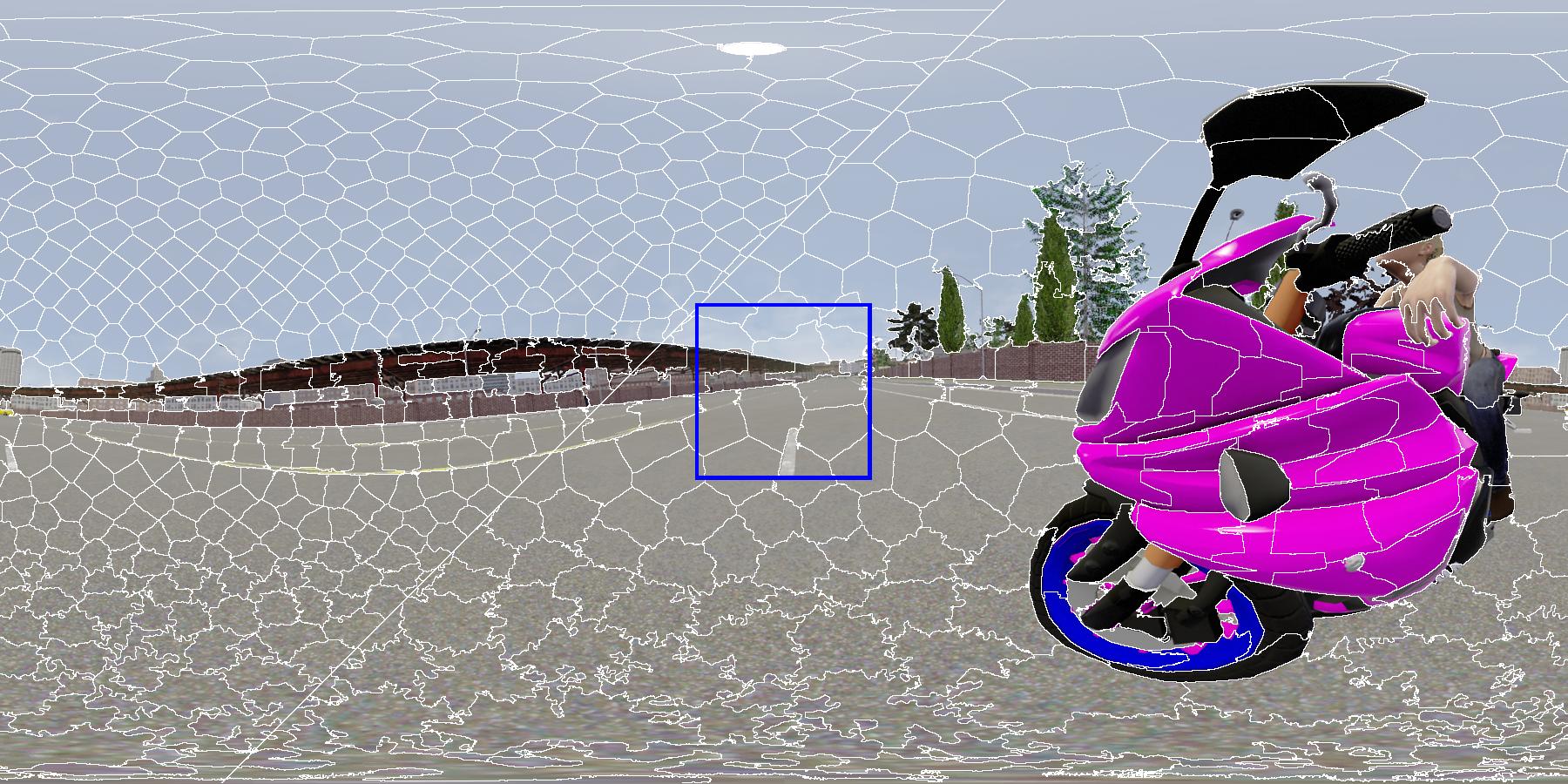}&
\includegraphics[width=\ppp,height=\ppp]{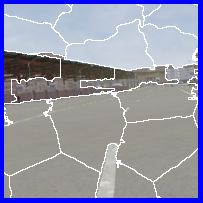}&
\includegraphics[width=\ppp,height=\ppp]{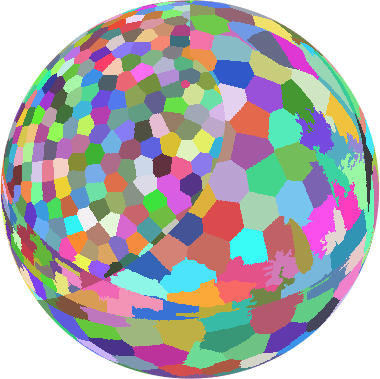}\\[-0.5ex]
\multicolumn{3}{c}{GMMSP \cite{Ban18}} &
\multicolumn{3}{c}{\textbf{SphSPS}}\\[2ex]
\includegraphics[width=\ww,height=\ppp]{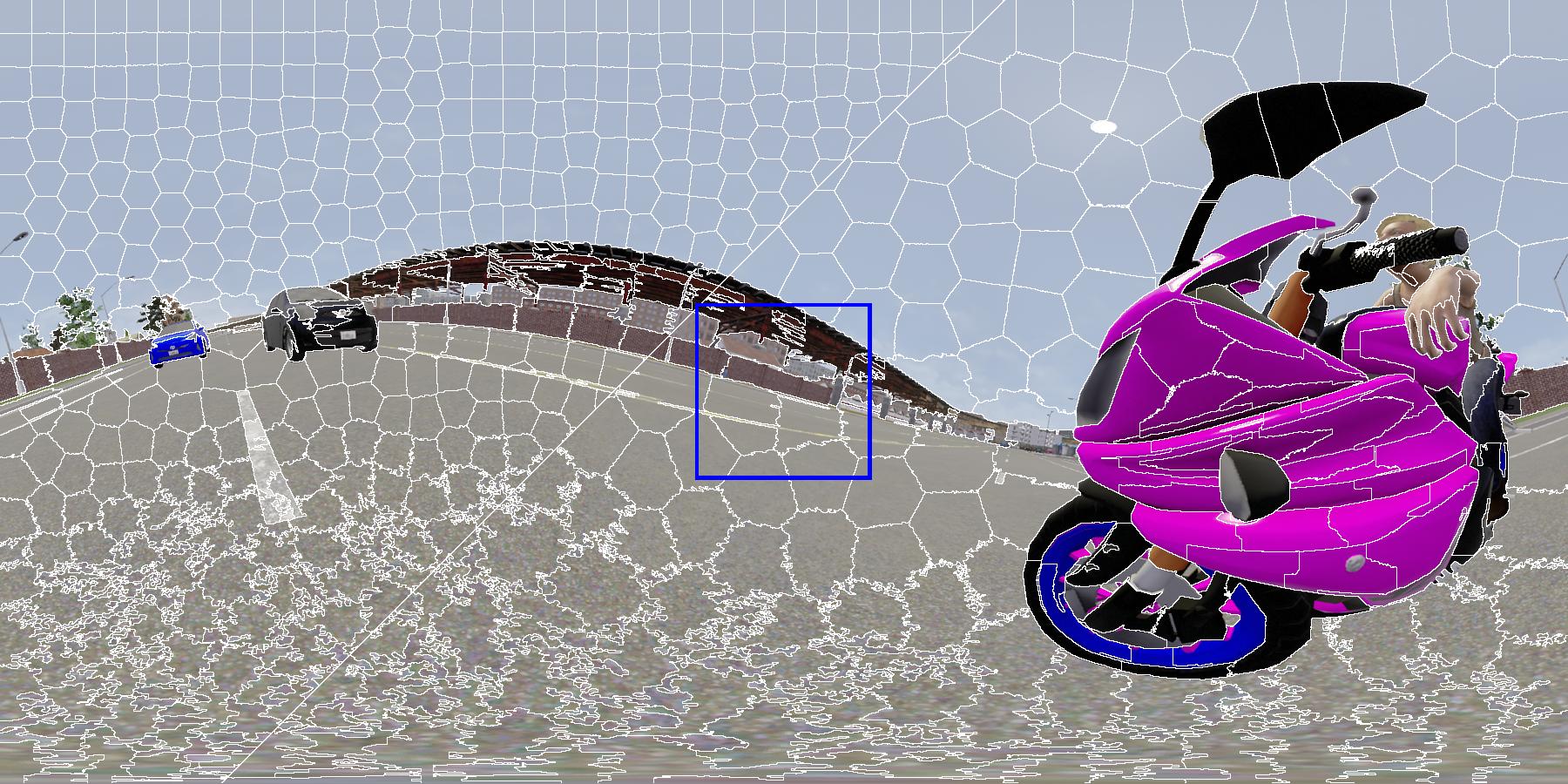}&
\includegraphics[width=\ppp,height=\ppp]{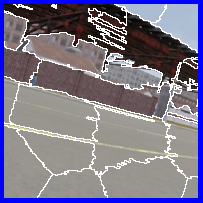}&
\includegraphics[width=\ppp,height=\ppp]{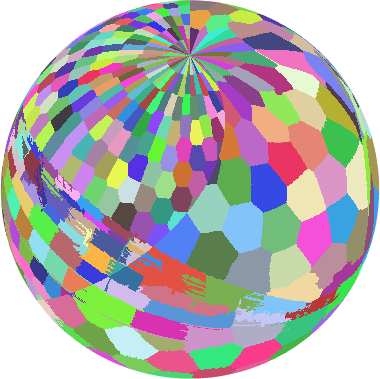}&
% \rotatebox{90}{\hspace{0.35cm} (Euclidean)}&
\includegraphics[width=\ww,height=\ppp]{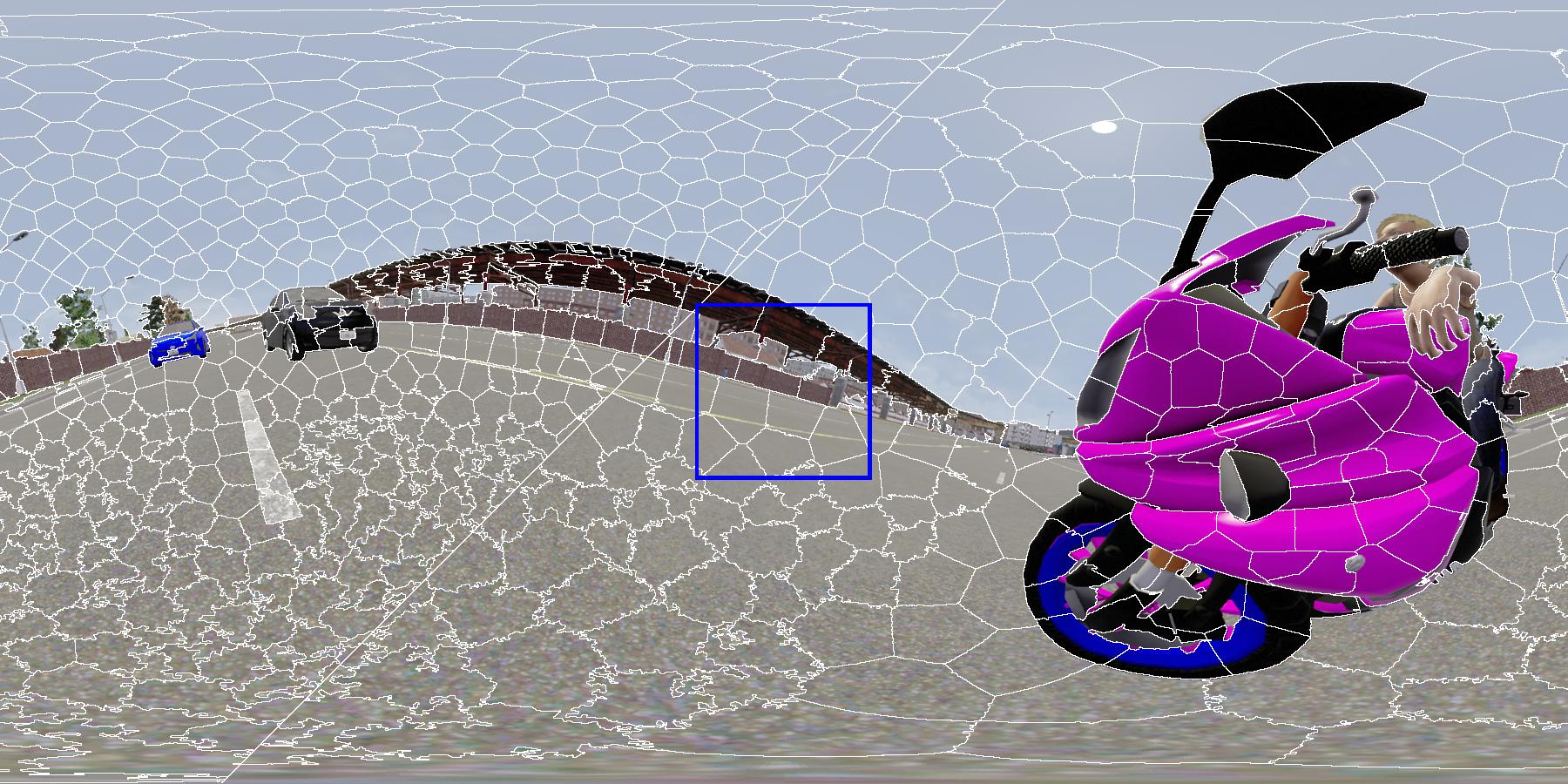}&
\includegraphics[width=\ppp,height=\ppp]{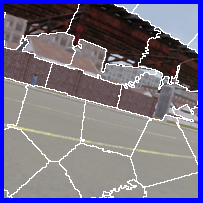}&
\includegraphics[width=\ppp,height=\ppp]{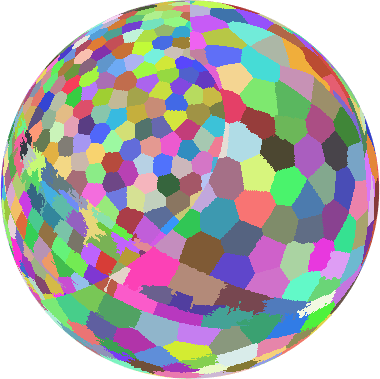}\\[-0.5ex]
\multicolumn{3}{c}{LSC \cite{chen2017}}&
\multicolumn{3}{c}{{SphSLIC-Euc \cite{zhao2018}}} \\[0.75ex]
\includegraphics[width=\ww,height=\ppp]{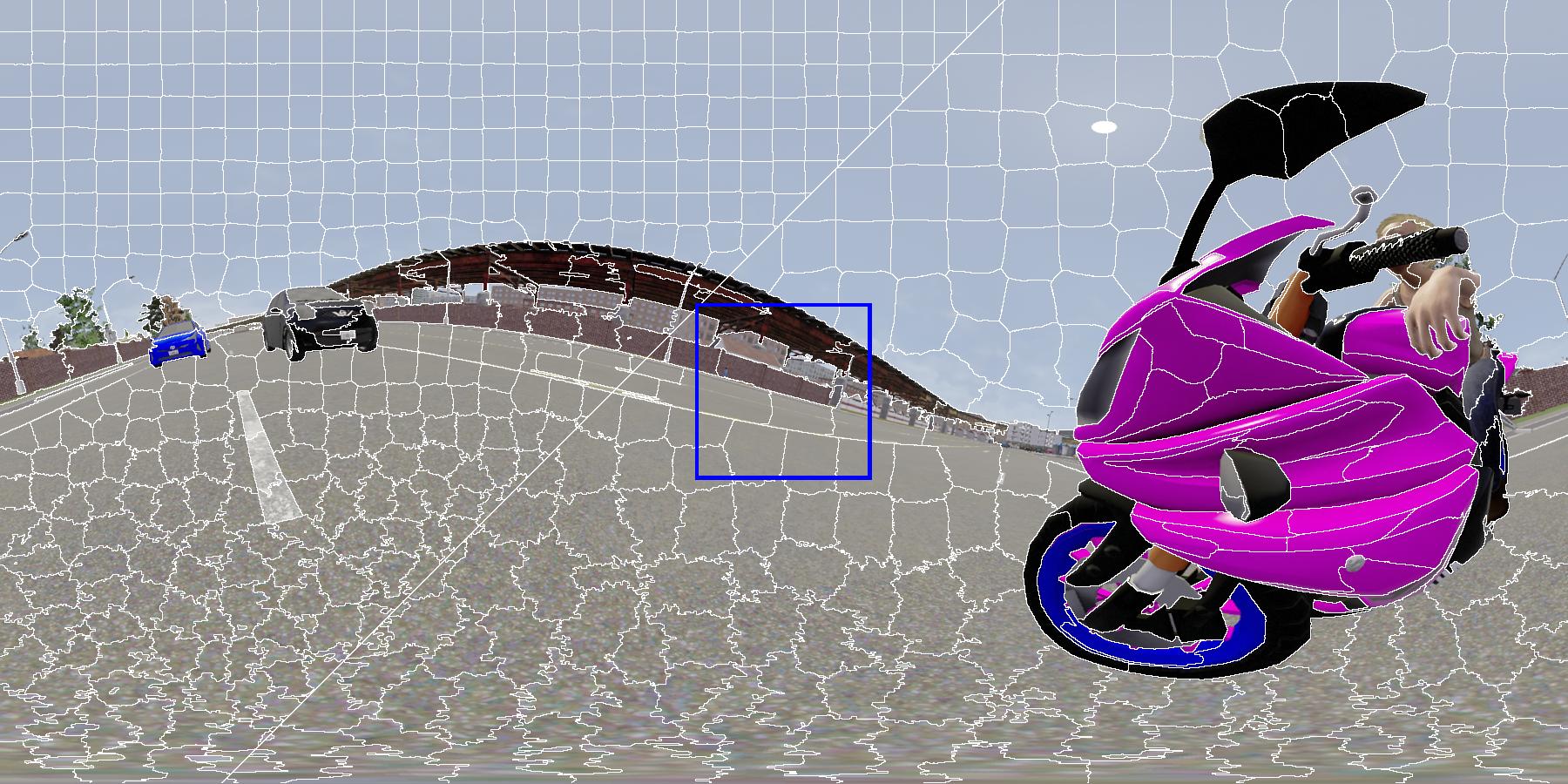}&
\includegraphics[width=\ppp,height=\ppp]{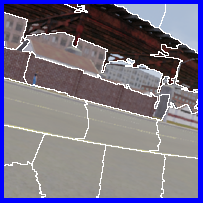}&
\includegraphics[width=\ppp,height=\ppp]{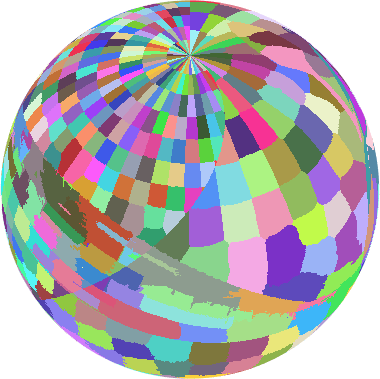}&
\includegraphics[width=\ww,height=\ppp]{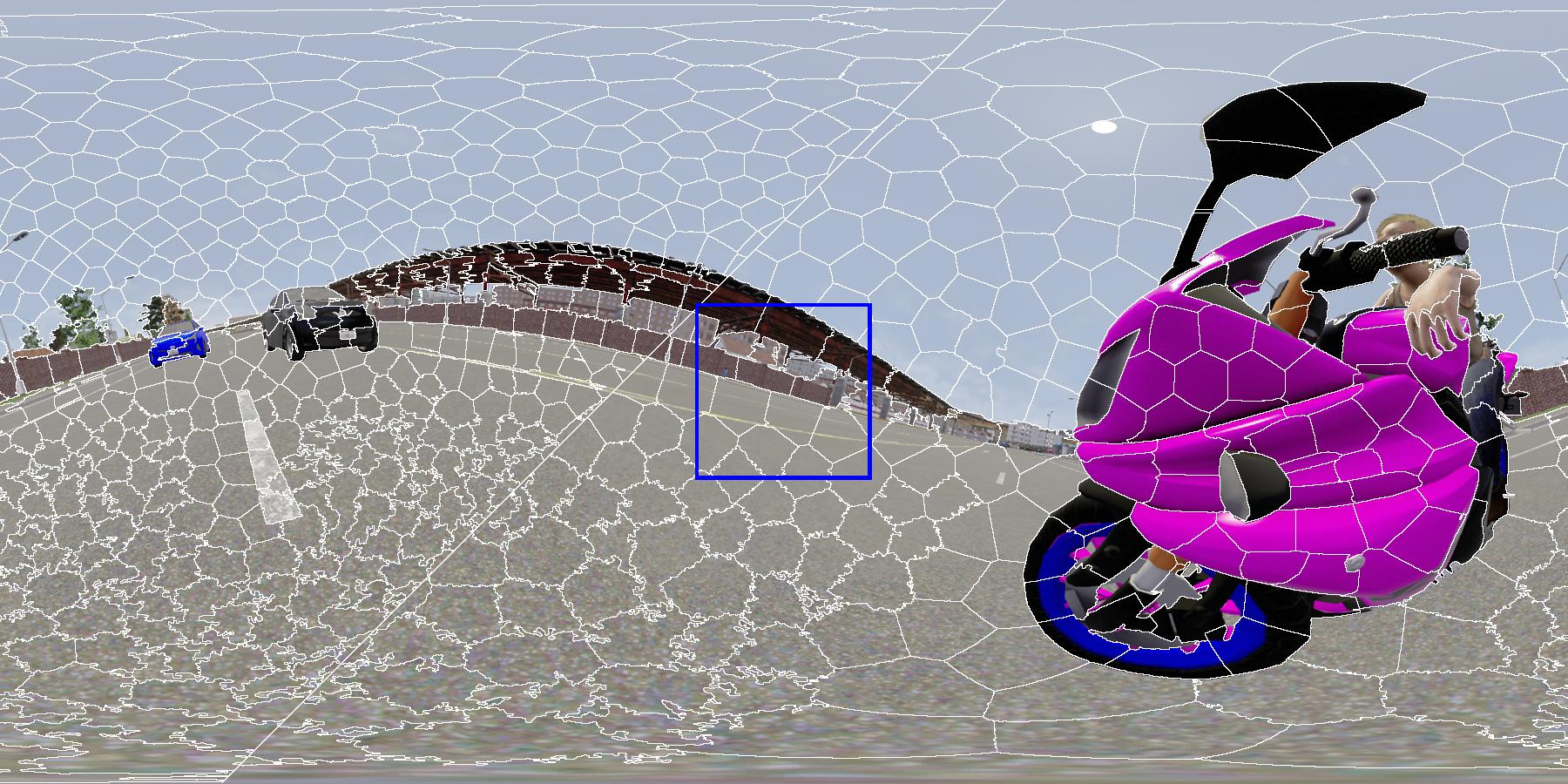}&
\includegraphics[width=\ppp,height=\ppp]{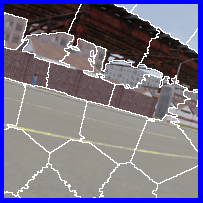}&
\includegraphics[width=\ppp,height=\ppp]{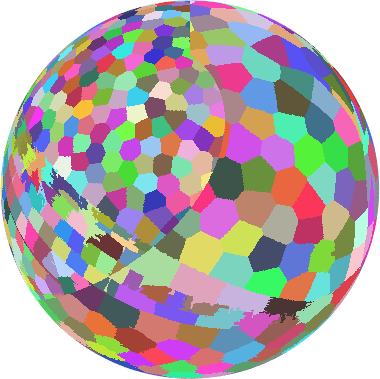}\\[-0.5ex]
\multicolumn{3}{c}{SNIC \cite{achanta2017superpixels}}&
\multicolumn{3}{c}{{SphSLIC-Cos \cite{zhao2018}}} \\[0.75ex]
\includegraphics[width=\ww,height=\ppp]{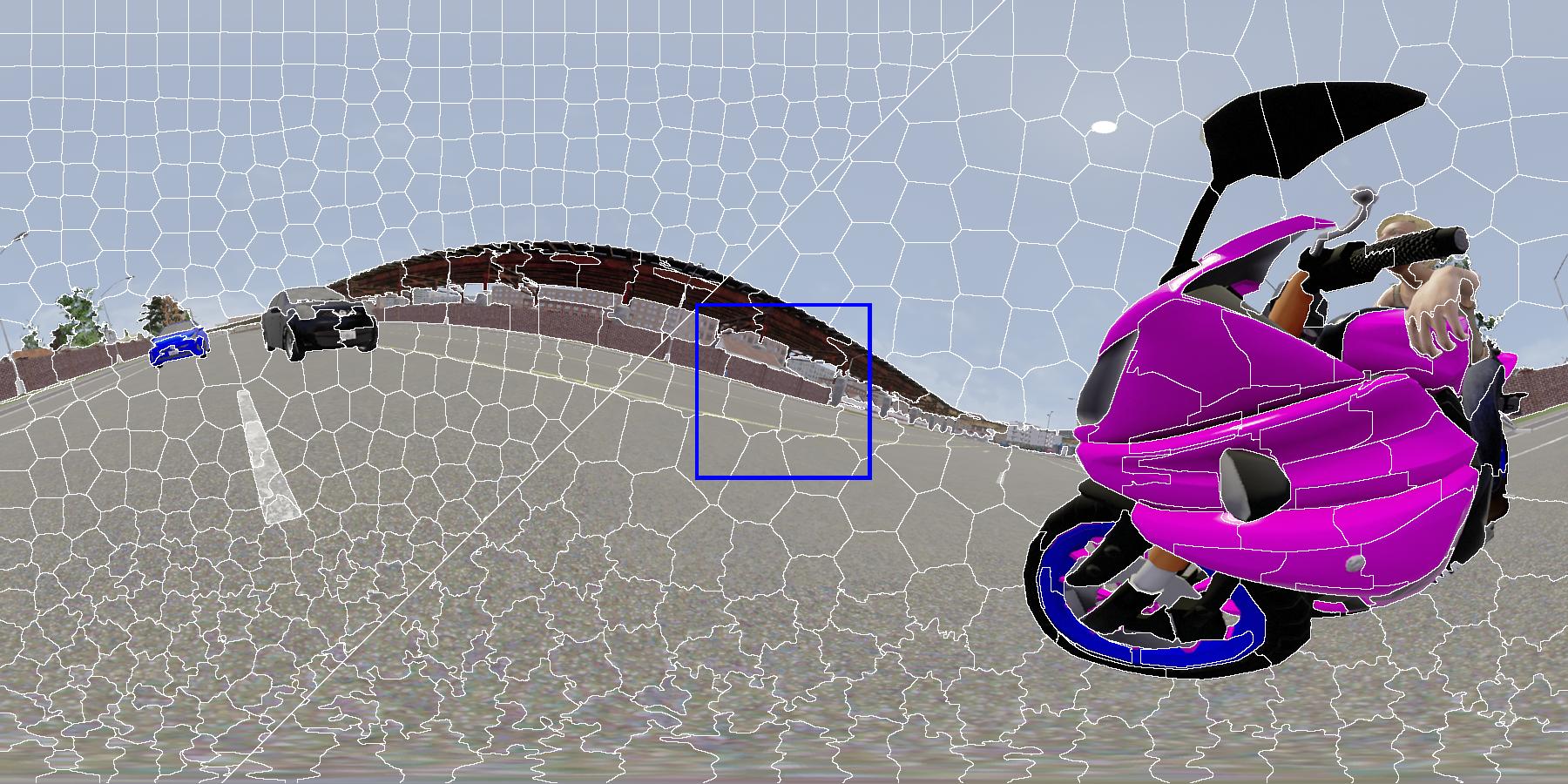}&
\includegraphics[width=\ppp,height=\ppp]{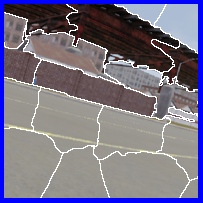}&
\includegraphics[width=\ppp,height=\ppp]{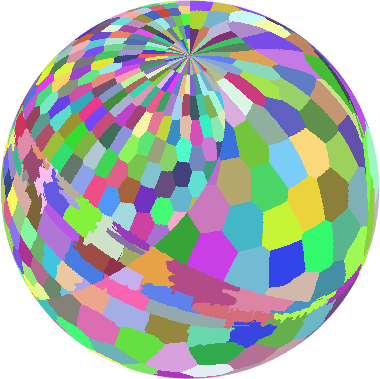}&
\includegraphics[width=\ww,height=\ppp]{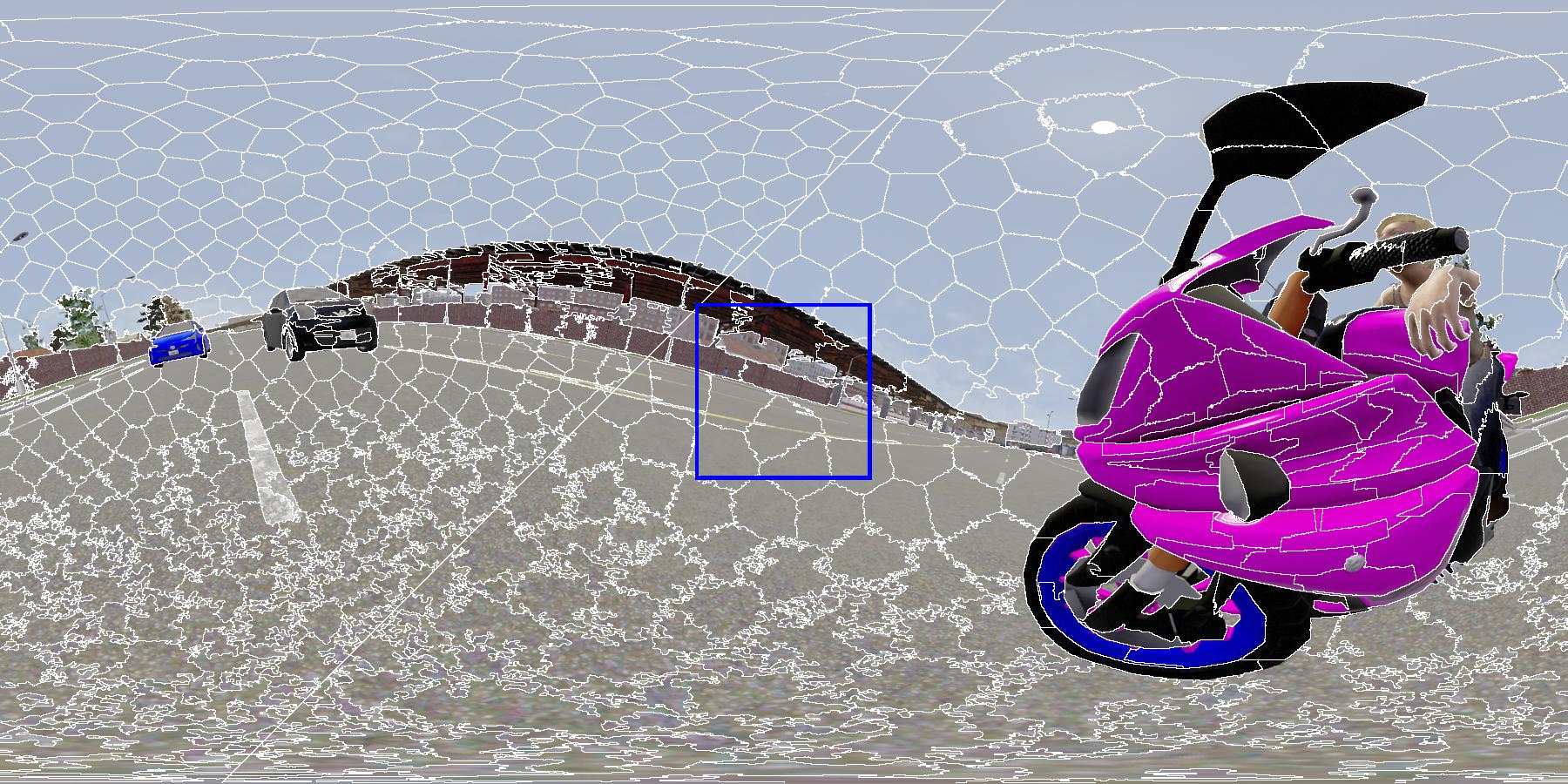}&
\includegraphics[width=\ppp,height=\ppp]{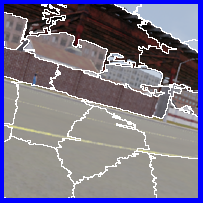}&
\includegraphics[width=\ppp,height=\ppp]{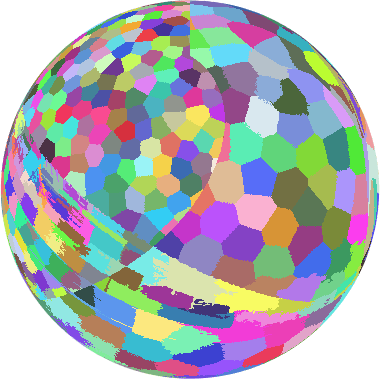}\\[-0.5ex]
\multicolumn{3}{c}{SCALP \cite{giraud2018_scalp}} &
\multicolumn{3}{c}{SphLSC \cite{chen2017,zhao2018}}\\[0.75ex]
\includegraphics[width=\ww,height=\ppp]{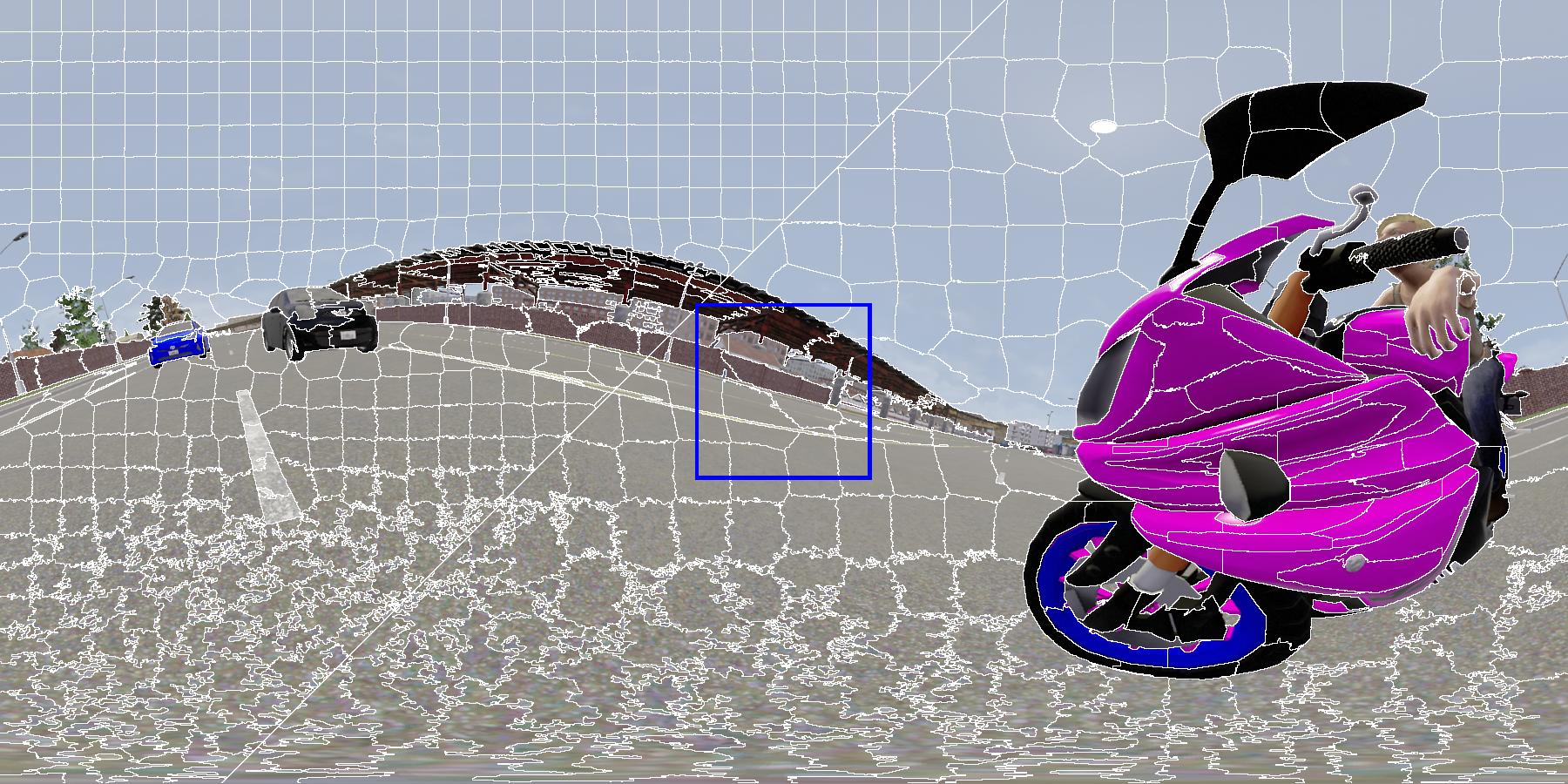}&
\includegraphics[width=\ppp,height=\ppp]{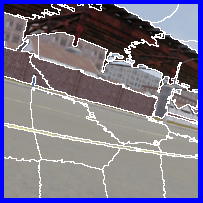}&
\includegraphics[width=\ppp,height=\ppp]{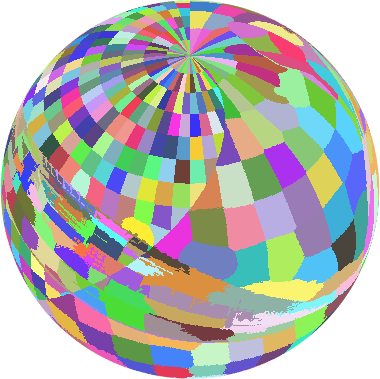}&
\includegraphics[width=\ww,height=\ppp]{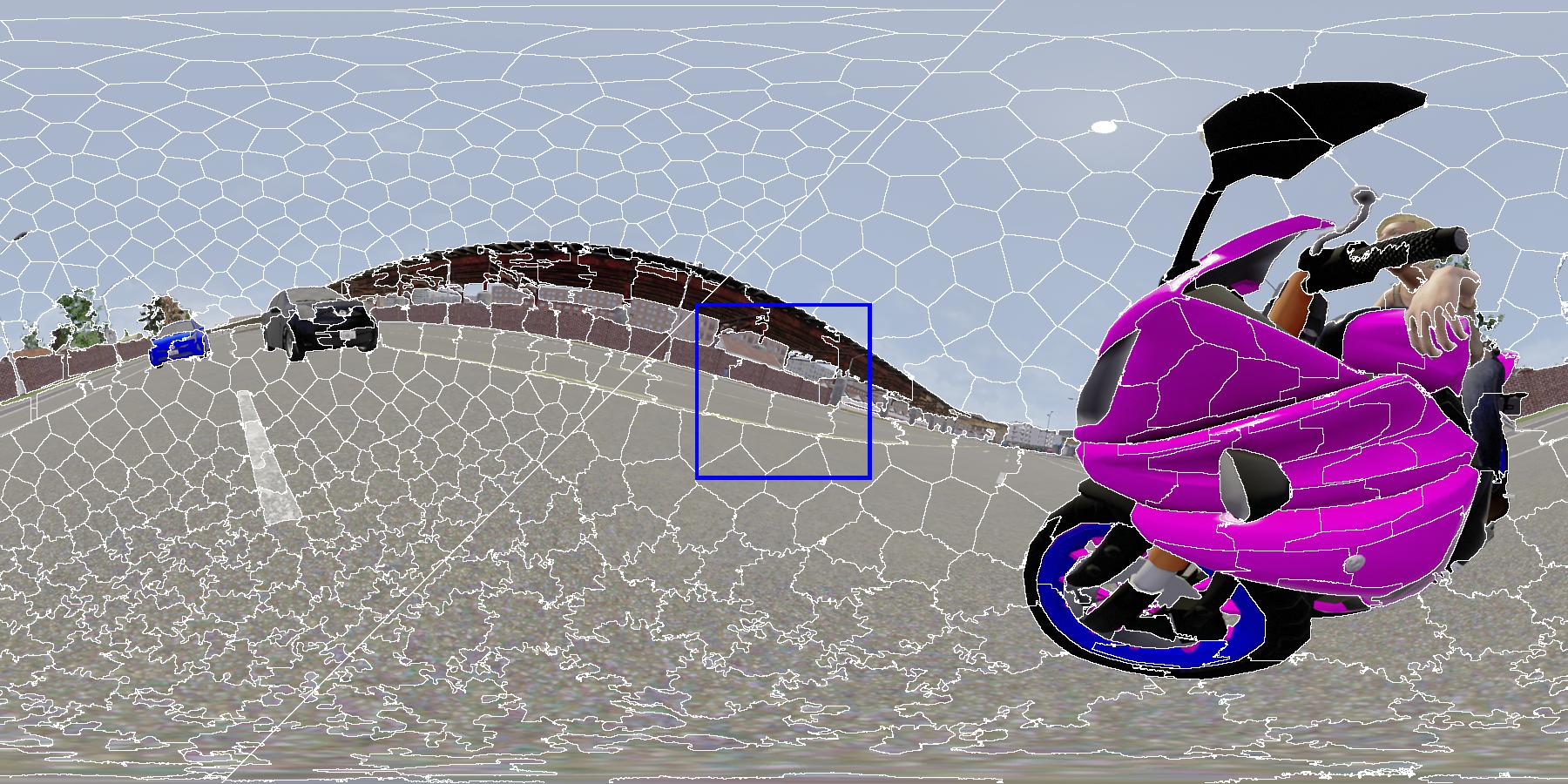}&
\includegraphics[width=\ppp,height=\ppp]{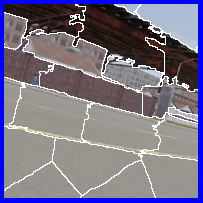}&
\includegraphics[width=\ppp,height=\ppp]{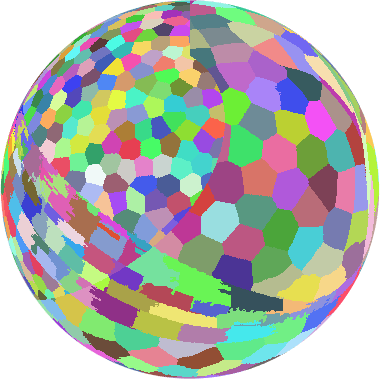}\\[-0.5ex]
\multicolumn{3}{c}{GMMSP \cite{Ban18}} &
\multicolumn{3}{c}{\textbf{SphSPS}}\\
\end{tabular}
}
\caption{
Visual comparison between SphSPS and the
best planar and spherical (underlined) state-of-the-art
methods on Omniscape images, for two superpixel numbers $K=1200$ (top-left) and $K=400$ (bottom right).
SphSPS produces regular spherical superpixels with smooth boundaries
  that adhere well to the image contours}%
\label{fig:sps_soa_img_omni}
\end{figure*}

 \subsubsection{Robustness to noise}

\begin{figure*}[t!]
\centering
{\scriptsize
\begin{tabular}{@{\hspace{0mm}}c@{\hspace{1mm}}c@{\hspace{1mm}}c@{\hspace{0mm}}}
\includegraphics[width=\wwh,height=\hhh]{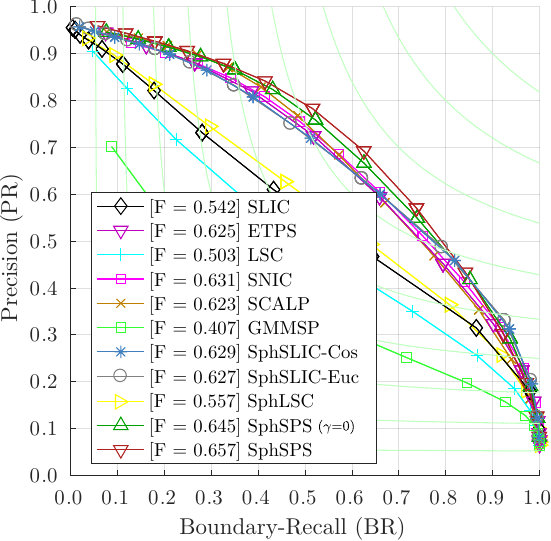}&
\includegraphics[width=\wwh,height=\hhh]{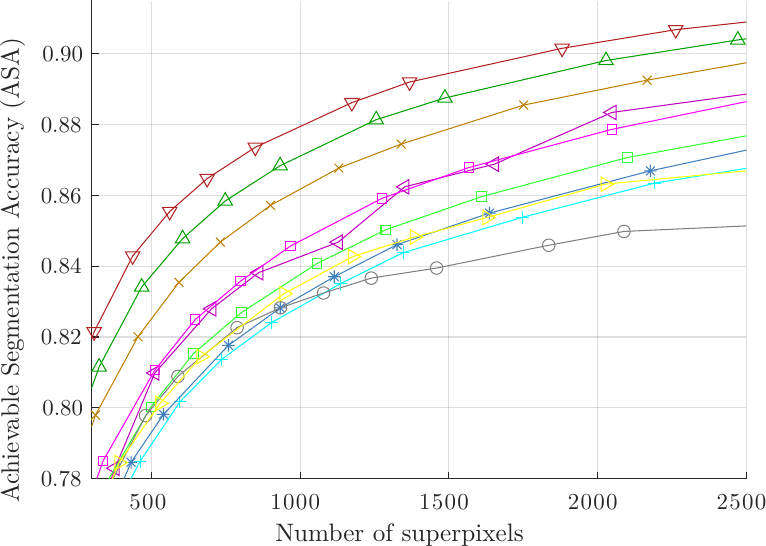}&
\includegraphics[width=\wwh,height=\hhh]{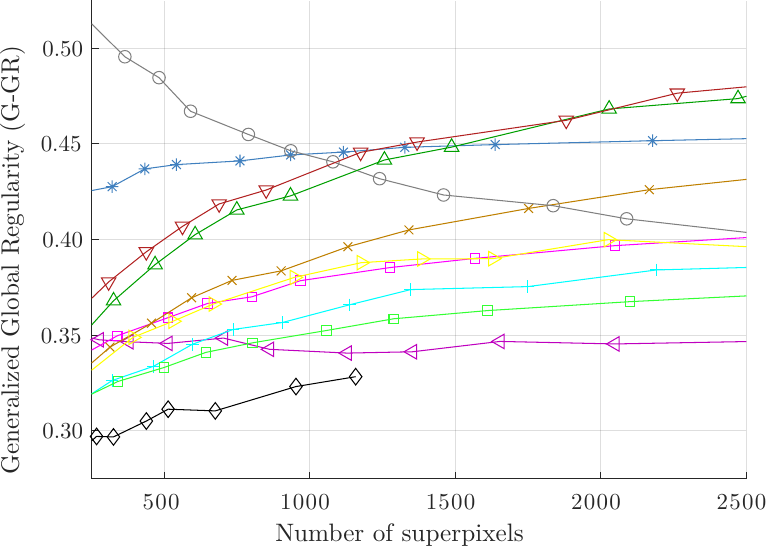}\\
% (a) P-R (F-measure) & (b) BR vs CD & (c) ASA & (d) GR \\
\end{tabular}
}
\caption{
Quantitative comparison on the noisy PSD images \cite{wan2018} (addition of a white Gaussian noise of variance 20), on PR/BR, %BR/CD,
ASA and G-GR metrics
of the proposed SphSPS method to the state-of-the-art ones}%
\label{fig:sps_soa_noise}
\end{figure*}

\begin{figure*}[ht!]

\centering
{\scriptsize
\begin{tabular}{@{\hspace{1mm}}c@{\hspace{1mm}}c@{\hspace{1mm}}c@{\hspace{3mm}}c@{\hspace{1mm}}c@{\hspace{1mm}}c@{\hspace{0mm}}}
\multicolumn{3}{c}{\small \textbf{Planar methods}} &
\multicolumn{3}{c}{\small \textbf{Spherical methods}} \\[1ex]
\includegraphics[width=\ww]{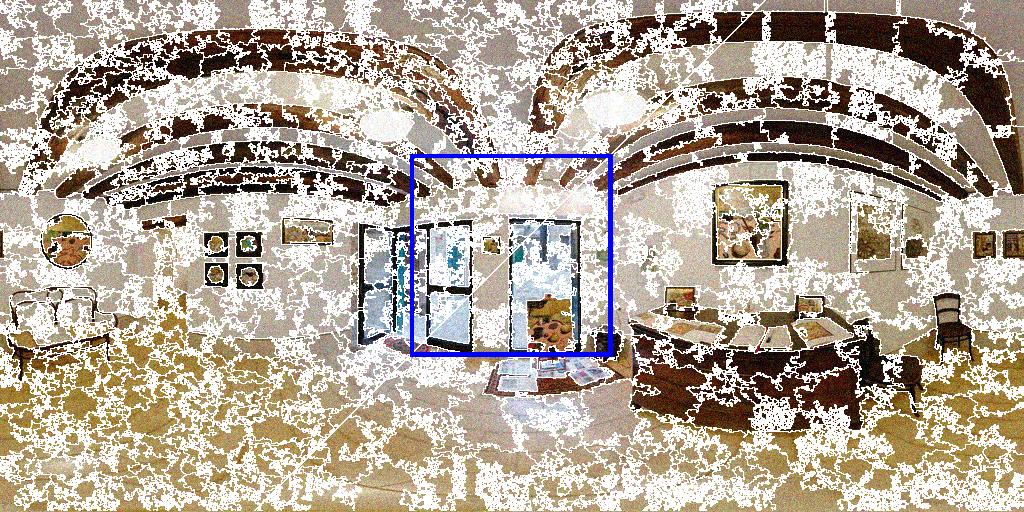}&
\includegraphics[width=\ppp,height=\ppp]{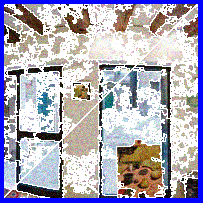}&
\includegraphics[width=\ppp,height=\ppp]{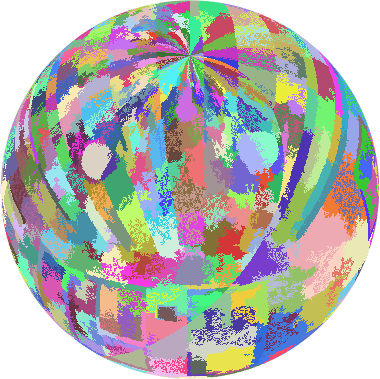}&
% \rotatebox{90}{\hspace{0.35cm} (Euclidean)}&
\includegraphics[width=\ww]{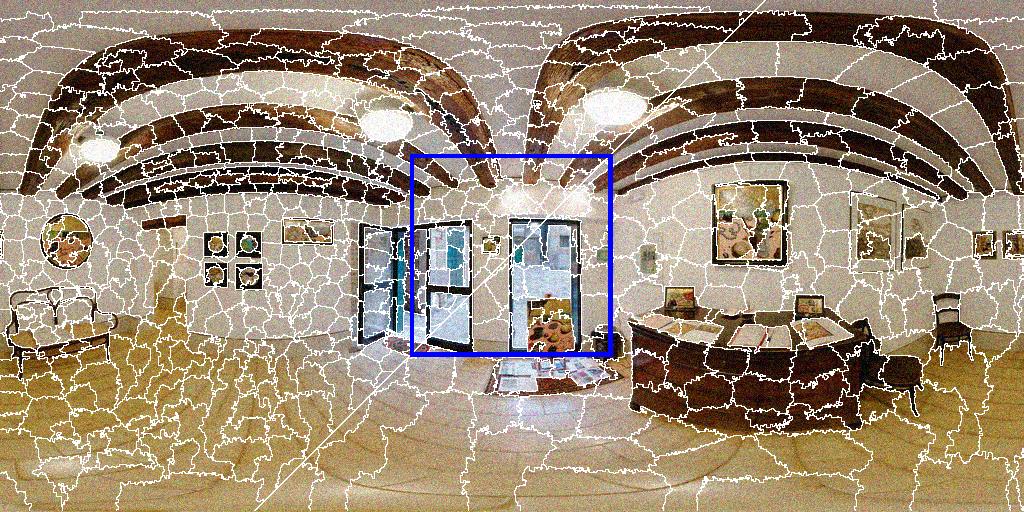}&
\includegraphics[width=\ppp,height=\ppp]{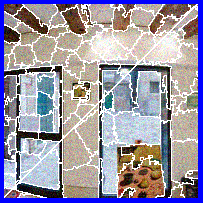}&
\includegraphics[width=\ppp,height=\ppp]{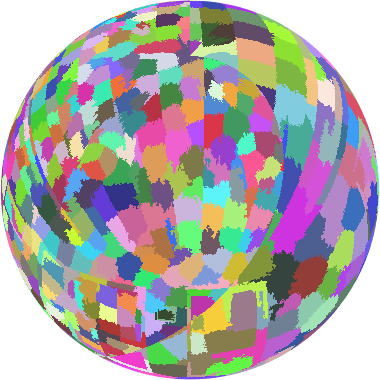}\\[-0.5ex]
\multicolumn{3}{c}{LSC \cite{chen2017}}&
\multicolumn{3}{c}{{SphSLIC-Euc \cite{zhao2018}}} \\[0.5ex]
\includegraphics[width=\ww]{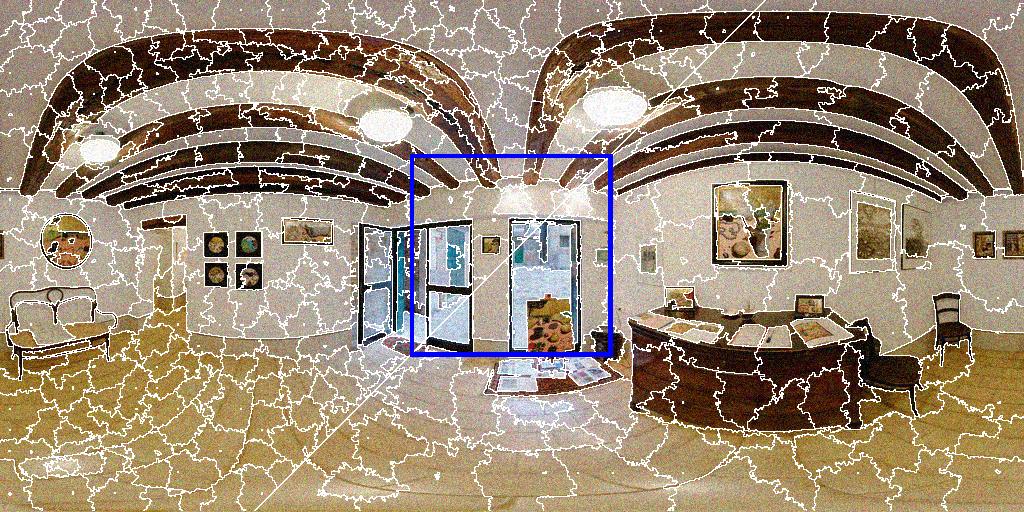}&
\includegraphics[width=\ppp,height=\ppp]{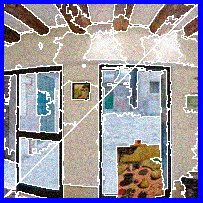}&
\includegraphics[width=\ppp,height=\ppp]{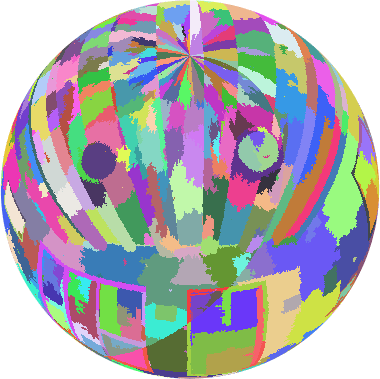}&
% \rotatebox{90}{\hspace{0.35cm} (Cosine-opt)}&
\includegraphics[width=\ww]{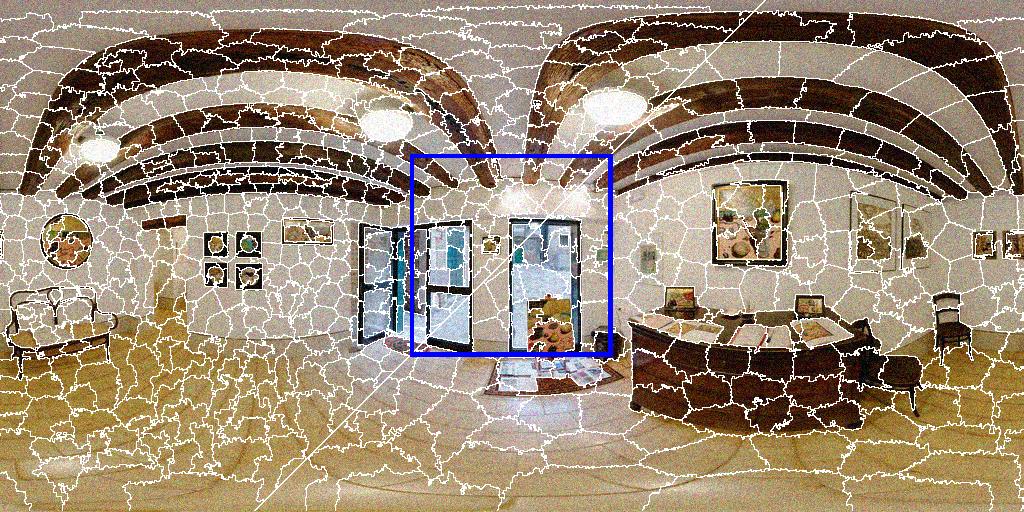}&
\includegraphics[width=\ppp,height=\ppp]{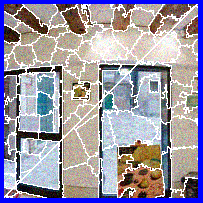}&
\includegraphics[width=\ppp,height=\ppp]{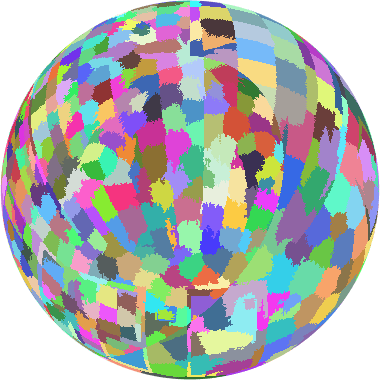}\\[-0.5ex]
\multicolumn{3}{c}{SNIC \cite{achanta2017superpixels}}&
\multicolumn{3}{c}{{SphSLIC-Cos \cite{zhao2018}}} \\[0.5ex]
\includegraphics[width=\ww]{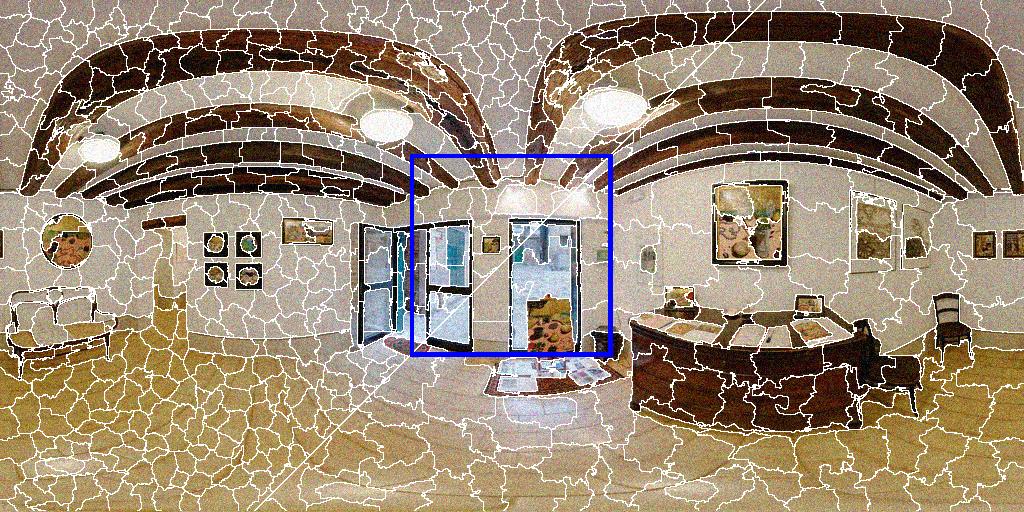}&
\includegraphics[width=\ppp,height=\ppp]{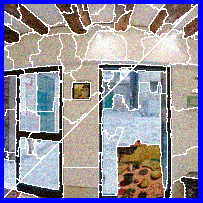}&
\includegraphics[width=\ppp,height=\ppp]{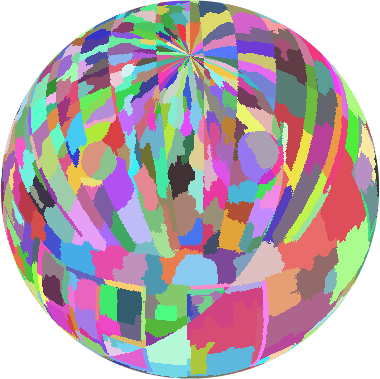}&
\includegraphics[width=\ww]{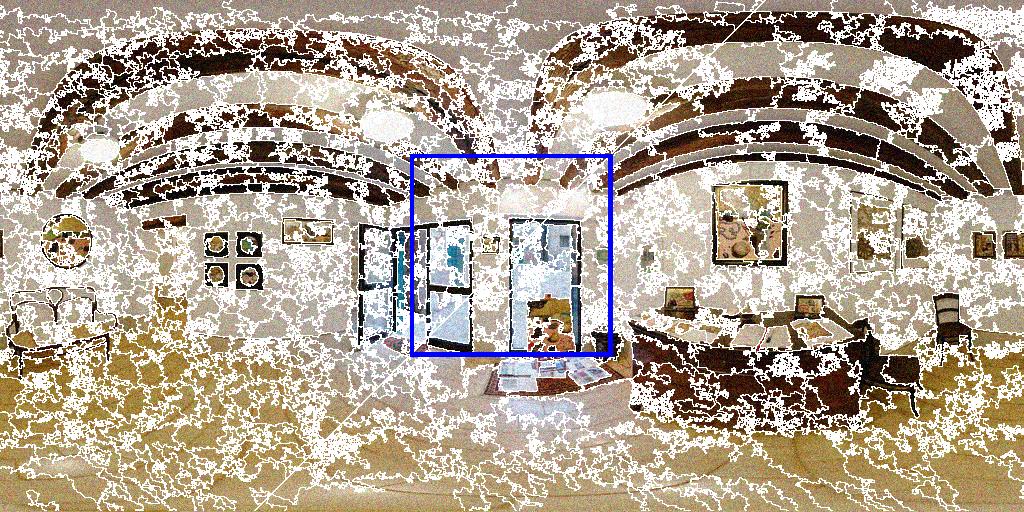}&
\includegraphics[width=\ppp,height=\ppp]{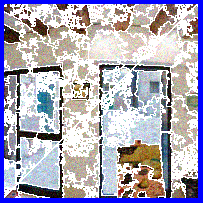}&
\includegraphics[width=\ppp,height=\ppp]{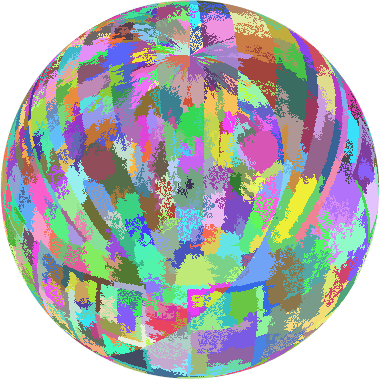}\\[-0.5ex]
\multicolumn{3}{c}{SCALP \cite{giraud2018_scalp}}&
\multicolumn{3}{c}{SphLSC \cite{chen2017,zhao2018}} \\[0.75ex]
\includegraphics[width=\ww]{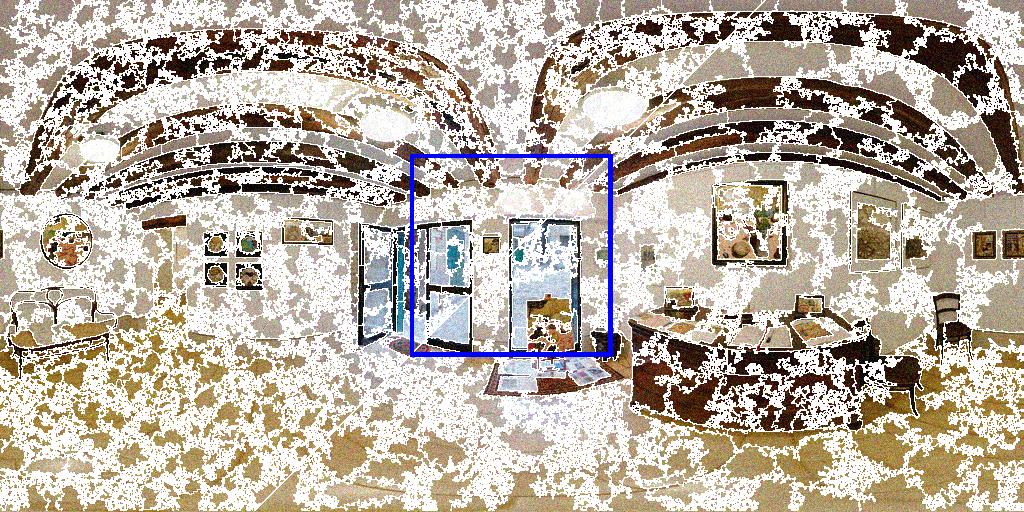}&
\includegraphics[width=\ppp,height=\ppp]{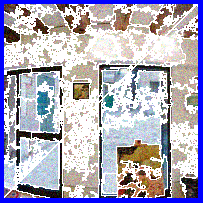}&
\includegraphics[width=\ppp,height=\ppp]{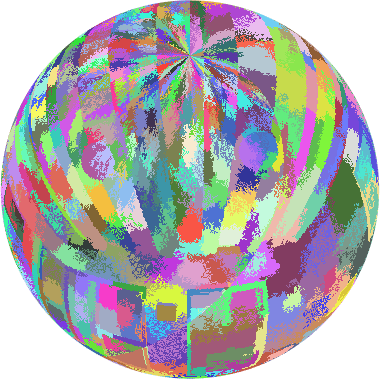}&
\includegraphics[width=\ww]{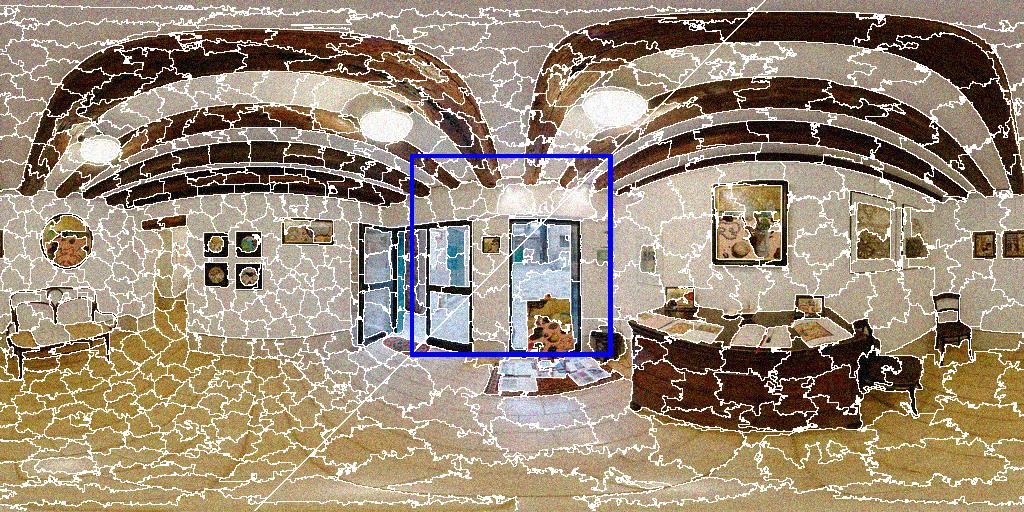}&
\includegraphics[width=\ppp,height=\ppp]{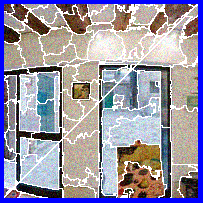}&
\includegraphics[width=\ppp,height=\ppp]{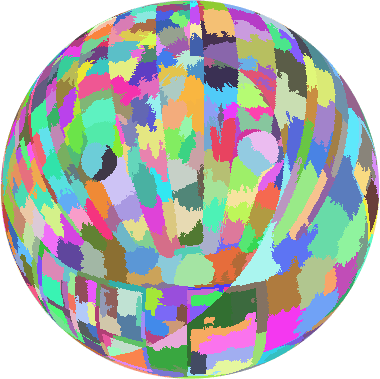}\\[-0.5ex]
\multicolumn{3}{c}{GMMSP \cite{Ban18}} &
\multicolumn{3}{c}{{\textbf{SphSPS}}} \\[2ex]
\includegraphics[width=\ww]{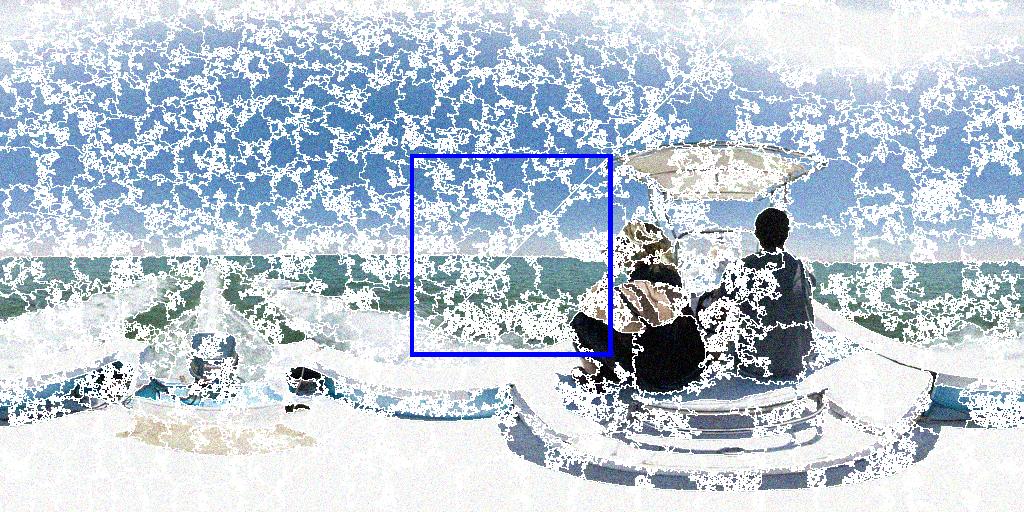}&
\includegraphics[width=\ppp,height=\ppp]{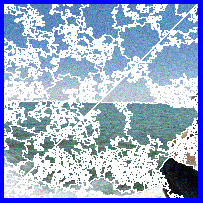}&
\includegraphics[width=\ppp,height=\ppp]{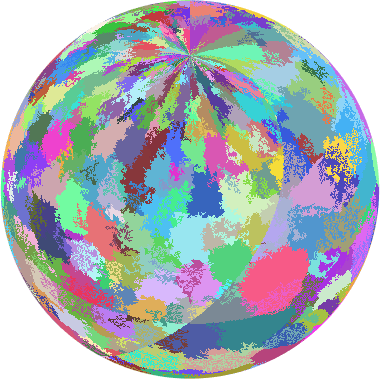}&
% \rotatebox{90}{\hspace{0.35cm} (Euclidean)}&
\includegraphics[width=\ww]{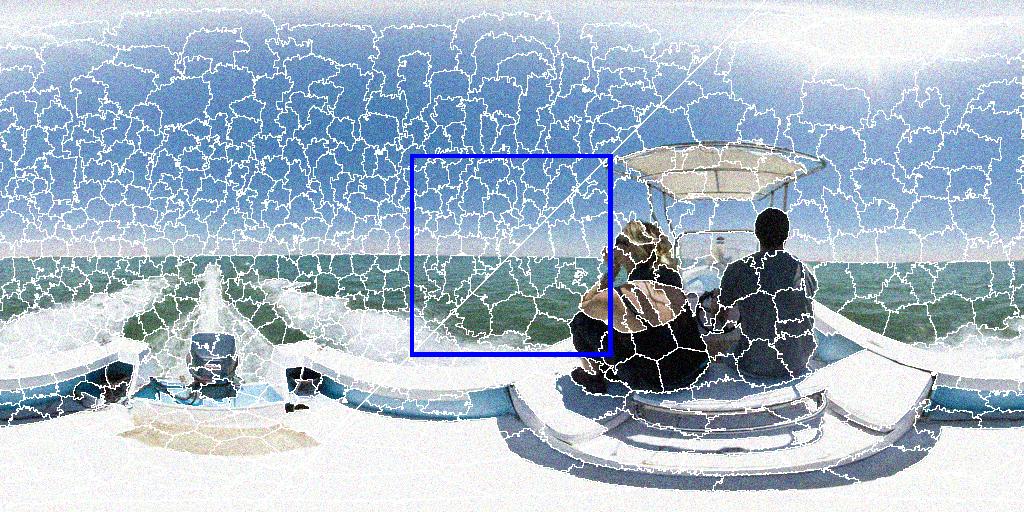}&
\includegraphics[width=\ppp,height=\ppp]{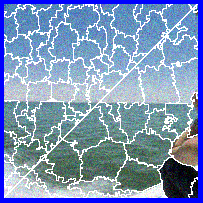}&
\includegraphics[width=\ppp,height=\ppp]{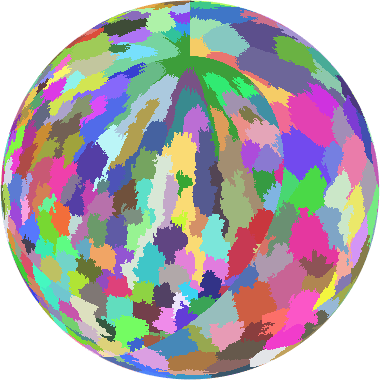}\\[-0.5ex]
\multicolumn{3}{c}{LSC \cite{chen2017}}&
\multicolumn{3}{c}{{SphSLIC-Euc \cite{zhao2018}}} \\[0.5ex]
\includegraphics[width=\ww]{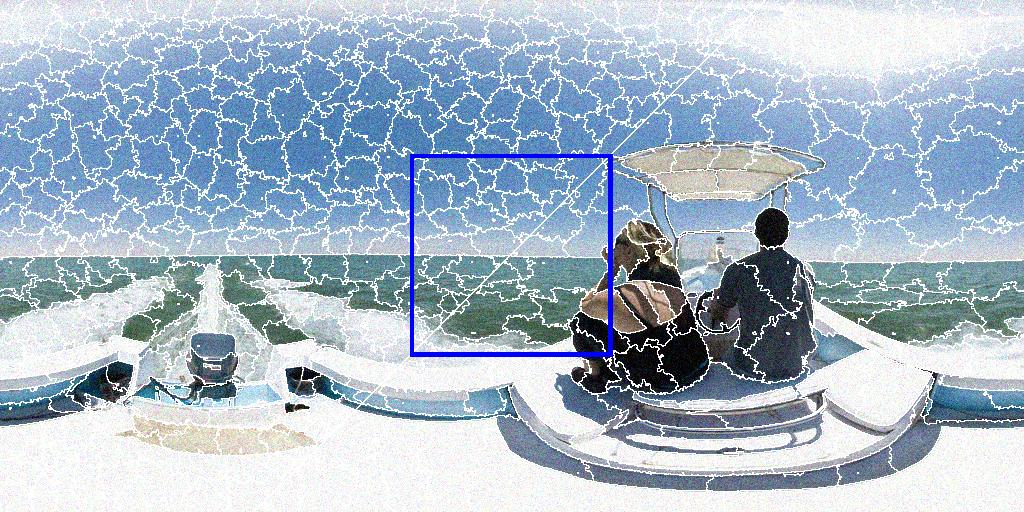}&
\includegraphics[width=\ppp,height=\ppp]{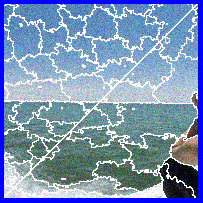}&
\includegraphics[width=\ppp,height=\ppp]{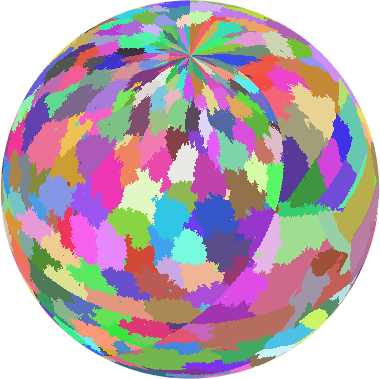}&
% \rotatebox{90}{\hspace{0.35cm} (Cosine-opt)}&
\includegraphics[width=\ww]{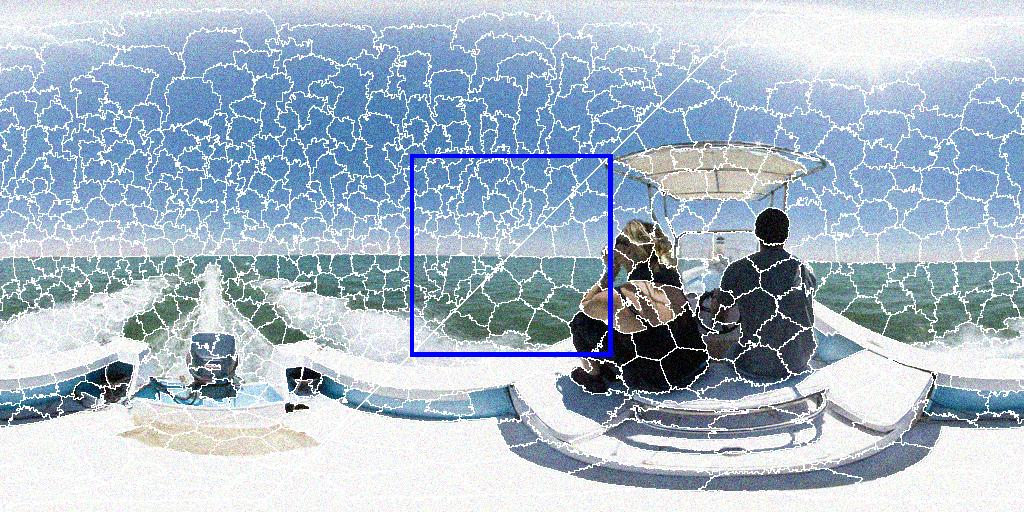}&
\includegraphics[width=\ppp,height=\ppp]{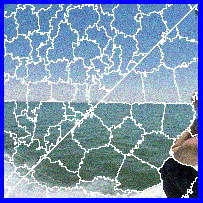}&
\includegraphics[width=\ppp,height=\ppp]{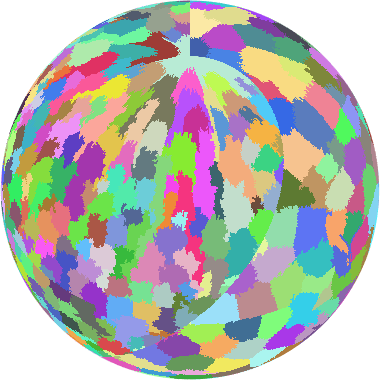}\\[-0.5ex]
\multicolumn{3}{c}{SNIC \cite{achanta2017superpixels}}&
\multicolumn{3}{c}{{SphSLIC-Cos \cite{zhao2018}}} \\[0.5ex]
\includegraphics[width=\ww]{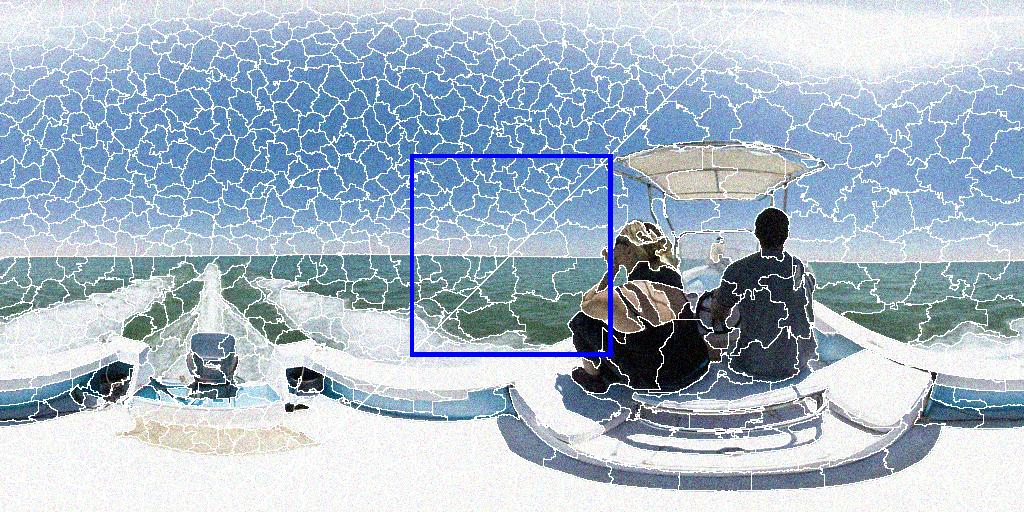}&
\includegraphics[width=\ppp,height=\ppp]{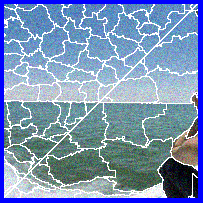}&
\includegraphics[width=\ppp,height=\ppp]{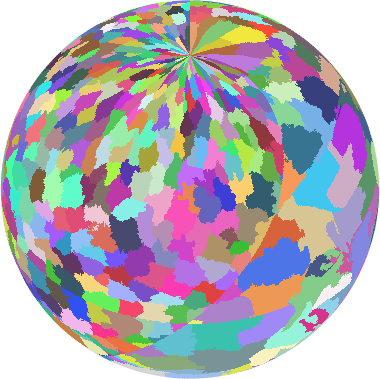}&
\includegraphics[width=\ww]{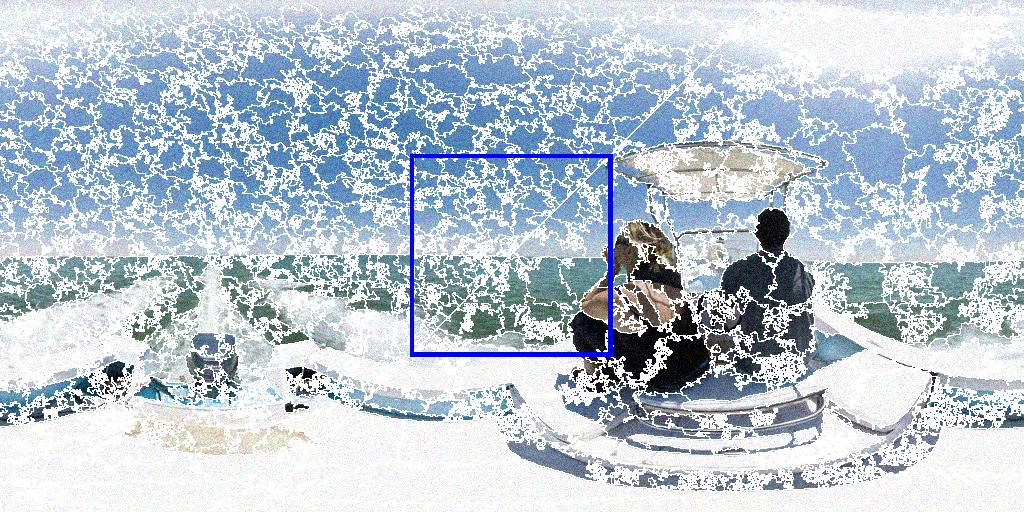}&
\includegraphics[width=\ppp,height=\ppp]{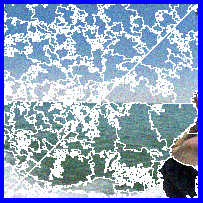}&
\includegraphics[width=\ppp,height=\ppp]{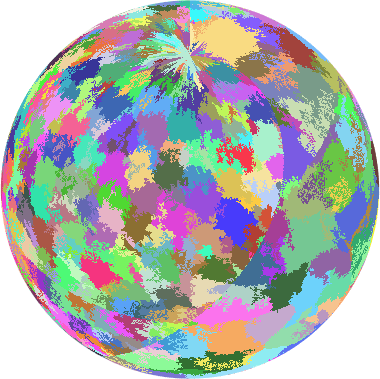}\\[-0.5ex]
\multicolumn{3}{c}{SCALP \cite{giraud2018_scalp}} &
\multicolumn{3}{c}{SphLSC \cite{chen2017,zhao2018}}\\[0.75ex]
\includegraphics[width=\ww]{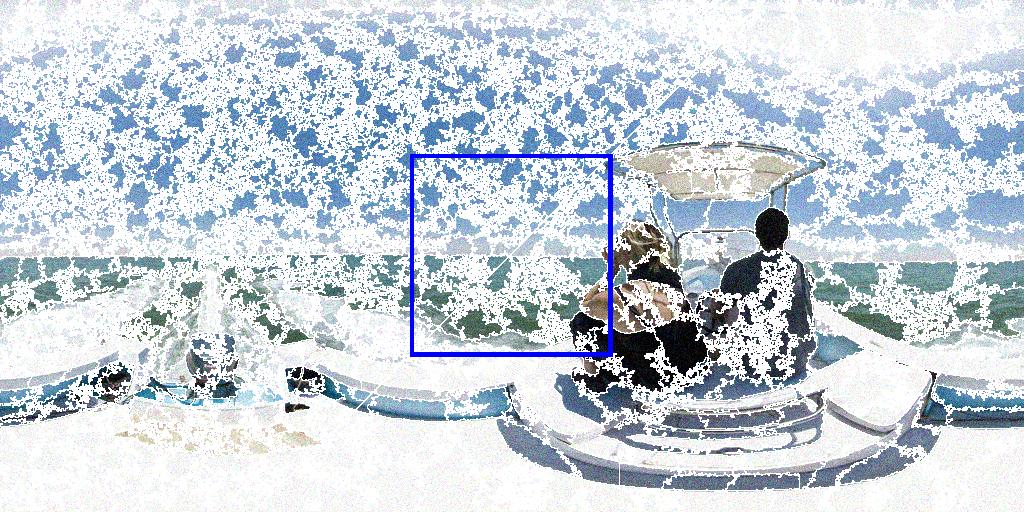}&
\includegraphics[width=\ppp,height=\ppp]{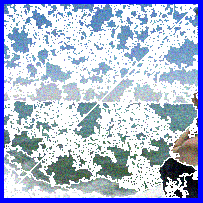}&
\includegraphics[width=\ppp,height=\ppp]{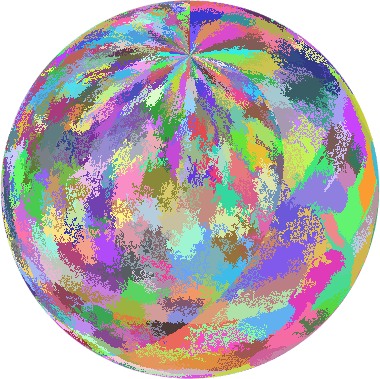}&
\includegraphics[width=\ww]{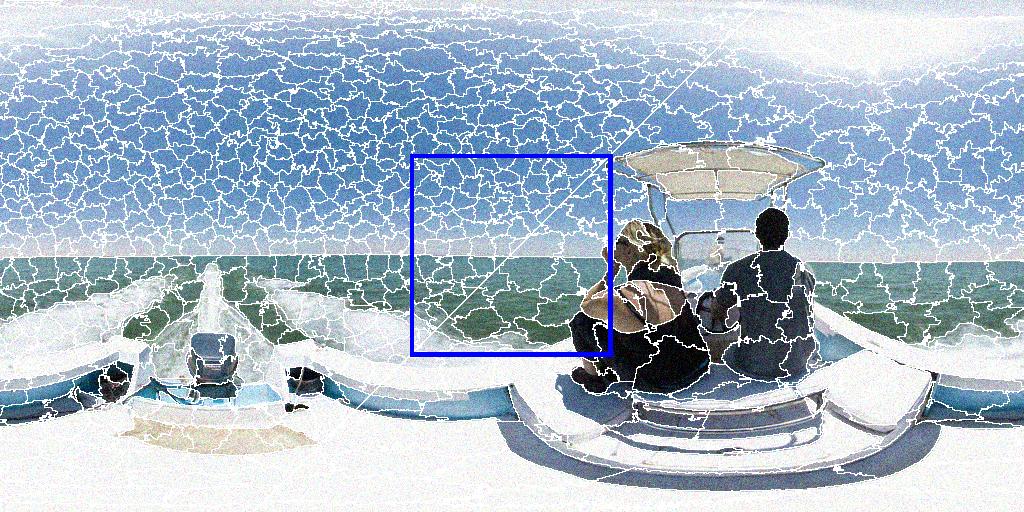}&
\includegraphics[width=\ppp,height=\ppp]{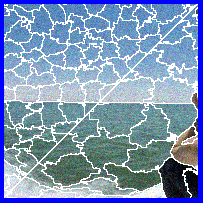}&
\includegraphics[width=\ppp,height=\ppp]{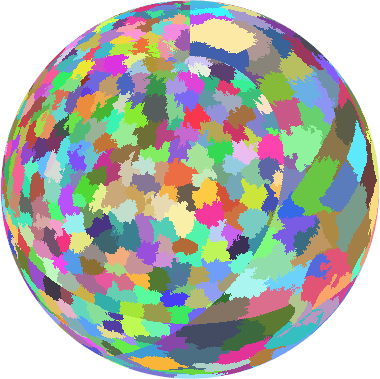}\\[-0.5ex]
 \multicolumn{3}{c}{GMMSP \cite{Ban18}} &
\multicolumn{3}{c}{\textbf{SphSPS}}\\
\end{tabular}
}
\caption{Visual comparison between SphSPS and the
best planar and spherical (underlined) state-of-the-art
methods on noisy PSD images, for two superpixel numbers $K=1200$ (top-left) and $K=400$ (bottom right).
The compared methods may generate very inaccurate superpixels with noisy borders,
while SphSPS remains robust to noise and produces regular superpixels with
smooth boundaries that adhere well to the image contours}
\label{fig:sps_soa_noise_img}
\end{figure*}

 To demonstrate the robustness to noise,
 we also report the results obtained on PSD images affected by the addition of a white Gaussian noise of variance 20.
 Performance in terms of PR/BR, ASA and G-GR are reported in Figure \ref{fig:sps_soa_noise} with segmentation examples given in Figure \ref{fig:sps_soa_noise_img}.

 {\color{review2}
 The proposed SphSPS method appears to be significantly robust to noise.
 While a few planar methods such as SCALP \cite{giraud2018_scalp} or SNIC \cite{achanta2017superpixels} may be moderately impacted by noise,
 all other spherical methods suffer a dramatic loss of accuracy and regularity,
 making SphSPS the only spherical method that is robust to noise.}
 Visually, the superpixels generated by SphSPS remain regular while other methods provide very fuzzy superpixels, or may even severely fail to capture the image objects.
Finally, in Table \ref{table:sps_soa}, we report the quantitative results
 for a given number of $K=1500$ superpixels for both types (initial and noisy PSD images).
 In this Table, we also report Boundary Recall performance (BR) \eqref{br} with respect to Contour Adherence (CD), \emph{i.e.}, the percentage of superpixel border among image pixels, to further express the fuzzy behaviour of several state-of-the-art methods.

\newcommand{\tes}{4mm}
\begin{table}[t!]
 \caption{
Comparison to the state-of-the-art methods on initial$|$noisy PSD images \cite{wan2018},
on the metrics presented in Section \ref{subsec:metrics}.
The ASA \eqref{asa} and G-GR \eqref{circu} results are given for $K=1500$ superpixels, and
CD results for $\text{BR}=0.8$  \eqref{br}.
Blue (bold) and red (underlined) respectively indicate best and second results.
 Some results for the SLIC method on noisy images are estimated by a linear fitting due to an insufficient number of generated superpixels
}
\renewcommand{\arraystretch}{1.1}
\begin{center}
{ \small
 \begin{tabular}{@{\hspace{1mm}}p{3.25cm}@{\hspace{\tes}}c@{\hspace{\tes}}c@{\hspace{\tes}}c@{\hspace{\tes}}c@{\hspace{1mm}}}
 \cline{1-5}
 { Method}& F \eqref{fmeasure} $\uparrow$ &{ CD/BR \eqref{br}} $\downarrow$ &{ ASA \eqref{asa}} $\uparrow$ &{ G-GR \eqref{circu}} $\uparrow$ \\
 \hline
 { SLIC } \cite{achanta2012}   &$0.684|0.542$  	&$0.123|0.156$	&$0.883|0.789$	&$0.438|0.340$\\
 { ETPS } \cite{yao2015}   &$0.695|0.625$	&$0.118|0.194$	&$0.881|0.866$	&$0.426|0.344$\\
 { LSC  }  \cite{chen2017}&$0.668|0.503$	 	&$0.130|0.313$	&$0.890|0.848$	&$0.407|0.374$\\
 { SNIC  } \cite{achanta2017superpixels} 	&$0.672|0.631$		&$0.125|\underline{\color{red}0.137}$	&$0.884|0.866$	& $0.405|0.389$\\
 { SCALP  } \cite{giraud2018_scalp}	&$0.688|0.623$	 &$0.114|\mathbf{\color{blue}0.130}$	&$0.887|0.879$	&$0.451|0.409$\\
 { GMMSP  } \cite{Ban18} &$0.681|0.407$	 	&$0.129|0.332$	&$0.893|0.856$	& $0.375|0.361$\\
 { SphSLIC-Euc  } \cite{zhao2018} &$0.656|0.629$	 	&$0.130|0.142$	&$0.884|0.840$	&$0.495|0.423$\\
 { SphSLIC-Cos  } \cite{zhao2018} &$0.656|0.627$	 	&$0.129|0.141$	&$0.884|0.851$	&$0.501|0.449$\\
 { SphLSC} \cite{chen2017,zhao2018} &$0.683|0.557$	 	&$0.103|0.223$	&$0.895|0.851$	&$0.469|0.390$\\
 { SphSPS ($\gamma$=$0$)}  &${\underline{\color{red}0.706}}|{\underline{\color{red}0.645}}$
 &$\underline{\color{red}0.112}|0.145$&
 ${\underline{\color{red}0.895}}|\underline{\color{red}0.888}$&$\mathbf{\color{blue}0.519}|\underline{\color{red}0.449}$\\
 { SphSPS }&  $\mathbf{\color{blue}0.710}|\mathbf{\color{blue}0.657}$&
 $\mathbf{\color{blue}0.111}|0.141$&
 ${\mathbf{\color{blue}0.899}|\mathbf{\color{blue}0.894}}$&$\underline{\color{red}0.518}|\mathbf{\color{blue}0.454}$\\
 \hline  \\[-5ex]
  \end{tabular}
 }
 \end{center}
 \label{table:sps_soa}
 \end{table}

\subsubsection{Processing time}
The computational cost induced
from the larger 6 dimensional feature space \cite{chen2017} and
the extraction of color and contour information on the shortest path,
is compensated by the fast convergence in a low number of iterations of SphSPS.
Without using the shortest path,
SphSPS generates superpixels in $0.85$s per image of size $512{\times}1024$ pixels and already obtains higher accuracy  ($\text{F}=0.697$)
 than the state-of-the-art methods (see Figure \ref{fig:sps_param_curves}).
With the information on the shortest path, SphSPS obtains significantly higher accuracy in only $2.30$s per image,
that is faster than existing spherical approaches \cite{zhao2018}.

Hence, the proposed optimizations in Section \ref{subsec:sps_path},
enable to consider the large number of pixels contained into the shortest path, while
only increasing the processing time by less than a factor 3.
Finally, with basic multi-threading implementation,
we easily reduce the processing time of SphSPS
to $0.7$s per image.

Further comparison of processing times would necessitate to consider the differences of
implementation and optimization to reflect the computational potential of each method \cite{stutz2016}.
Nevertheless, since our method uses the same $K$-means clustering framework as SLIC \cite{achanta2012},
it is very likely to be able to perform in real-time,
as other works have already proposed such GPU implementations
\cite{neubert2014compact,choi2016subsampling,ban2018glsc}.

\section{Conclusion}

In this work,
we proposed a new paradigm to compute accurate superpixels directly in the 3D acquisition space considering the distorsions induced by the spherical projection to the planar image space.
To this end, we generalized the shortest path approach to extract features between a pixel and a superpixel barycenter.
To consider the color and contour feature information on such path enables
to increase the segmentation performance and the regularity of the generated regions.
We especially showed that respecting the acquisition space geometry
enables to more accurately capture the image objects.

To evaluate the performance of the proposed SphSPS method,
we first addressed the limitation of the existing spherical regularity measure by introducing
a generalized regularity metric measuring the spatial convexity and consistency in the 3D spherical space.
{\color{review2}
We extensively compared SphSPS to the state-of-the-art methods
on the standard natural dataset and a new synthetic set of 360$^\text{o}$ road images.
We demonstrate its superior performance in terms of accuracy, regularity and robustness to noise, with a low computational time.
}

{\color{review3}
Since having both} border accuracy with high
regularity in the acquisition space are crucial for relevant
display and processing of neighboring relationships
in computer vision pre-processing, we truly believe
that the proposed work can be of high interest to the community.
Future works will focus on extending our method to other acquisition spaces and
omnidirectional video acquisitions.

\appendix{}

\subsection*{A. Comparison of proposed G-GR and standard COM metrics}

\begin{figure}[ht!]
\centering
{\scriptsize
\begin{tabular}{@{\hspace{0mm}}c@{\hspace{2mm}}c@{\hspace{2mm}}c@{\hspace{0mm}}}
\includegraphics[width=0.32\textwidth,height=0.24\textwidth]{gr_sph_soa.pdf}&
\includegraphics[width=0.32\textwidth,height=0.24\textwidth]{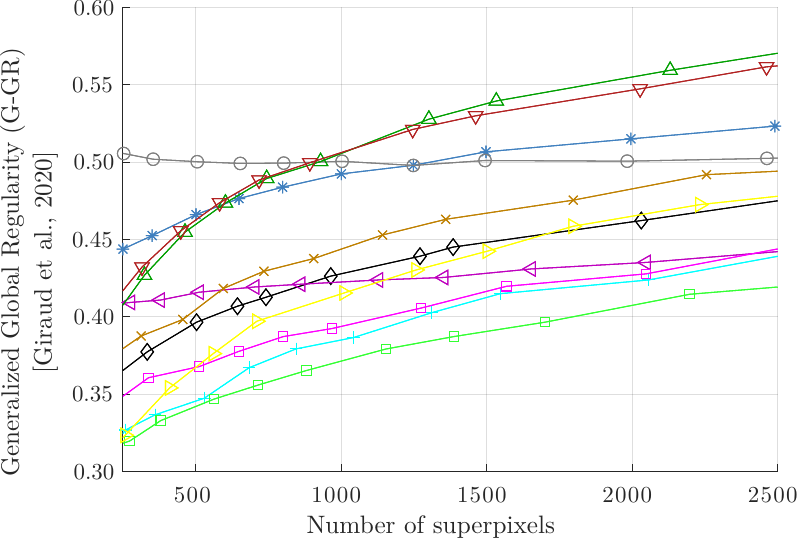}&
\includegraphics[width=0.32\textwidth,height=0.24\textwidth]{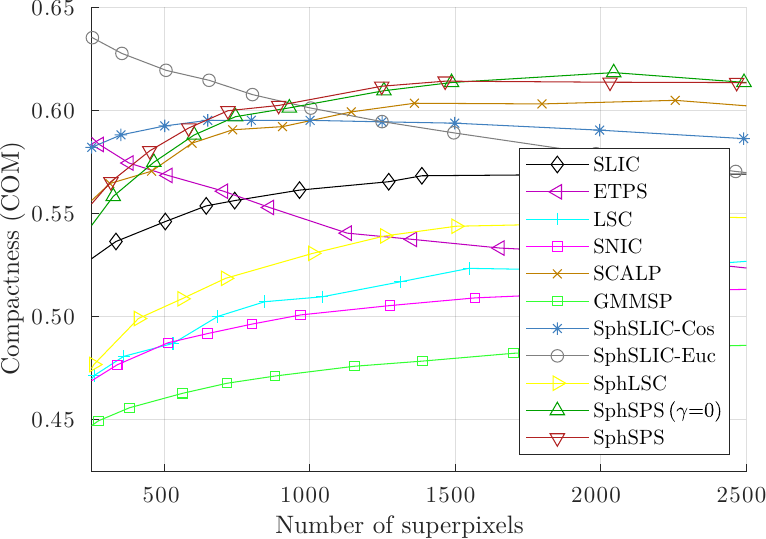}\\
\end{tabular}
}
\caption{Comparison between proposed G-GR (left),
our previous implementation of the G-GR metric \cite{giraud2020_icpr} (center)
and the
spherical compactness COM \cite{zhao2018} metric (right) on the PSD dataset.
The proposed G-GR is able to differentiate spherical from planar methods while being more correlated to the number of generated superpixels
}%
\label{fig:com_res}
\end{figure}

In Figure \ref{fig:com_res}, we report the evaluation of methods
on our previous implementation of the G-GR metric \cite{giraud2020_icpr},
and on the standard compactness measure COM \eqref{circu}, adapted to the spherical space in \cite{zhao2018}.
As discussed in Section \ref{subsec:soa},
the proposed version of the G-GR metric enables to clearly differentiate spherical from planar methods,
\emph{i.e.}, with a significant number of superpixels, all spherical methods obtain higher regularity measure.
Our previous implementation \cite{giraud2020_icpr} uses a downscaling instead of a concave hull to
obtain the final 2D projection of spherical superpixel shapes.
It appears to be more impacted by fuzzy borders, so that the SphLSC method that produces very irregular spherical superpixels (see Figures \ref{fig:sps_soa_img_1}, \ref{fig:sps_soa_img_2} and \ref{fig:sps_soa_img_omni}), is measured less regular than the planar SCALP method.

We also observe that the COM metric even  fails at differentiating planar and spherical methods.
For instance, the SCALP method \cite{giraud2018_scalp} gets higher results than the SphSLIC method \cite{zhao2018}.
It may come from the non robustness of this criteria, which is only seen as a circularity notion and does not take into account the consistency of shape within the decomposition,
while the proposed G-GR metric \eqref{grs}, %based on \cite{giraud2017_jei},
evaluates the regularity of shapes and their consistency according to the acquisition space.

Finally, to further demonstrate the relevance of G-GR over the COM metric,
we compute the correlation between the measured regularity and the number of generated superpixels.
As the number of superpixel increases, their shape is very likely to be more regular, since they less have to stretch to gather homogeneous pixels.
Hence, the regularity is supposed to increase with the superpixel scale, \emph{i.e.}, the number of superpixels.
For all the compared methods in Section \ref{sec:results}, on the PSD dataset,
we report an average correlation between the number of superpixels and the regularity
of $0.3725$ for COM and of $0.6518$ for the proposed G-GR.

\section*{Acknowledgments}
We would like to thank
Ahmed Rida Sekkat, Yohan Dupuis, Pascal Vasseur and Paul Honeine
for sharing some images and their annotations in the frame of the Omniscape projet \cite{sekkat2020omniscape}.

\bibliographystyle{plain}
\bibliography{biblio}

\end{document}